\documentclass[preprint,12pt]{elsarticle}
\usepackage{lineno}
\usepackage[hyphens,spaces,obeyspaces]{url}
\usepackage[colorlinks=true,linkcolor=blue,citecolor=blue]{hyperref}
\usepackage{adjustbox}
\usepackage{array}
\usepackage{xcolor}
\usepackage{colortbl}
\usepackage{multirow}
\usepackage{comment}
\usepackage{pdflscape}
\usepackage{lscape}
\usepackage[flushleft]{threeparttable}
\usepackage{tikz}
\def\checkmark{\tikz\fill[scale=0.4](0,.35) -- (.25,0) -- (1,.7) -- (.25,.15) -- cycle;} 
\usepackage{longtable}
\usepackage{stackengine}
    \newcommand\xrowht[2][0]{\addstackgap[0.5\dimexpr#2\relax]{\vphantom{#1}}}
\usepackage{colortbl}
\usepackage{tablefootnote}
\usepackage{makecell}                         
\usepackage{geometry}

\usepackage{amsmath}
\usepackage[utf8]{inputenc}
\DeclareUnicodeCharacter{2212}{-}
\DeclareUnicodeCharacter{2217}{*} 
\usepackage{pifont}

\usepackage{textcomp}
\modulolinenumbers[1]

\journal{Transportation Research Part C: Emerging Tech.}

\biboptions{authoryear}

\begin{document}

\begin{frontmatter}

\title{Traffic Prediction using Artificial Intelligence: Review of Recent~Advances and Emerging Opportunities}

\author[mymainaddress]{Maryam Shaygan\fnref{collaboration}}
\ead{mshaygan@udel.edu}

\author[mymainaddress]{Collin Meese\fnref{collaboration}}
\ead{cmeese@udel.edu}

\author[xjtlu]{Wanxin Li}
\ead{wanxin.li@xjtlu.edu.cn}

\author[mysecondaryaddress]{Xiaoliang Zhao}
\ead{Xiaoliang.Zhao@bluehalo.com}

\author[mymainaddress]{Mark Nejad\corref{cor1}} \cortext[cor1]{Corresponding author
}
\ead{nejad@udel.edu}

\fntext[collaboration]{These authors contributed equally to this work.}
\address[mymainaddress]{Department of Civil and Environmental Engineering, University of Delaware, Newark, DE 19716, United States}
\address[xjtlu]{Department of Communications and Networking, Xi'an Jiaotong-Liverpool University, Suzhou, Jiangsu 215123, China}
\address[mysecondaryaddress]{BlueHalo, Rockville, MD 20855, United States}

\begin{abstract}
Traffic prediction plays a crucial role in alleviating traffic congestion which represents a critical problem globally, resulting in negative consequences such as lost hours of additional travel time and increased fuel consumption. Integrating emerging technologies into transportation systems provides opportunities for improving traffic prediction significantly and brings about new research problems. In order to lay the foundation for understanding the open research challenges in traffic prediction, this survey aims to provide a comprehensive overview of traffic prediction methodologies. Specifically, we focus on the recent advances and emerging research opportunities in Artificial Intelligence (AI)-based traffic prediction methods, due to their recent success and potential in traffic prediction, with an emphasis on multivariate traffic time series modeling. We first provide a list and explanation of the various data types and resources used in the literature. Next, the essential data preprocessing methods within the traffic prediction context are categorized, and the prediction methods and applications are subsequently summarized. Lastly, we present primary research challenges in traffic prediction and discuss some directions for future research.

\end{abstract}

\begin{keyword}
Traffic prediction \sep Artificial intelligence \sep Intelligent transportation systems \sep Traffic data \sep Deep learning \sep Survey
\end{keyword}

\end{frontmatter}


\section{Introduction}
Traffic congestion is a critical problem with increasingly adverse impacts globally. The consequences are widespread, including increased accident rates, less reliable travel times, additional fuel consumption, excessive air pollution, and deterioration of societal health. For example, the 2019 Urban Mobility Report estimates that congestion in the United States resulted in a yearly 8.8 billion hours of additional travel time and 3.3 billion gallons of increased fuel consumption, totaling a cost of about \$179 billion \citep{schrank2019}. Thus, it becomes critical to optimally utilize transportation infrastructure capacity to reduce congestion, especially in dense urban areas. 

Accurate traffic prediction significantly improves network capacity utilization while also helping alleviate congestion by empowering traffic management centers (TMCs) and road operators to control traffic more effectively. In Table \ref{project}, we provide some practical examples of projects that focus on improving traffic management or deploying traffic predictors to facilitate real-world applications.

In addition, other applications such as route guidance and navigation systems (e.g., Google Maps\footnote{\url{https://www.google.com/maps}}, Waze\footnote{\url{https://www.waze.com/}}, MapQuest\footnote{\url{https://maps.questfordirections.com/}}, Apple Maps \footnote{\url{https://www.apple.com/maps/}}, TomTom GO Navigation \footnote{\url{https://www.tomtom.com/en_us/}}, INRIX\footnote{\url{https://inrix.com/}}, Traffic Spotter \footnote{\url{http://www.trafficspotter.com/}}) can also leverage traffic prediction methods to provide travelers with more accurate information in real-time and more quickly alleviate, or even prevent, congestion. Therefore, developing more intelligent and robust traffic prediction methodologies will be a crucial approach to relieving traffic congestion.

\begin{table}[!ht]
\centering
\begin{adjustbox}{width=\textwidth}
\begin{threeparttable}
\caption{Sample ITS-based urban traffic management projects.}
\label{project}

  \begin{tabular}{llcc}
\hline
\multirow{2}{*}{\textbf{Project}}&\multirow{2}{*}{\textbf{ Objectives }}&\multirow{2}{*}{\textbf{Project Sponsor}}&\textbf{Year of start}\\

&&&\textbf{$\&$ completion}\\\hline

\multirow{2}{*}{Artificial Intelligence and ML enhanced
\tablefootnote{\url{https://deldot.gov/Programs/itms/index.shtml?dc=projects}}}&\textendash Adopting
new technologies such as &  &\multirow{4}{*}{2020-Ongoing}\\
\multirow{2}{*}{Integrated Transportation Management}
&AI and ML to achieve better transportation &Federal Highway \\
\multirow{2}{*}{ System (AI-ITMS) Deployment Program}&
management results $\&$ AI and ML enhanced &Administration, US&\\
&integrated transportation&\\\hline

An Evaluation of the
Valley Metro–Waymo
\tablefootnote{\url{https://www.transit.dot.gov/research-innovation/evaluation-valley-metro-waymo-automated-vehicle-ridechoice-mobility-demand}}
&\multirow{2}{*}{\textendash  Understand the potential behavioral impacts} 
 &Research and Innovative 
&\multirow{3}{*}{2019-2020}\\
 Automated Vehicle RideChoice
Mobility on &\multirow{2}{*}{of AV
 mobility on demand services}&Technology \\
Demand Demonstration&&Administration, US&\\\hline

Mobility on Demand (MOD)
Sandbox \tablefootnote{\url{https://www.transit.dot.gov/sites/fta.dot.gov/files/2021-06/FTA-Report-No-0197.pdf}}
&\multirow{3}{*}{\textendash  Improves transit trip planning}&Research and Innovative 
&\multirow{3}{*}{2017-2019}\\
Demonstration: (TriMet)
OpenTripPlanner
&&Technology &\\
 (OTP) Shared-Use Mobility&&Administration, US&\\\hline

Mobility on Demand (MOD)
Sandbox \tablefootnote{\url{https://www.transit.dot.gov/research-innovation/mobility-demand-mod-sandbox-demonstration-vermont-agency-transportation-vtrans}}
&\multirow{2}{*}{\textendash  Provide an advanced statewide trip
planner} &Research and Innovative 
&\multirow{3}{*}{2017-2019}\\
Demonstration: Vermont Agency of  &\multirow{2}{*}{to present flexible transit services}&Technology &\\
Transportation
(VTrans) OpenTripPlanner&&Administration, US&\\\hline

Traffic Signal Control in the Transition  \tablefootnote{\url{https://www.its-edulab.nl/wp-content/uploads/MSc_Thesis_Femke_van_Giessen.pdf}}&\textendash Present how existing traffic signal controllers &Dutch Traffic and  &\multirow{2}{*}{2021-2021}\\
from Manual to Automated Driving& can cope with hybrid traffic environment &Transport Laboratory&\\\hline

\multirow{2}{*}{CAVITE-LAGUNA (CALA) Expressway \tablefootnote{\url{https://www.dpwh.gov.ph/dpwh/PPP/projs/cala}}}&\multirow{3}{*}{\textendash Reduce traffic congestion}&Department of Public&\multirow{3}{*}{2019-Ongoing}\\
\multirow{2}{*}{Project}&& Works and Highways,\\
&& Philippines&\\\hline

\multirow{3}{*}{Keys COAST Connected and Automated  \tablefootnote{\url{https://sunguide.info/its-program/projects/}}}&\textendash Improve mobility and safety performance &Office of Research and 
&\multirow{4}{*}{2020-Ongoing}\\
\multirow{3}{*}{Vehicle (CAV) Project}&by establishing connectivity between various &Development,State of &\\
& modes and a connected vehicle traffic &Florida, Department &\\
&signal system (CVTSS)&of Transportation&\\\hline

\multirow{3}{*}{5G for Connected and Automated Road  \tablefootnote{\url{https://cordis.europa.eu/project/id/825012}}}&\textendash Build a 5G-enabled corridor to conduct  &\multirow{4}{*}{European Union}&\multirow{4}{*}{2018-2022}\\
\multirow{3}{*}{Mobility in the European Union}& cross-border trials and will deploy a mixture &&\\
&of 5G micro-/macro-cells for ubiquitous &&\\
&C-V2X connectivity&\\\hline

End-to-end Hardware Implementation of  \tablefootnote{\url{https://cordis.europa.eu/project/id/849921}}&\textendash Develop an alternative for DL tasks to reduce &\multirow{3}{*}{European Union}&\multirow{3}{*}{2019-2020}\\
Artificial Neural
Networks for Edge  &space-, power- and cost-consuming  while  also &&\\
Computing in Autonomous Vehicles&allowing  for the advancement of AV technologies&&\\\hline

\end{tabular}
  \end{threeparttable}
  \end{adjustbox}
\end{table}

There are two primary classifications for traffic congestion: recurrent and non-recurrent. While recurrent congestion generally occurs when the road capacity is insufficient to accommodate the existing vehicle volume, non-recurrent congestion mainly results from accidents, disabled vehicles, inclement weather, work zones, and special events. A primary problem in traffic forecasting is accurately predicting the outcome of non-recurrent traffic events, which account for about 50\% of all traffic congestion according to the Federal Highway Administration (FHWA) \citep{fhwaCongestion}. Thus, traffic prediction during non-recurrent events is a critical research area that needs more attention. Despite this, most existing studies focus on short-term and long-term recurrent traffic prediction problems, especially during rush hours.

In short-term traffic forecasting, the prediction horizon usually ranges from seconds to multiple hours ahead of time and utilizes both historical and current traffic information \citep{qu2019daily}, making it applicable for traffic control route planning, etc. However, the precise definition of long-term forecasting, to differentiate it from short-term approaches, is largely ambiguous, and many researchers classify short-term and long-term forecasts differently \citep{manibardo2021deep}. More specifically, some papers differentiate the two approaches solely based upon a quantitative threshold of the prediction horizon \citep{abdulhai2002short, chrobok2004different,bogaerts2020graph}. In contrast, others assume short-term relates to the number of forecasting steps output by the model \citep{zhang2014hybrid, cai2016spatiotemporal,ma2020multi}. However, both classification approaches result in ambiguous definitions for some models (e.g., multi-output models).

Moreover, it is imperative to examine emerging technologies, such as autonomous vehicles (AVs), Internet of Things (IoT), novel machine learning techniques, and secure wireless connectivity and information sharing schemes, as they have the potential to impact how we predict and manage traffic drastically. For example, blockchain is a secure-by-design network technology that has recently been explored as a way to manage decentralized traffic data for modeling traffic prediction states, such as congestion probability estimation \citep{9107472}. Additionally, new machine learning concepts, including federated learning (FL), have attracted significant attention in traffic flow prediction applications \citep{meese2022bfrt}. Specifically, their ability to train new models in a privacy-preserving and decentralized manner presents a more efficient and secure method for utilizing the enormous amounts of existing traffic data \citep{9082655}. Furthermore, AVs can transform existing mobility and congestion patterns, laying the foundation for sustainable smart cities. In particular, AVs can better cope with many of the current problems (e.g., congestion) experienced on the roads today thanks to the real-time Vehicle-to-everything (V2X) communication with critical infrastructure and stakeholders, such as vehicle to traffic control centers, Road Side Units (RSUs), and other vehicles \citep{miglani2019deep}. However, this reality only becomes possible if AVs become both common and affordable for society \citep{litman2020autonomous}. Additionally, there still exists a large knowledge gap regarding the impacts of these types of vehicles within mixed traffic scenarios \citep{di2021survey}.

Most of the new AI methods in this review seek to leverage DL approaches' strong non-linear modeling capabilities to capture the complex spatio-temporal relationships between the multivariate traffic time series. Over the years, research has uncovered many unique properties of multivariate traffic time series, compared to other multivariate time series (e.g., product recommendations), such as non-stationary relationships, global spatial correlations with strong locality considerations, high temporal variability across multiple time-frames (e.g., recent, daily-periodic, seasonal), and also recurrent, non-recurrent, and external network impacts (e.g., accidents, weather, events). Throughout the paper, we refer to these models and their associated applications as multivariate traffic prediction (MTP) approaches.

This survey provides an introduction to, and extensive review of, the recent advances and emerging opportunities in AI-based traffic prediction methodologies. To the best of our knowledge, this is the first survey to provide an exhaustive review of both historical and recent AI-based traffic prediction approaches and associated data, discussed within the context of the common traffic prediction states and their practical applications, as well as the existing challenges and necessary future research directions. Thus, we believe this survey will be of paramount interest to researchers seeking a survey paper within the context of the cutting-edge Deep Learning (DL) approaches.

Concretely, we focus the literature review on the following broad research questions:
\begin{itemize}
    \item How are different data type resources and classifications utilized within the literature for traffic prediction?
    \item Which preprocessing methods are applicable and demonstrated to be most effective for traffic prediction applications? 
    \item How are varying prediction models and methodologies leveraged within the literature; and how do they apply to various traffic prediction problems? 
    \item What are the fundamentally critical open research issues and challenges presenting barriers in existing traffic prediction research; and which of the emerging technologies are best suited for addressing these problems? 
\end{itemize}

The remainder of this paper is organized as follows. Section \ref{Sec2} provides an overview of traffic prediction survey papers, comparing our work to the existing surveys. Section \ref{Sec3} summarizes the most commonly used traffic prediction research datasets found in the literature. Section \ref{Sec4} discusses the traffic prediction methods, including classical and machine learning methods. Section \ref{Sec5} introduces some common traffic prediction states and associated applications. Next, we discuss the open research challenges and list promising future research directions in Section \ref{Sec6}. Finally, we conclude the paper in Section \ref{Sec7}. Table \ref{table:acr} has summarized all acronyms used in this survey.

\begin{table}[t]
\centering
\footnotesize
\caption{List of acronyms}
\setlength\extrarowheight{0pt}
\begin{adjustbox}{width=\textwidth}
    \begin{tabular}{l  p{6.5cm} |l  p{6.5cm}}
    \hline
     \textbf{Acro.} & \textbf{Description}  &  \textbf{Acro.} & \textbf{Description} \\ \hline
     AE&Autoencoder &GB&Gradient Boosting\\
     AI&Artificial Intelligence &HA & Historical Average \\
      ANN & Artificial Neural Network  &HMM & Hidden Markov Model\\
       API&Application Programming  &IoT& Internet of the Things\\
       ARIMA & Autoregressive Integrated Moving Average  &IoV&Internet of Vehicles\\
     AV& Autonomous Vehicle &ITS&Intelligent Transportation System\\
     AVI&Automatic Vehicle Identification &KF  &Kalman Filter\\ 

    BN& Bayesian Networks &kNN & K-nearest Neighbors\\
     BP&backpropagation &LSTM&Long short-term Memory\\
     CAV&Connected Autonomous vehicle&ML& Machine Learning\\
    CFCD &Cellular Floating Car Data & MLP&Multi-Layer Perceptron\\
    CNN&Convolutional Neural Network &MTP& Multivariate Traffic Prediction\\
     DBM& Deep Boltzmann Machine &NN&Neural Network\\
     DBN&Deep Belief Network &OD&Origin-Destination  \\
     DL&Deep Learning &POD&Points of Dispensing  \\
     DNN& Deep Neural Network  &RBM&Restricted Boltzmann Machine \\
     DT & Decision Tree &RF&Random Forest \\
ETC&Electronic Toll Collection &RNN&Recurrent Neural Network\\

      FCD& Floating Car Data &RSU&Road Side Unit   \\
       FCN&Fully-Connected Neural Network &STTP&Short-term Traffic Prediction \\
      FHWA&Federal Highway Administration&SAE&Stacked Autoencoder\\

 FL& Federated  Learning &SVR&Support Vector Regression\\

     FNN&Feedforward Neural Network (FFNN) &TCN&Temporal Convolution Network\\
     
 GCN&Graph Convolutional Network  &TMC&Traffic Management Center\\
 GMM&Gaussian Mixture Model  &VAR&Vector Auto-Regressive\\
 GPS&Global Positioning Systems  &WNN&Wavelet Neural Network\\
GRU&Gated Recurrent Unit&XGBoost&Extreme Gradient Boosting\\

     \hline
   \end{tabular}
   \end{adjustbox}
\label{table:acr}
\end{table}

\section{Prediction aspects assessed in previous reviews}
\label{Sec2}

This section outlines related survey papers on traffic prediction and briefly discusses their varying perspectives and approaches. \citet{vlahogianni2004short} contributed a comprehensive survey on traditional short-term traffic prediction methods and their applications. A decade later, the same authors discussed ten significant challenges for newly developed short-term traffic prediction approaches \citep{vlahogianni2014short}. Later, \citet{barros2015short} explored data-driven and model-driven approaches for short-term traffic prediction to highlight their capabilities. \citet{li2018brief} also suggested some possible future research directions after a review of several short-term traffic prediction techniques. Besides, \citet{nagy2018survey} summarized the data resources and provided an overview of existing Machine Learning (ML) traffic prediction methods. In the same year, \citet{ermagun2018spatiotemporal} conducted a comprehensive survey and examined long-term traffic forecasting literature to identify critical spatial components and their role in traffic forecasting. The authors also proposed possible future research directions for improving long-term traffic prediction. Meanwhile, \citet{angarita2019taxonomy} proposed a new taxonomy categorizing the traffic forecasting regression problem in terms of both ML modeling and traffic specifications, focusing on the core characteristics that may alter the complexity of modeling traffic forecasting regression problems. 

In addition, both \cite{boukerche2020machine} and \citet{lana2018road} reviewed models that leverage classical approaches and some early deep learning methods. Other relevant surveys attempted to elucidate how different deep learning models are utilized in various transportation applications, including \citet{wang2019enhancing} and \citet{do2019survey}. More recently, \citet{akhtar2021review} compared and contrasted some popular machine learning techniques for predicting traffic congestion. Furthermore, recurrent neural network, convolutional neural network, and feedforward neural network models for traffic prediction were reviewed by \citet{tedjopurnomo2020survey}. In contrast, \citet{ye2020build} instead concentrated on outlining graph-based deep learning architectures from the literature, emphasizing applications in the general traffic domain. 

Urban flow prediction using spatio-temporal data was systematically outlined by \citet{xie2020urban}, and some traditional machine learning and deep learning models were discussed within the context of the urban flow prediction application. Later, \citet{gobezie2020machine} reviewed robust traffic flow prediction concerning different approaches, development tools, and benchmark performance evaluation metrics. However, \citet{lee2021short} also suggested possible directions for future work in deep neural network (DNN)-based short-term traffic prediction (STTP) by reviewing input data representation methods, DNN approaches application domains, and dataset availability. 

In another work, \citet{yin2021deep} summarized various deep learning methods and collected and organized a comprehensive list of publicly available traffic datasets to help future research. Additionally, the authors conducted experiments to investigate the performance of the outlined methods. Meanwhile, \citet{shi2019survey,shi2021comprehensive} focused on reviewing the hybrid deep learning methods, detailing their advantages while highlighting their ability to increase the accuracy of traffic prediction and better capture dependency in multi-dimensional studies. Later, \citet{yuan2021survey} focused on surveying traffic data types based on their spatial and temporal dimensions and data preprocessing techniques. Specifically, traffic classification, generation, and forecasting are discussed to show how the existing methods can address traffic prediction challenges.

Moreover, \citet{wang2020deep} reviewed DL models' success in automatic feature representation learning and their ability to be widely applied in various spatio-temporal data mining (STDM) tasks such as predictive learning, anomaly detection, and classification. Another related survey, \citet{manibardo2021deep}, indicated why DL may not be the best modeling technique for every short-term traffic prediction application, which unveils some caveats unconsidered to date that the community in prospective studies should address. Lastly, \citet{lana2021data} identified the main actionability gaps in the data-based modeling workflow and how big data in ITS can be used to learn and adapt data-driven models for efficiently operating ITS assets, systems, and processes.

Recently, with the ongoing advancements in ITS and autonomous vehicle technologies, some efforts have been made to explore and design traffic prediction methods concentrated on autonomous vehicle environments. \citet{nguyen2018deep} focused on reviewing not only the recent DL studies in traffic data processing, such as transportation network representation, traffic flow forecasting, traffic signal control, automatic vehicle detection, traffic incident processing, and travel demand prediction, but also the effects and challenges of the autonomous driving and driver behavior on traffic prediction. In \cite{kumar2020review}, the authors reviewed some recent traffic prediction research within the self-driving application context. A comprehensive survey on DL applications and practical approaches for traffic prediction in autonomous vehicles is also provided in \cite{miglani2019deep}, where the main unresolved challenges in this area are presented to help guide future research.

\begin{figure}[t]
    \centering
    \includegraphics[width=.88\textwidth]{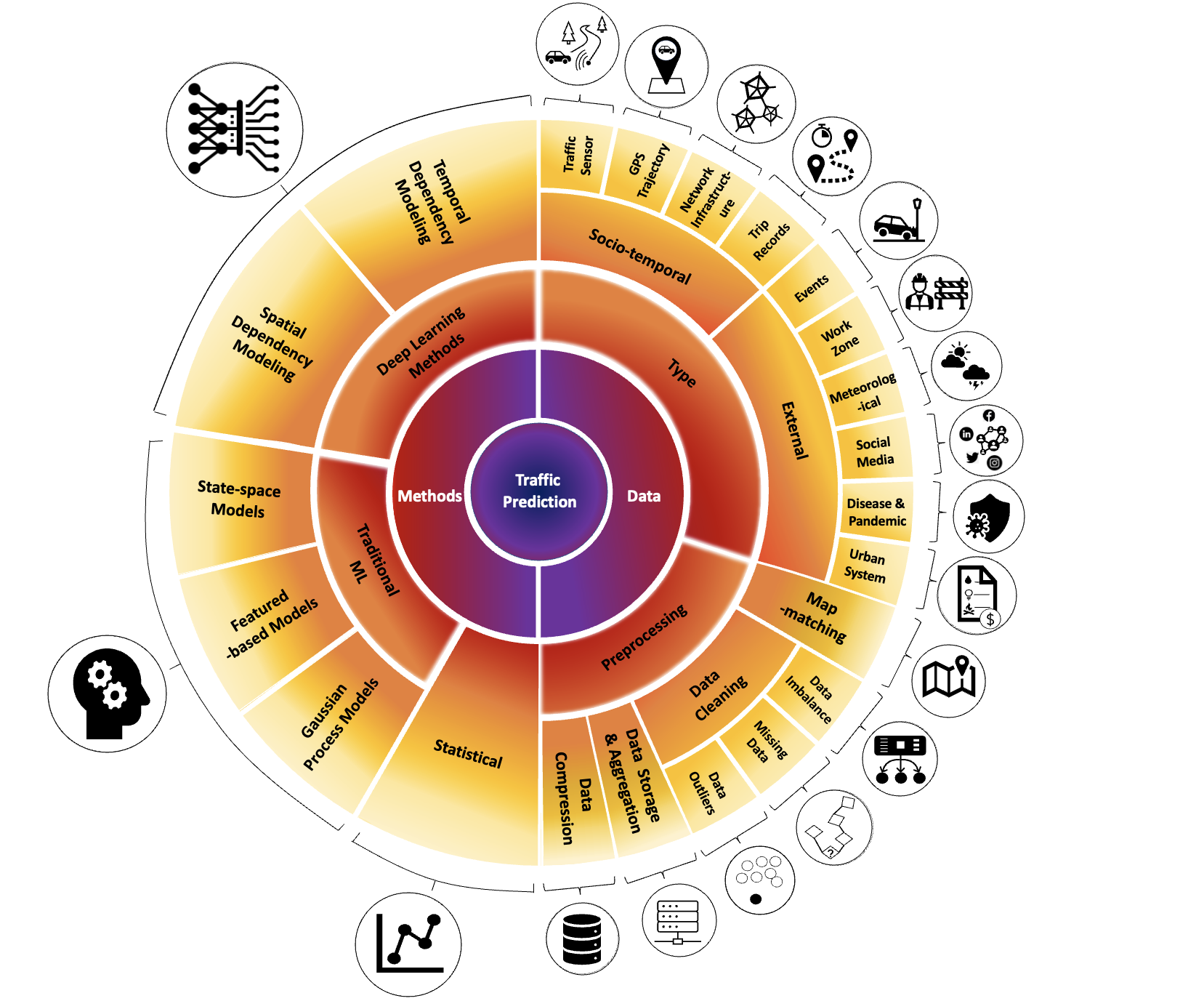}
    \caption{The structure of traffic prediction based on the literature}
      \label{fig:ppp}
\end{figure}

While many related traffic prediction survey papers exist, they are generally focused on specific technical areas or subsets of the methods due to the density of approaches within the literature. On the other hand, not all surveys relate the recent advances to practical applications and future research directions. Besides, these surveys are also segmented across various discipline-specific journals and publication venues, making it challenging for readers to assess the current state of traffic prediction methods altogether. As the literature continues to become more diverse and interdisciplinary, it is necessary to provide a review that considers the literature comprehensively. Consequently, the primary contribution of this survey paper is to provide a systematic and extensive review of the existing traffic prediction literature as a whole, focusing on highlighting the recent advances and emerging research opportunities in Artificial Intelligence (AI)-based traffic prediction methods. To this end, we categorize the structure of the existing traffic prediction literature as shown in Fig. \ref{fig:ppp}. Additionally, to aid the reader in distinguishing our specific contributions from the myriad of existing surveys, we compare our survey to the related works in Table \ref{allrev}. 

\begin{landscape}

 \begin{table}[!ht]
\tiny
\centering
\begin{threeparttable}
\caption{Comparison with related survey articles.}
\label{allrev}
\begin{tabular}{lllllp{0.001\textwidth}p{0.001\textwidth}p{0.001\textwidth}p{0.001\textwidth}p{0.001\textwidth}p{0.001\textwidth}p{0.001\textwidth}p{0.001\textwidth}p{0.001\textwidth}p{0.001\textwidth}p{0.001\textwidth}p{0.001\textwidth}p{0.001\textwidth}p{0.001\textwidth}p{0.001\textwidth}p{0.001\textwidth}p{0.001\textwidth}p{0.001\textwidth}p{0.001\textwidth}p{0.001\textwidth}p{0.001\textwidth}p{0.001\textwidth}p{0.001\textwidth}p{0.001\textwidth}p{0.001\textwidth}p{0.001\textwidth}p{0.001\textwidth}p{0.001\textwidth}p{0.001\textwidth}p{0.001\textwidth}p{0.001\textwidth}l}
\hline
\multicolumn{5}{c}{\multirow{-12}{*}{\textbf{Criteria}}}&
\rotatebox{50}{\citet{vlahogianni2004short}}&
\rotatebox{50}{\citet{vlahogianni2014short}}&
\rotatebox{50}{\citet{barros2015short}}&
\rotatebox{50}{\citet{nguyen2018deep}}&
\rotatebox{50}{\citet{li2018brief}}&
\rotatebox{50}{\citet{nagy2018survey}}&
\rotatebox{50}{\citet{lana2018road}}&
\rotatebox{50}{\citet{ermagun2018spatiotemporal}}&
\rotatebox{50}{\cite{miglani2019deep}}&
\rotatebox{50}{\citet{wang2019enhancing}}&
\rotatebox{50}{\citet{do2019survey}}&
\rotatebox{50}{\citet{shi2019survey}}&
\rotatebox{50}{\citet{angarita2019taxonomy}}&
\rotatebox{50}{\cite{boukerche2020machine}} &
\rotatebox{50}{\citet{tedjopurnomo2020survey}}&
\rotatebox{50}{\citet{ye2020build}}&
\rotatebox{50}{\citet{xie2020urban}}&
\rotatebox{50}{\citet{wang2020deep}}&
\rotatebox{50}{\citet{gobezie2020machine}}&
\rotatebox{50}{\citet{akhtar2021review}}&
\rotatebox{50}{\citet{lee2021short}} &
\rotatebox{50}{\citet{yin2021deep}}&
\rotatebox{50}{\citet{shi2021comprehensive} }&
\rotatebox{50}{\citet{yuan2021survey}}&
\rotatebox{50}{\citet{manibardo2021deep}}&
\rotatebox{50}{\citet{lana2021data}}&\rotatebox{50}{ Current Survey}&&\\\hline

\multirow{25}{*}&\multirow{25}{*}{Data}&\multirow{9}{*}{Type}&\multirow{4}{*}{\shortstack[c]{Spatio-\\Temporal}}&Fixed Position Sensors&$\checkmark$&$\checkmark$&$\checkmark$&$ \times$&$ \times$&$\checkmark$&$\checkmark$&$ \mathtt{\sim}$&$\checkmark$&$\checkmark$&$\checkmark$&$\checkmark$&$\checkmark$&$\mathtt{\sim}$&$\checkmark$&$ \checkmark$&$ \checkmark$&$ \checkmark$&$ \checkmark$&$ \checkmark$&$ \mathtt{\sim}$&$ \checkmark$&$\times$&$ \checkmark$&$\checkmark$&$ \checkmark$&$ \checkmark$\\\cline{5-32}

&&&&Trajectory&$ \mathtt{\sim}$&$\checkmark$&$\checkmark$&$ \times$&$ \times$&$\checkmark$&$\checkmark$&$ \mathtt{\sim}$&$\checkmark$&$\checkmark$&$\checkmark$&$\checkmark$&$\checkmark$&$\mathtt{\sim}$&$\checkmark$&$ \checkmark$&$ \checkmark$&$ \checkmark$&$ \checkmark$&$ \checkmark$&$ \mathtt{\sim}$&$ \checkmark$&$\times$&$ \checkmark$&$\checkmark$&$ \checkmark$&$ \checkmark$\\\cline{5-32}

&&&&Trip Records&$ \mathtt{\sim}$&$ \times$&$ \mathtt{\sim}$&$ \times$&$ \times$&$ \times$&$\mathtt{\sim}$&$ \times$&$ \times$&$ \times$&$ \times$&$ \times$&$ \times$&$ \times$&$ \times$&$ \times$&$ \times$&$ \times$&$ \times$&$ \times$&$ \times$&$ \mathtt{\sim}$&$\times$&$\times$&$\times$&$ \times$&$ \checkmark$\\\cline{5-32}

&&&&Network Infrastructure&$ \times$&$ \times$&$ \times$&$ \times$&$ \times$&$ \times$&$ \times$&$ \times$&$ \times$&$ \times$&$ \times$&$ \times$&$ \times$&$ \times$&$ \checkmark$&$ \times$&$ \times$&$ \times$&$ \times$&$ \times$&$ \times$&$ \times$&$\times$&$\times$&$\times$&$\mathtt{\sim}$&$ \checkmark$\\\cline{4-32}

&&&\multirow{6}{*}{External}&Events&$ \times$&$ \mathtt{\sim}$&$ \mathtt{\sim}$&$ \times$&$ \times$&$ \mathtt{\sim}$&$\checkmark$&$ \times$&$\checkmark$&$\mathtt{\sim}$&$\mathtt{\sim}$&$\checkmark$&$ \times$&$\mathtt{\sim}$&$ \mathtt{\sim}$&$ \mathtt{\sim}$&$ \mathtt{\sim}$&$ \checkmark$&$\mathtt{\sim}$&$\mathtt{\sim}$&$\mathtt{\sim}$&$\checkmark$&$\times$&$ \checkmark$&$ \checkmark$&$ \checkmark$&$ \checkmark$\\\cline{5-32}

&&&&Meteorological&$ \times$&$ \mathtt{\sim}$&$\checkmark$&$ \times$&$ \times$&$\checkmark$&$\checkmark$&$ \times$&$\checkmark$&$\mathtt{\sim}$&$\mathtt{\sim}$&$\checkmark$&$\checkmark$&$\mathtt{\sim}$&$ \mathtt{\sim}$&$ \mathtt{\sim}$&$ \mathtt{\sim}$&$ \checkmark$&$\mathtt{\sim}$&$\mathtt{\sim}$&$\mathtt{\sim}$&$\checkmark$&$\times$&$ \checkmark$&$ \checkmark$&$ \checkmark$&$ \checkmark$\\\cline{5-32}

&&&&Social Media&$ \times$&$ \times$&$\mathtt{\sim}$&$ \times$&$ \times$&$ \times$& $\checkmark$&$ \times$&$\checkmark$&$\mathtt{\sim}$&$\mathtt{\sim}$&$ \times$&$ \times$&$ \times$&$ \times$&$ \times$&$ \times$&$ \times$&$ \times$&$\mathtt{\sim}$&$ \times$&$ \times$&$\times$&$\mathtt{\sim}$&$\mathtt{\sim}$&$\checkmark$&$ \checkmark$\\\cline{5-32}

&&&&Disease $\&$ Pandemics&$ \times$&$ \times$&$ \times$&$ \times$&$ \times$&$ \times$&$ \times$&$ \times$&$ \times$&$ \times$&$ \times$&$ \times$&$ \times$&$ \times$&$ \times$&$ \times$&$ \times$&$ \times$&$ \times$&$ \times$&$ \times$&$ \times$&$\times$&$\times$&$\times$&$ \times$&$ \checkmark$\\\cline{5-32}

&&&&Work Zone&$ \times$&$ \times$&$\mathtt{\sim}$&$ \times$&$ \times$&$ \mathtt{\sim}$&$ \times$&$ \times$&$ \checkmark$&$ \times$&$\mathtt{\sim}$&$ \times$&$ \times$&$ \times$&$ \times$&$ \times$&$ \times$&$ \times$&$ \times$&$ \times$&$ \times$&$ \times$&$ \times$&$\times$&$\times$&$ \times$&$ \checkmark$\\\cline{5-32}

&&&&Urban System&$ \times$&$ \times$&$\mathtt{\sim}$&$ \times$&$ \times$&$ \times$&$ \times$&$ \times$&$ \times$&$ \times$&$ \times$&$ \times$&$ \times$&$ \times$&$ \times$&$ \times$&$ \times$&$ \times$&$ \times$&$ \times$&$ \times$&$ \times$&$\times$&$\times$&$\times$&$ \mathtt{\sim}$&$ \checkmark$\\\cline{3-32}

&&\multicolumn{3}{l}{Resolution}&$\mathtt{\sim}$&$ \mathtt{\sim}$&$ \mathtt{\sim}$&$ \times$&$ \times$&$ \mathtt{\sim}$&$ \mathtt{\sim}$&$\mathtt{\sim}$&$ \mathtt{\sim}$&$\mathtt{\sim}$&$\mathtt{\sim}$&$ \times$&$\checkmark$&$ \mathtt{\sim}$&$\checkmark$&$\checkmark$&$\times$&$ \mathtt{\sim}$&$\checkmark$&$ \mathtt{\sim}$&$\checkmark$&$\mathtt{\sim}$&$\times$&$\mathtt{\sim}$&$\checkmark$&$ \mathtt{\sim}$&$ \checkmark$\\\cline{3-32}

&&\multirow{4}{*}{\shortstack[l]{Prepro-\\cessing}}&\multicolumn{2}{l}{Map Matching}&$ \times$&$ \times$&$ \times$&$ \times$&$ \times$&$ \times$&$ \times$&$ \times$&$ \times$&$\mathtt{\sim}$&$ \times$&$ \times$&$\checkmark$&$ \mathtt{\sim}$&$\mathtt{\sim}$&$ \times$&$\checkmark$&$\mathtt{\sim}$&$ \times$&$ \times$&$ \times$&$ \times$&$\times$&$\checkmark$&$\times$&$\checkmark$&$ \checkmark$\\\cline{4-32}

&&&\multicolumn{2}{l}{Data Cleaning}&$ \times$&$ \mathtt{\sim}$&$ \times$&$ \times$&$ \times$&$ \times$&$ \times$&$ \times$&$ \times$&$\mathtt{\sim}$&$ \times$&$ \times$&$\checkmark$&$\checkmark$&$\mathtt{\sim}$&$ \times$&$\checkmark$&$ \times$&$ \times$&$ \times$&$ \times$&$ \times$&$\times$&$\checkmark$&$\times$&$\checkmark$&$ \checkmark$\\\cline{4-32}

&&&\multicolumn{2}{l}{Data Storage $\&$ Aggregation}&$ \times$&$ \times$&$ \times$&$ \times$&$ \times$&$ \times$&$ \times$&$ \times$&$ \mathtt{\sim}$&$ \times$&$ \times$&$ \times$&$\mathtt{\sim}$&$ \mathtt{\sim}$&$\mathtt{\sim}$&$ \times$&$\checkmark$&$ \times$&$ \times$&$ \times$&$ \times$&$ \times$&$\times$&$\checkmark$&$\times$&$\checkmark$&$ \checkmark$\\\cline{4-32}

&&&\multicolumn{2}{l}{Data Compression}&$ \times$&$ \times$&$ \times$&$ \times$&$\times$&$ \times$&$ \times$&$ \times$&$ \times$&$ \times$&$ \times$&$ \times$&$\checkmark$&$\checkmark$&$\mathtt{\sim}$&$ \times$&$\checkmark$&$ \times$&$ \times$&$ \times$&$ \times$&$ \times$&$\times$&$\checkmark$&$\times$&$\checkmark$&$ \checkmark$\\\cline{3-32}

&&\multicolumn{3}{l}{Anomaly Detection$\&$
Mitigation}&$ \times$&$ \times$&$ \times$&$ \times$&$ \times$&$\times$&$\mathtt{\sim}$&$ \times$&$\mathtt{\sim}$&$\mathtt{\sim}$&$\mathtt{\sim}$&$\times$&$\times$&$\mathtt{\sim}$&$\times$&$\times$&$\times$&$\checkmark$&$\mathtt{\sim}$&$\mathtt{\sim}$&$\mathtt{\sim}$&$\mathtt{\sim}$&$\times$&$ \checkmark$&$\mathtt{\sim}$&$\mathtt{\sim}$&$ \checkmark$\\\cline{3-32}

&&\multicolumn{3}{l}{Mixed Traffic Transportation Networks}&$ \times$&$ \times$&$ \times$&$ \times$&$ \times$&$\times$&$\checkmark$&$ \times$&$ \times$&$\mathtt{\sim}$&$ \times$&$\times$&$\times$&$\times$&$\times$&$\times$&$\times$&$\times$&$\times$&$\times$&$\mathtt{\sim}$&$\mathtt{\sim}$&$\times$&$\times$&$ \times$&$ \times$&$ \checkmark$\\\cline{3-32}

&&\multicolumn{3}{l}{Limited Data Accessibility}&$ \times$&$\mathtt{\sim}$&$\mathtt{\sim}$&$ \times$&$ \times$&$\mathtt{\sim}$&$\checkmark$&$ \times$&$ \checkmark$&$ \times$&$ \times$&$\times$&$\mathtt{\sim}$&$\mathtt{\sim}$&$\checkmark$&$\mathtt{\sim}$&$\checkmark$&$\times$&$\mathtt{\sim}$&$\times$&$\mathtt{\sim}$&$ \checkmark$&$\times$&$\mathtt{\sim}$&$\checkmark$&$\mathtt{\sim}$&$ \checkmark$\\\cline{3-32}

&&\multicolumn{3}{l}{Benchmarking Traffic Prediction Datasources}&$ \times$&$ \times$&$ \times$&$ \times$&$ \times$&$\times$&$\checkmark$&$ \times$&$ \checkmark$&$ \times$&$ \times$&$\times$&$\mathtt{\sim}$&$\mathtt{\sim}$&$\checkmark$&$ \times$&$\times$&$\times$&$\times$&$\times$&$\mathtt{\sim}$&$ \checkmark$&$\times$&$\mathtt{\sim}$&$\checkmark$&$\checkmark$&$ \checkmark$\\\cline{3-32}

&&\multicolumn{3}{l}{External Data Restrictions}&$ \times$&$\checkmark$&$\mathtt{\sim}$&$ \times$&$\mathtt{\sim}$&$\checkmark$&$\checkmark$&$ \times$&$ \checkmark$&$ \checkmark$&$ \checkmark$&$\times$&$\mathtt{\sim}$&$\mathtt{\sim}$&$\checkmark$&$\checkmark$&$\checkmark$&$\times$&$\times$&$\times$&$\checkmark$&$ \checkmark$&$\times$&$\mathtt{\sim}$&$\checkmark$&$\checkmark$&$ \checkmark$\\\cline{3-32}

&&\multicolumn{3}{l}{Multi-source Data}&$ \times$&$ \times$&$ \times$&$ \times$&$ \times$&$\checkmark$&$\checkmark$&$ \times$&$ \checkmark$&$ \checkmark$&$\mathtt{\sim}$&$\times$&$\mathtt{\sim}$&$\times$&$\times$&$\checkmark$&$\mathtt{\sim}$&$\times$&$\mathtt{\sim}$&$\times$&$\checkmark$&$ \checkmark$&$\times$&$\checkmark$&$\checkmark$&$\checkmark$&$ \checkmark$\\\cline{3-32}

&&\multicolumn{3}{l}{Data Collection Area}&$ \checkmark$&$\checkmark$&$ \times$&$ \times$&$ \times$&$\mathtt{\sim}$&$ \checkmark$&$ \checkmark$&$\mathtt{\sim}$&$ \times$&$\mathtt{\sim}$&$\times$&$\mathtt{\sim}$&$\checkmark$&$\times$&$\times$&$\times$&$\times$&$\times$&$\times$&$\mathtt{\sim}$&$\times$&$\times$&$\times$&$\mathtt{\sim}$&$\checkmark$&$ \checkmark$\\\hline

&\multirow{8}{*}{\shortstack[l]{Method-\\ology}}&\multirow{5}{*}{Type}&\multicolumn{2}{l}{Statistical}&$ \checkmark$&$ \checkmark$&$ \checkmark$&$ \times$&$\mathtt{\sim}$&$ \checkmark$&$ \checkmark$&$ \checkmark$&$ \mathtt{\sim}$&$ \checkmark$&$\times$&$ \times$&$\checkmark$&$ \times$&$ \times$&$ \times$&$\times$&$ \times$&$ \times$&$\checkmark$&$ \times$&$ \checkmark$&$\times$&$ \checkmark$&$\checkmark$&$\mathtt{\sim}$&$ \checkmark$\\\cline{4-32}

&&&\multicolumn{2}{l}{Traditional ML}&$ \checkmark$&$ \checkmark$&$ \checkmark$&$ \times$&$\mathtt{\sim}$&$ \checkmark$&$ \checkmark$&$ \checkmark$&$ \mathtt{\sim}$&$ \checkmark$&$\mathtt{\sim}$&$ \times$&$ \checkmark$&$ \checkmark$&$ \mathtt{\sim}$&$ \times$&$\checkmark$&$ \times$&$\checkmark$&$\checkmark$&$ \mathtt{\sim}$&$ \checkmark$&$ \mathtt{\sim}$&$ \checkmark$&$ \checkmark$&$\mathtt{\sim}$&$ \checkmark$\\\cline{4-32}

&&\multirow{2}{*}{}&\multicolumn{2}{l}{DL}&$ \times$&$ \times$&$ \times$&$ \mathtt{\sim}$&$\mathtt{\sim}$&$\mathtt{\sim}$&$\mathtt{\sim}$&$ \times$&$\checkmark$&$ \checkmark$&$ \checkmark$&$ \checkmark$&$ \checkmark$&$ \checkmark$&$ \checkmark$&$ \checkmark$&$\checkmark$&$\checkmark$&$\checkmark$&$\checkmark$&$\checkmark$&$ \checkmark$&$ \mathtt{\sim}$&$ \checkmark$&$ \checkmark$&$\mathtt{\sim}$&$ \checkmark$\\\cline{4-32}

&&&\multicolumn{2}{l}{Limitations}&$ \times$&$ \times$&$ \mathtt{\sim}$&$ \mathtt{\sim}$&$\mathtt{\sim}$&$\mathtt{\sim}$&$\times$&$ \times$&$ \mathtt{\sim}$&$ \mathtt{\sim}$&$ \checkmark$&$ \mathtt{\sim}$&$ \checkmark$&$ \checkmark$&$ \checkmark$&$ \checkmark$&$\checkmark$&$\checkmark$&$\checkmark$&$\checkmark$&$\checkmark$&$ \checkmark$&$ \mathtt{\sim}$&$ \checkmark$&$ \checkmark$&$\mathtt{\sim}$&$ \checkmark$\\\cline{3-32}

&&\multicolumn{3}{l}{Evaluation Matrices}&$ \times$&$ \times$&$\mathtt{\sim}$&$ \times$&$\mathtt{\sim}$&$ \times$&$ \mathtt{\sim}$&$\times$&$\checkmark$&$ \mathtt{\sim}$&$ \mathtt{\sim}$&$\times$&$\times$&$ \mathtt{\sim}$&$ \mathtt{\sim}$&$\times$&$ \checkmark$&$\times$&$\checkmark$&$\checkmark$&$\times$&$ \checkmark$&$\times$&$\times$&$\times$&$\checkmark$&$ \checkmark$\\\cline{3-32}

&&\multicolumn{3}{l}{Long-term Traffic Prediction}&$ \times$&$ \times$&$\mathtt{\sim}$&$\mathtt{\sim}$&$\checkmark$&$\times$&$ \checkmark$&$ \times$&$ \checkmark$&$\times$&$\times$&$\times$&$\times$&$\times$&$\times$&$\times$&$\times$&$\times$&$\times$&$\times$&$ \times$&$ \checkmark$&$\times$&$\times$&$\mathtt{\sim}$&$\checkmark$&$ \checkmark$\\\cline{3-32}

&&\multicolumn{3}{l}{Real-time Prediction}&$ \times$&$ \times$&$ \times$&$ \times$&$ \times$&$ \times$&$ \times$&$ \times$&$ \times$&$ \times$&$ \times$&$ \times$&$ \times$&$ \times$&$ \checkmark$&$ \times$&$\times$&$\times$&$\times$&$\checkmark$&$ \times$&$ \checkmark$&$\mathtt{\sim}$&$\mathtt{\sim}$&$\mathtt{\sim}$&$ \checkmark$&$ \checkmark$\\\cline{3-32}

&&\multicolumn{3}{l}{Model Selection, Improve $\&$ Combine and Testing }&$\mathtt{\sim}$&$\checkmark$&$\mathtt{\sim}$&$\mathtt{\sim}$&$\mathtt{\sim}$&$ \checkmark$&$ \checkmark$&$\mathtt{\sim}$&$ \checkmark$&$ \checkmark$&$ \checkmark$&$ \checkmark$&$ \checkmark$&$ \checkmark$&$ \checkmark$&$ \checkmark$&$\times$&$ \checkmark$&$ \checkmark$&$ \checkmark$&$ \checkmark$&$ \checkmark$&$\checkmark$&$\checkmark$&$\checkmark$&$\checkmark$&$ \checkmark$\\\hline

&\multicolumn{4}{l}{Applications}&$ \times$&$ \times$&$\mathtt{\sim}$&$\mathtt{\sim}$&$\mathtt{\sim}$&$\times$&$\times$&$\times$&$\mathtt{\sim}$&$\checkmark$&$\times$&$\times$&$\times$&$ \mathtt{\sim}$&$\times$&$ \mathtt{\sim}$&$\times$&$\times$&$\times$&$\times$&$\checkmark$&$\times$&$\times$&$\checkmark$&$\mathtt{\sim}$&$\checkmark$&$ \checkmark$\\\hline

&\multicolumn{4}{l}{Prediction States}&$\mathtt{\sim}$&$\mathtt{\sim}$&$\mathtt{\sim}$&$\mathtt{\sim}$&$ \times$&$ \mathtt{\sim}$&$\times$&$ \mathtt{\sim}$&$\mathtt{\sim}$&$\mathtt{\sim}$&$\mathtt{\sim}$&$\mathtt{\sim}$&$\mathtt{\sim}$&$ \mathtt{\sim}$&$ \mathtt{\sim}$&$ \checkmark$&$\mathtt{\sim}$&$\mathtt{\sim}$&$\mathtt{\sim}$&$\mathtt{\sim}$&$\checkmark$&$\checkmark$&$\mathtt{\sim}$&$\checkmark$&$\mathtt{\sim}$&$\mathtt{\sim}$&$ \checkmark$\\\hline

\multirow{14}{*}&\multicolumn{4}{l}{Emergence of AV$\&$CAV}&$ \times$&$ \times$&$ \times$&$\mathtt{\sim}$&$ \times$&$\times$&$\mathtt{\sim}$&$ \times$&$\checkmark$&$\mathtt{\sim}$&$\times$&$\times$&$\times$&$ \times$&$ \times$&$\times$&$\times$&$\times$&$\times$&$\times$&$\times$&$\times$&$\times$&$\times$&$\times$&$\checkmark$&$ \checkmark$\\\hline

&\multicolumn{4}{l}{FL Approach}&$ \times$&$ \times$&$ \times$&$ \times$&$ \times$&$\times$&$\times$&$ \times$&$ \times$&$ \times$&$\times$&$\times$&$\times$&$ \times$&$ \times$&$\times$&$\times$&$\times$&$\times$&$\times$&$\times$&$\times$&$\times$&$\times$&$ \times$&$\mathtt{\sim}$&$ \checkmark$\\\hline

&\multicolumn{4}{l}{Blockchain-enabled Traffic Prediction}&$ \times$&$ \times$&$ \times$&$ \times$&$ \times$&$\times$&$\times$&$ \times$&$ \times$&$ \times$&$\times$&$\times$&$\times$&$ \times$&$ \times$&$\times$&$\times$&$\times$&$\times$&$\times$&$\times$&$\times$&$\times$&$\times$&$ \times$&$\mathtt{\sim}$&$ \checkmark$\\\hline

&\multicolumn{4}{l}{Actionability}&$ \times$&$ \times$&$ \times$&$ \times$&$ \times$&$ \times$&$ \times$&$ \times$&$ \times$&$ \times$&$ \times$&$ \times$&$ \times$&$ \times$&$ \times$&$ \times$&$\times$&$\times$&$\times$&$ \times$&$ \times$&$ \times$&$ \times$&$ \times$&$\checkmark$&$\checkmark$&$ \times$\\\hline

\hline
\end{tabular}

\begin{tablenotes}
      \tiny
      \item $ \checkmark$ indicates topic is covered, $\times$ indicates that topic is not covered and $\mathtt{\sim}$ indicates that topic is partially covered.

\end{tablenotes}
  \end{threeparttable}
\end{table} 
\end{landscape}

\section{Traffic Data}
\label{Sec3}

This section extensively outlines and categorizes the varying data types and existing public data resources available for traffic prediction research. We also discuss data resolution and the different approaches in the literature for preprocessing the raw data into valuable and actionable information.

\subsection{Data Types and Resources}
The first step in accurate traffic prediction is acquiring high-quality data. Generally, the data collection operation is conducted by various stakeholders, including government organizations, municipalities, researchers, and commercial companies, on different time scales, utilizing various techniques. Government agencies and municipalities often measure road traffic for planning future roads, operation and maintenance, and analysis of accident risks and environmental impacts. The timescale of interest is often days, months, or years, but measurements are sometimes collected and used in real-time for traffic control and incident detection. On the other hand, commercial companies such as Google, Waze\footnote{\url{https://www.waze.com/}}, Here\footnote{\url{https://www.here.com/platform/traffic-solutions/real-time-traffic-information}}, TomTom, and INRIX, collect and provide traffic information to drivers (and in some cases also to businesses, urban planners, and road authorities). However, some of the collected data is not accessible since it is maintained privately. That being said, many publicly available datasets are of interest to researchers. We direct the reader to a recent survey conducted by \citet{jiang2022big} for a list of open datasets. Specifically, the data types used within the literature can be categorized into two areas based on their various characteristics and associated resources: (1) the spatio-temporal sequence data; and (2) the external data. 

\subsubsection{Spatio-temporal Sequence Data}
The most commonly used spatio-temporal, open-access datasets within the literature are summarized based on their characteristics in Table \ref{openacc}. In Table \ref{table:datasets1ref}, we outline some recent research works utilizing these datasets. More specifically, we summarize the available spatio-temporal sequence data commonly used in traffic forecasting into the following categories:

\begin{itemize}
\item \textbf{Fixed Position Sensor Data.} A primary method for collecting traffic data leverages fixed-position sensors, which detect nearby vehicles at a specific point on the roadway. Inductive-loop detector, camera, radar, and light detecting and ranging (LIDAR) are standard sensors used to collect data. These sensors are usually limited in spatial coverage and measure traffic volume, speed, occupancy, and other traffic flow parameters. However, the main reason for this type of data's popularity is its availability and compatibility with deep neural network models; this data usually does not require significant transformation steps and can be used in its collected format. For example, PEMS\footnote{\url{https://pems.dot.ca.gov/}} and its subsets (PeMS-BAY\footnote{\url{https://github.com/liyaguang/DCRNN}}, PeMSD3, PeMSD4, PeMSD7 and PeMSD8) are widely considered as the most used datasets within the literature. PEMS contains data from all major metropolitan areas of California from 2001 to 2019. Another frequently used dataset is METR-LA\footnote{\url{https://github.com/liyaguang/DCRNN}}, which contains traffic speed and volume data collected from the highways of Los Angeles County in 2012 from March 1st to June 30th, aggregated in 5-minute intervals. Additionally, LOOP\footnote{\url{https://github.com/zhiyongc/Seattle-Loop-Data}} and Los-loop\footnote{\url{https://github.com/lehaifeng/T-GCN/tree/master/data}} are the other popular traffic sensor data that gathered data from Seattle and Los Angeles freeways, respectively. Furthermore, Electronic Toll Collection (ETC) systems, with their multitude of fixed-position sensors, have also been leveraged to collect traffic data in some studies \citep{chen2020long}. 

\item \textbf{Trajectory Data.} This type of data is generally collected using probe vehicle techniques designed for dynamic real-time data collection. In addition, with the emergence of IoT and AVs, mobile sensors can provide more detailed information \citep{miglani2019deep, lippi2013short}. These systems, known as moving sensors, can be utilized for travel time data collection and assist in real-time applications, including routine monitoring of traffic operations, anomaly event detection, and route guidance. 
Based on the literature, the ITS probe vehicle data collection systems, such as Automatic Vehicle Identification (AVI) systems \citep{wang2019traffic}, Cellular Geo-location systems, and Global Positioning Systems (GPS) \citep{li2020autost} help improve traffic prediction by quantifying the exact trajectory and motion patterns of nearby vehicles, bicycles, and pedestrians. Additionally, they can help identify road segment connections where limited or no fixed sensors are deployed \citep{nagy2018survey, hu2021urban}. Floating Car Data (FCD) is another subset of trajectory which can be generated by smartphones or vehicles equipped with GNSS receivers \citep{elleuch2020neural}. When the data is collected by cellular phones using the cellular network, it is generally referred to as Cellular Floating Car Data (CFCD). On the other hand, the data generated by GNSS receivers are classified as GNSS probe data. Notably, most GNSS probe data is collected by fleet vehicles (e.g., taxi services, public transport services, private fleet companies). 

Notably, there are some operational differences between moving and fixed sensors. While stationary sensors capture spatio-temporal traffic data without any confusion regarding the location of vehicles, due to their fixed detection range, these sensors can be less reliable, and operation may be interrupted dynamically. Thus, it is vital to consider the possibility of outage periods when working with large datasets collected from fixed sensor deployments. Meanwhile, the shared and privately-owned moving sensors do not incur general maintenance costs (e.g., weather damage, part failure, accidents, vandalism) associated with fixed sensors, providing a cost-benefit. However, while a single fixed sensor on the roadway can collect all the data, moving sensors require a high proportion of vehicles or pedestrians to collect and provide data to represent the approximate ground-truth flow and all the possible trajectories \citep{nagy2018survey}. Moreover, it becomes imperative to preprocess the GPS trajectory data since measurement or interpolation errors related to sampling rates can cause uncertainties.

The available public trajectory data, which are provided by several probe-data vendors, such as INRIX, HERE,
TomTom, NAVTEQ\footnote{\url{https://www.here.com/navteq}} and TrafficCast\footnote{\url{https://www.iteris.com/TCI-redirect}}, can be divided into various categories. Some of them are recognized as Taxi Data, which contains many collected taxicab trajectories derived from GPS data. T-Drive\footnote{\url{https://www.microsoft.com/en-us/research/publication/t-drive-driving-directions-based-on-taxi-trajectories/}}, TaxiBJ  \footnote{\url{https://github.com/lucktroy/DeepST/tree/master/data/TaxiBJ}}, SZ-taxi\footnote{\url{https://github.com/lehaifeng/T-GCN}}, NYC taxi\footnote{\url{https://www1.nyc.gov/site/tlc/about/tlc-trip-record-data.page}} and TaxiCD\footnote{\url{https://js.dclab.run/v2/cmptDetail.html?id=175}}  are some of the well-known datasets used for traffic prediction. Ridehailing data gathered by some companies, such as Uber\footnote{\url{https://github.com/fivethirtyeight/uber-tlc-foil-response}} and Didi\footnote{\url{https://outreach.didichuxing.com/research/opendata/}}, represents another type of GPS data which contains ride-hailing trip requests with their travel time, pick-up, and drop-off locations. Thus, this type of data is often applied for traffic demand prediction problems. Furthermore, Bus Data (e.g., NYC BUS\footnote{\url{https://new.mta.info/system\_modernization/bus-network}}, CHI BUS\footnote{\url{http://www.gbis.go.kr/}}), Bike Data (e.g., CHI-Bike\footnote{\url{https://www.divvybikes.com/system-data}}, DCBike\footnote{\url{https://www.capitalbikeshare.com/system-data}}, NYC-Bike\footnote{\url{https://www.citibikenyc.com/system-data}}) and Subway Data (e.g. SHMetro\footnote{\url{https://github.com/ivechan/PVCGNZ}}, HZMetro\footnote{\url{https://github.com/ivechan/PVCGN}}) represent other types of trajectory data frequently used in the current traffic prediction literature. 

\item \textbf{Network Infrastructure Data.} Network infrastructure data (e.g., GIS data) is becoming increasingly available within open-access traffic datasets due to its importance in representing both spatial and temporal correlations \citep{liao2018deep,pan2019urban,peng2020spatial,james2021citywide}. More specifically, in applications such as map matching, network infrastructure data can be leveraged to map trajectory data sequences to actual positions on the roadway infrastructure, improving traffic prediction capabilities \citep{atif2020internet, james2021citywide}. This data includes information about the topology of roadway, subway, and bus networks. Much of the available network infrastructure data is open-access and can be obtained from government agencies (e.g., FHWA) or extracted from applications such as Open Street Map (OSM)\footnote{\url{https://www.openstreetmap.org/$\#$map=4/38.01/-95.84}}. As one of the most common open-access map databases, OSM provides information about road networks worldwide. Notably, this map data is collected by hand from volunteers performing systematic ground surveys using tools such as a handheld GPS unit, a notebook, a digital camera, or a voice recorder. In addition to OSM, A-map is a popular smartphone-based navigation app utilized in recent studies to collect network infrastructure data \citep{zhang2019multistep}.
    
\item \textbf{Trip Records Data}. The trip information, such as departures and arrivals times, dates, and locations, is necessary for traffic speed and demand prediction. Trip record data can be collected in many ways and from various sources, including taxis, ride-hailing services, e-scooters, bicycles, buses, and navigation systems. Another method for collecting trip record data is extracting it from existing datasets that may include trip data in forms other than the traditional origin-destination pairs. Much of this data can be automatically collected by the AFC (Automatic Fare Collection) system in subways and buses. In addition, trip record data can elucidate how drivers' characteristics influence their driving patterns, preferred route choices, and other driving-related decisions. When the collected information is labeled with additional driver-related attributes, such as driver ID, age, and gender, it is possible to improve prediction accuracy and provide personalized predictions \citep{zou2020estimation}.

\end{itemize}
 
 \begin{table}[!ht]
\tiny
\centering
\begin{threeparttable}
\caption{Commonly used and large-scale real-world datasets in traffic prediction}
\label{openacc}
  \begin{tabular}{llcll}
\hline
\multicolumn{2}{c}{\textbf{dataset}}&\textbf{Area}
&\textbf{Spatial coverage}&\textbf{Time}\\\hline
\multirow{9}{*}{\rotatebox{90}{PeMS}}&PeMS3&\multirow{9}{*}{California, USA}&\textendash   358 sensors&9/1/2018 $\sim$ 11/30/2018\\\cline{2-2}\cline{4-5}

&\multirow{1}{*}{PeMS4}&
&\multirow{1}{*}{\textendash   3848 sensors on 29 roads}&\multirow{1}{*}{1/1/2018 $\sim$2/28/2018}\\\cline{2-2}\cline{4-5}

&\multirow{1}{*}{PeMS7(M)}&
&\multirow{1}{*}{\textendash   228 stations}&\multirow{1}{*}{May and June, 2012 }\\\cline{2-2}\cline{4-5}

&PeMS7(L)&&\textendash   1026 stations&May and June, 2012\\\cline{2-2}\cline{4-5}

&\multirow{1}{*}{PeMS7}&&\multirow{1}{*}{\textendash   883
sensor stations}&\multirow{1}{*}{7/1/2016 $\sim$ 8/31/2016}\\\cline{2-2}\cline{4-5}

&\multirow{2}{*}{PeMS8}&
&\multirow{2}{*}{\textendash   1979 sensors on 8 roads}&7/1/2016 $\sim$ 8/31/2016\\
&&&&7/1/2016 $\sim$ 8/31/2016\\\cline{2-2}\cline{4-5}

&\multirow{2}{*}{PeMS-SF}&

&\textendash 440 days of traffic data &\multirow{2}{*}{1/1/2008 $\sim$ 3/30/2009}\\
&&&\textendash 963 sensors in the San Francisco&\\
\cline{2-2}\cline{4-5}

&PeMS-BAY&
&\textendash   325 sensors&1/1/2017 $\sim$ 6/30/2017\\\hline

\multicolumn{2}{c}{\multirow{1}{*}{METR-LA}}&\multirow{1}{*}{Los Angeles, USA}&\multirow{1}{*}{\textendash   207 sensors on the highways}&\multirow{1}{*}{3/1/2012 $\sim$ 6/30/2012}\\\hline

\multicolumn{2}{c}{\multirow{3}{*}{LOOP}}& 
\multirow{3}{*}{Seattle,USA}&
\textendash   323 sensor stations on&\multirow{3}{*}{2015} 
\\
& &
&four connected freeways: &
\\
&& &I-5, I-405, I-90 \& SR-520&\\\hline
\multicolumn{2}{c}{Los-loop}&Los Angeles, USA&\textendash   207 sensors on the highways&3/1/2012 $\sim$ 3/7/2012\\

\hline
\multicolumn{2}{c}{\multirow{3}{*}{Q-Traffic}}
&\multirow{3}{*}{Beijing, China}&
 \textendash   15073 road segments &\multirow{3}{*}{4/1/2017 $\sim$ 5/31/2017}
\\
&
&&\textendash  Approximately 73891 km (6th ,& Map)
\\
&
&&ring road) from the Baidu Map&\\\hline

\multicolumn{2}{c}{TOPIS}&Seoul, South Korea&\textendash  1153 sensors and over 70000 taxis&1/2014 $\sim$\\\hline

\multicolumn{2}{c}{ENG-HW}&Three cities,Britain&\textendash  Inter-city highways&2006 $\sim$ 2014\\
\hline

\multicolumn{2}{c}{WebTRIS}&England&\textendash  All motorways and "A" roads
&4/2015 $\sim$\\\hline

\multicolumn{2}{c}{HK}&Hong Kong, China&\textendash  Four regions in Hong Kong&12/28/2015 $\sim$\\\hline

\multicolumn{2}{c}{DRIVENet}&Seattle, USA&\textendash  323 stations, 85 miles&2011\\\hline

\multicolumn{2}{c}{Travel Time }&Shenzhen,
Suzhou, 
&\textendash Include city-level, district-level,
&\multirow{2}{*}{1/1/2018 $\sim$
12/31/2018}\\
\multicolumn{2}{c}{Index data}&Jinan, and Haikou
&and road-level&\\
\hline
\multirow{8}{*}{\rotatebox{90}{DiDi}}&DiDiChengdu&\multirow{2}{*}{Chengdu, China}
&\textendash   5476 geographical blocks&11/1/2016$\sim$11/30/2016\\\cline{2-2}\cline{5-5}
&DiDiTTIChengdu&
&(each block size = 1$km^2$)&2018\\\cline{2-3}\cline{4-5}
&DiDiXi’an&Xi’an, China&\textendash  Contain DiDi
Express and DiDi&10/2016$\sim$11/2016 2016.\\\cline{2-3}\cline{5-5}
&DiDiHaikou&Haikou, China&Premier drivers&5/1/2017 $\sim$ 10/31/2017\\\cline{2-3}\cline{4-5}
&\multirow{3}{*}{Didi Chuxing GAIA}& \multirow{3}{*}{JiNan $\&$ XiAn,}&\textendash Total sample number of the two&\multirow{4}{*}{10/2017 $\sim$ }  \\
&\multirow{3}{*}{Initiative}&\multirow{3}{*}{China}&datasets is 52286 each&\\
&&&\textendash 561 
road segments for JiNan \\
&&&\textendash 792
road segments for XiAn&\\\hline

\multicolumn{2}{c}{\multirow{2}{*}{Uber}}&\multirow{2}{*}{USA}&\multirow{2}{*}{\textendash  Over 18.8 mililon pick-up}&4/1/2014$\sim$9/30/2014\\
&&&&1/1/2015$\sim$6/30/2015\\\hline

\multicolumn{2}{c}{T-Drive}&Beijing, China&\textendash 10,357 taxis&2/1/2015 $\sim$ 6/2/2015\\\hline

\multicolumn{2}{c}{SHSpeed}&Shanghai,China&\textendash for 156 urban road segments
&4/1/2015$\sim$4/30/2015\\\hline

\multicolumn{2}{c}{\multirow{4}{*}{TaxiBJ}}&\multirow{4}{*}{Beijing, China}&\multirow{2}{*}{\textendash Contains inflow and outflow data}
&7/1/2013 $\sim$10/30/2013\\
&&
&\multirow{2}{*}{\textendash   Over 34000 taxicab GPS data} &3/1/2014 $\sim$
6/30/2014\\
&&&\multirow{2}{*}{\textendash  Include meteorology data}&3/1/2015 $\sim$6/30/2015\\
&&&& 11/1/2015 $\sim$ 4/10/2016\\\hline
\multicolumn{2}{c}{\multirow{1}{*}{SZ-taxi}}& \multirow{1}{*}{Shenzhen, China}&\multirow{1}{*}{\textendash   156 major roads of Luohu district}&\multirow{1}{*}{1/1/2015 $\sim$ 1/31/2015}\\\hline

\multicolumn{2}{c}{\multirow{6}{*}{TaxiCD}}&\multirow{6}{*}{Chengdu, China}&\textendash 1.4 billion GPS records from 14,864&\multirow{6}{*}{8/3/2014 $\sim$ 8/30/2014}\\
&&&taxis\\
&&&\textendash Consists of a taxi ID, latitude, \\ 
&&& longitude, an indicator of whether the \\ 
&&&taxi is occupied, and a timestamp\\\hline

\multicolumn{2}{c}{\multirow{2}{*}{NYC Taxi}}&\multirow{2}{*}{New York City}&\textendash   Including Bronx,Brooklyn, Manhattan&\multirow{2}{*}{2009 $\sim$ 2018}\\
&& &Queen, Staten Island&\\\hline

\multicolumn{2}{c}{\multirow{2}{*}{Porto}}&\multirow{2}{*}{Porto, Portugal}&\textendash  442 taxis, 420000 trajectories,&\multirow{2}{*}{7/1/2013$\sim$6/30/2014}\\
&&&16735m*14389m&\\\hline
\multicolumn{2}{c}{\multirow{2}{*}{BikeDC}}&\multirow{2}{*}{Washington, USA}&\textendash   Over 500 stations  across 6 jurisdictions&\multirow{2}{*}{20/9/2010 $\sim$
}\\
&&&\textendash   4300 bikes&
\\

\hline
\multicolumn{2}{c}{\multirow{2}{*}{NYC Bike}}&\multirow{2}{*}{New York City}&\textendash   800 stations
&\multirow{2}{*}{5/7/2013 $\sim$}\\
&&&\textendash 13000 bikes&\\\hline
\multicolumn{2}{c}{\multirow{2}{*}{CHI Bike}}&\multirow{2}{*}{Chicago, USA}&\textendash  580 stations across chicagoland,&\multirow{2}{*}{6/27/2013$\sim$}\\
&&&5800 bikes&\\\hline

\multicolumn{2}{c}{NYC Bus}&New York, USA&\textendash  All public buses&2014\\\hline

\multicolumn{2}{c}{\multirow{2}{*}{CHI Bus}}&\multirow{2}{*}{Chicago, USA}&\textendash  Buses on arterial streets in real time,&\multirow{2}{*}{8/2/2011$\sim$5/3/2018}\\
&&&\textendash 
1250 road segments covering 300 miles\\\hline

\multicolumn{2}{c}{Chengdu}&Chengdu, China&\textendash 14864 taxis, 9737557 trajectories&8/2014\\\hline
\multicolumn{2}{c}{Shanghai}&Shanghai, China&\textendash 1.5 thousand main roads&3/1/2015$\sim$ 4/31/2015\\\hline
\multicolumn{2}{c}{Beijing}&Beijing, China&\textendash  More than 2 million trajectories &3/1/2016$\sim$7/31/2016\\\hline

\multicolumn{2}{c}{\multirow{4}{*}{SHMetro}}&\multirow{4}{*}{Shanghai, China}&\textendash811.8 million
transaction records of&\multirow{4}{*}{7/1/2016$\sim$ 9/30/2016}\\
&&&metro\\
&&&\textendash288 metro stations and 958 physical&\\
&&&edges\\\hline
\multicolumn{2}{c}{HZMetro}&Hangzhou, China&\textendash 80 metro stations and 248 physical edges&1/2019 \\\hline

\end{tabular}

  \end{threeparttable}
\end{table}

 \begin{table}[!ht]
\tiny
\centering
\caption{Detailed classification of current traffic prediction surveys based on the spatio-temporal Sequence Data}
\label{table:datasets1ref}
  \begin{tabular}{clc}
  \hline
  \textbf{Data Category} &\textbf{Ref.}&\textbf{Dataset}\\
  \hline
\multirow{38}{*}{ Fixed Position} &\citet{song2020spatial}&PeMS3\\\cline{2-3}
\multirow{38}{*}{Sensor Data}&\citet{shi2020spatial,bai2020adaptive,guo2019attention}&\multirow{2}{*}{PeMS4}\\
&\citet{li2018diffusion, huang2020lsgcn,song2020spatial}\\\cline{2-3}
&\citet{yu2018spatio,zhang2021fastgnn}&\multirow{2}{*}{PeMS7(M)}\\
&\citet{zhang2020spatio,sen2019think}\\\cline{2-3}
&\citet{song2020spatial,li2018diffusion,huang2020lsgcn}&PeMS7\\\cline{2-3}
&\citet{shi2020spatial,bai2020adaptive,guo2019attention}&\multirow{2}{*}{PeMS8}\\
&\citet{huang2020lsgcn,song2020spatial}\\\cline{2-3}
&\citet{zhao2018parallel}&PeMS-SF\\\cline{2-3}
&\citet{wu2019graph,zheng2020gman,shang2020end}&\multirow{10}{*}{PeMS-BAY}\\
&\citet{sun2021modeling,zhang2020spatial,lu2020st}\\
&\citet{jin2021hetgat,oreshkin2021fc,he2020stnn}\\
&\citet{boukerche2020performance,cui2020graph}\\
&\cite{wang2020forecast,kong2020stgat,cai2020traffic}\\
&\cite{wu2020connecting,pan2020spatio,zhou2020reinforced}\\
&\citet{chen2020multi,fu2016using}\\
&\citet{wang2020traffic,huang2019diffusion}\\
&\citet{zhang2020spatio,park2020st}\\
&\citet{chen2019gated,li2018diffusion}\\\cline{2-3}
&\citet{li2018diffusion,wu2019graph}&\multirow{13}{*}{METR-LA}\\
&\citet{jin2021hetgat,shang2020end,he2020stnn}\\
&\citet{chen2019gated,chen2020multi,zhou2020reinforced}\\
&\citet{zhang2020spatio,zhang2020spatial,jiang2022graph}\\
&\citet{sun2021modeling,lu2020st}\\
&\citet{chen2020autoreservoir,boukerche2020performance}\\
&\citet{oreshkin2021fc,kong2020stgat}\\
&\citet{cai2020traffic,pan2020spatio,wang2020traffic}\\
&\citet{wu2020connecting,wang2020forecast,wu2020graph}\\
&\citet{yang2019relational, huang2019diffusion, park2020st}\\
&\citet{zhang2018gaan,pan2019urban,cui2020graph}\\
\cline{2-3}
&\cite{liao2018deep}&Q-Traffic\\\cline{2-3}
&\cite{shin2020incorporating}&TOPIS\\\cline{2-3}
&\cite{zhao2018layerwise,zhao2019layerwise}&ENG-HW\\\cline{2-3}
&\cite{he2018stann}&HK\\\cline{2-3}
&\cite{zhao2019enlstm,qu2019daily}&DRIVENet\\\cline{2-3}
&\cite{zhang2010comparison}&Travel Time Index data\\\hline
Ride-hailing&\cite{lin2019spatial,jiang2021deepcrowd}&DiDi\\\cline{2-3}
Data&\cite{zhao2020unifying,faghih2019predicting}&Uber\\\hline
\multirow{5}{*}{ Taxi Data}&\cite{pan2019urban}&T-Drive\\\cline{2-3}
&\cite{li2020autost,bai2019stg2seq}&TaxiBJ\\\cline{2-3}
&\cite{zhao2019t}&SZ-taxi\\\cline{2-3}
&\citet{ye2019co,wang2020effective}&NYC Taxi\\\cline{2-3}
&\cite{karimzadeh2021reinforcement}&Porto\\\hline
\multirow{2}{*}{ Bike Data}
&\citet{bai2019stg2seq,wang2020effective}&NYC Bike\\\cline{2-3}
&\cite{chai2018bike}&CHI Bike\\\hline
\multirow{2}{*}{ Bus Data}&\cite{wang2021public}&NYC Bus\\\cline{2-3}
&\cite{jang2018integration}&CHI Bus\\\hline
\multirow{2}{*}{ Subway Data}&\cite{liu2020physical}&SHMetro \\\cline{2-3}
&\cite{liu2020physical}&HZMetro \\\hline
Network &\citet{james2021citywide}&OSM\\\cline{2-3}
Infrastructure&\multirow{2}{*}{\citet{hou2021effect}}&\multirow{2}{*}{A-map}\\
Data&&\\\hline
\end{tabular}
\end{table}

\subsubsection{External Data}

In addition to spatio-temporal sequence data, the prediction models must also be adaptable and responsive to the dynamic traffic and road environment changes, making it necessary to consider external data. Specifically, external data refers to related factors that impact traffic dynamics and roadway conditions that are not directly present within the collected spatio-temporal sequence data (e.g., meteorological and social media data). Problematically, this data is sometimes difficult to collect in practice or is unavailable. The existing studies leveraging external data are presented as part of Table \ref{table:exter}. Below, we provide an outline of the most common types of external data and their importance in producing accurate traffic prediction models:

\begin{table}[t]
\tiny
\centering
\begin{threeparttable}
\caption{Applying external datasets in traffic prediction studies}
\label{table:exter}
  \begin{tabular}{llll}
\hline
\textbf{Ref.}&\textbf{External Data Type}&\textbf{Characteristics}&\textbf{Time}\\\hline

\multirow{8}{*}{\citet{yao2021twitter}}&\multirow{2}{*}{Event Traffic Data} &\textendash Use PennDOT RCRS dataset including 2,696 &\multirow{2}{*}{2014}\\
&& traffic incidents on Pittsburg&\\\cline{2-4}

&\multirow{2}{*}{Meteorological data }&\textendash Use Weather underground dataset &\multirow{2}{*}{NA}\\
&&collected in Pittsburgh International Airport&\\\cline{2-4}
&\multirow{4}{*}{Social  Media  Data}&\textendash Use twitter dataset contains 1,782,636 tweets&\multirow{4}{*}{1/23/2014$\sim$12/31/2014}\\
&&\textendash Include date/time, text, user ID, language, \\
&&latitude and longitude (if available), user \\
&&profile location, etc.\\\hline

 \multirow{2}{*}{\citet{liu2019contextualized}}&\multirow{2}{*}{Meteorological Data } &\textendash Use Wunderground dataset collected & \multirow{2}{*}{NA}\\
 &&from Central Park and  Manhattan Stations&\\\hline
 
 \multirow{2}{*}{\citet{liu2020predicting}}&Meteorological data& \textendash Collect in  NYC, USA&1/1/2016$\sim$6/30/2016\\\cline{2-4}
 &Event Data& \textendash contains 22
weekdays, 8 weekends and holidays&1/11/2016$\sim$30/11/2016\\\hline

  \multirow{2}{*}{\citet{yao2018deep}}&Event Traffic Data&\textendash Collect from Didi Chuxing dataset &2/1/2017$\sim$3/26/2017\\\cline{2-4}
 &Meteorological Data&&\\
 \hline
 
 \citet{li2021hybrid}&Event Traffic Data&\textendash Collect in England&1/7/2018$\sim$
28/1/2020\\\hline

 \multirow{14}{*}{\citet{taghipour2020dynamic}}&\multirow{6}{*}{Meteorological Data}&\textendash Include weather conditions provided by the  &\multirow{14}{*}{4/2017$\sim$9/2017}\\
 &&National Weather Service (NWS) for 94 different&\\
 && states in US&\\
 &&\textendash Include sun glare data based on sun’s angle &\\
&&in relation to cars, solar elevation, solar azimuth,
&\\
&&driving direction, and slope&\\\cline{2-3}
&\multirow{6}{*}{Event Traffic Data}&\textendash Contain Incident, stadium events, parades, and &\\
&& road races data collected by Illinois Department &\\
&&of Transportation (IDOT)&\\
&&\textendash Include time and geographical locations of the  \\
&&accidents and events, severity of accident and type&\\
&& of events&\\\cline{2-3}
&\multirow{2}{*}{Work Zone Data}&\textendash Use Chicago highways network data collected &\\
&&by Illinois Department of Transportation (IDOT)&\\\hline

 \multirow{3}{*}{\citet{chen2018detecting}}& \multirow{3}{*}{Social Media Data}&\textendash Collect nearly forty thousand traffic relevent & \multirow{3}{*}{NA}\\
&&microblogs on china by searching Sina Weibo &\\
&&with key words&\\\hline

\multirow{2}{*}{\citet{lin2017road}}&\multirow{2}{*}{Social Media Data}&\textendash Collect 10000 tweets via the Twitter REST&\multirow{2}{*}{6/1/2013$\sim$3/31/2014}\\
&& search API from Washington D.C. \& Philadelphia&\\\hline

\multirow{3}{*}{\citet{zhang2018user}}&\multirow{3}{*}{Urban System Data}&\textendash Household-level electricity usage data from 322  &\multirow{3}{*}{2014}\\
&&household in Austin, Texas&\\
&&\textendash Collect from midnight to 6 am for 251 weekdays&\\\hline

\multirow{2}{*}{\cite{macioszek2021extracting}}& Disease  Spread  and
&\multirow{2}{*}{\textendash COVID19 cases reports from Poland } &\multirow{2}{*}{2019$\sim$2020}\\
& Pandemic  Data&
\\
\hline
\end{tabular}

  \end{threeparttable}
\end{table}

\begin{itemize}
\item \textbf{Event Traffic Data.} This type of data represents unexpected and non-recurrent traffic events on roads, such as accidents, sports events, and activities that may affect traffic flow and cause congestion. This data can be found in some reports, such as traffic accident reports. Notably, abnormal event data can also be found in some collected traffic datasets. Still, it may be difficult to distinguish it from other events and recurrent congestion without referencing other external data (e.g., accident). Event traffic data can also include trajectory data that is considered an anomaly because it appears with low frequency from an origin point to a destination point. Generally, anomaly trajectory data would be removed from the dataset during the cleaning process or repaired using various techniques \citep{liu2021experience} including activity sequence occurrence probability \citep{sani2019repairing}. However, this type of data can be helpful for specific applications, such as for detecting abnormal criminal behaviors in taxi drivers \citep{shen2015outlier, wang2018detecting}.
    
\item \textbf{Work Zone Data.} Construction activities on the roadways can disrupt expected traffic flows and create challenges for predicting traffic accurately. For example, work zones generally result in modifications to the existing roadway structure, such as reducing or eliminating lanes, shifting lanes, removing shoulders, using temporary signals, and reducing speed limits. As a result, additional traffic congestion becomes unavoidable in many cases, and accurate traffic prediction during these events is critical to mitigating the negative impacts.

\item \textbf{Meteorological Data.} 
Broadly, meteorological data includes multiple weather factors such as temperature and humidity, wind speed and strength, precipitation (e.g., rain, snow, hail), and severe weather (e.g., storm, tornado). For example, meteorological factors, such as sun glare, can negatively impact roadway congestion by obstructing the perception ability of both drivers and sensors, leading to roadway accidents that result in unexpected congestion \citep{taghipour2020dynamic}. The impacts of meteorological events on traffic flow and roadway conditions (e.g., dry, wet, frost, icy) have become increasingly important in recent years with the advent of autonomous vehicles, which must have the appropriate sensors and configurations to develop safe driving strategies during periods of inclement weather \citep{huang2021data}.

\item \textbf{Social Media Data}. Online social media data (e.g., Tweets), also known as crowdsourced data, provide rich information for short- and long-term traffic prediction. Facebook, Twitter, and Snapchat are some of the online platforms that have seen remarkable usage in the last few years for communication, news reporting, and advertising events \citep{essien2020deep}. Since these social media platforms provide real-time data retrieval using application programming interfaces (APIs), many researchers have leveraged social media data to understand better the relationships between people, their decisions, and the urban transportation systems \citep{ni2014using,yao2021twitter}. However, while social media data is generally available free of charge, it must be mined and can be unreliable due to human factors \citep{cvetek2021survey}.

\item \textbf{Disease Spread and Pandemic Data.} The worldwide COVID-19 pandemic has attracted much traffic prediction attention due to drastic changes in societal movement patterns. Pandemic events can significantly impact the prediction performance of existing models. More research is needed to investigate and quantify the effects of a non-recurrent worldwide pandemic on short-term and long-term traffic prediction \citep{macioszek2021extracting}. Notably, public health agencies generally employ Points of Dispensing (PODs) during a large-scale biological emergency event for distributing medical countermeasures to affected populations. While the public health community has studied the clinical operations aspect of PODs, little attention has been paid to how the transportation system can support POD access and how the choice of POD location impacts roadway traffic. For example, to accurately model the traffic impacts of PODs operations, it is necessary to consider expected demand concerning the parking capacity of a POD in conjunction with the nearby road traffic capacity at any given time.
    
\item \textbf{Urban Systems Data.} There may be a strong correlation between utility systems usage and transportation patterns within the urban environment. For example, high electricity usage in a neighborhood may indicate that most residents are at home and not utilizing the transportation network. Thus, the spatio-temporal usage patterns of a community can be identified by connecting and analyzing user demand data from urban systems, including transportation, energy, water, and building infrastructure. Despite this intuition, little research has been done in this area. However, a recent study exploring the correlation between electricity usage and roadway congestion revealed that electricity usage data could be leveraged to make reliable predictions regarding the beginning of peak period highway congestion, highlighting the promise of considering urban systems data for improving traffic prediction performance \citep{zhang2018user}.
        
\end{itemize}

\subsection{Data Resolution}
For training traffic prediction models, data resolution is an important consideration. Presently, data collection technologies allow acquiring traffic data at a variety of resolutions to match the needs for traffic management and control applications. Specifically, defining the appropriate data resolution is a critical issue, especially in data-driven algorithms, because it affects the quality of spatio-temporal dependencies, which are learnable from the underlying data. This section outlines and describes the resolution of two types of data commonly used in traffic forecasting: temporal and spatial data resolutions.

\subsubsection{Temporal Resolution}
Data resolution is strongly related to the forecasting horizon and step size. The horizon describes the forecasting time window (e.g., one day, one week, one month). At the same time, the step size denotes the frequency of forecasts within the horizon (e.g., every five minutes, every hour, every day). For example, a prediction model may predict daily traffic for an entire month. In other words, determining the most suitable forecasting interval (step) relates to the type of ITS application for which the algorithms are to be integrated. On the one hand, a significant prediction time horizon is preferred to provide sufficient time for control and reactions. On the other hand, a small prediction error is desirable for accurately informing the response of both TMCs and individual users. Notably, there exists a trade-off between the time horizon and prediction accuracy: a model trained to predict traffic over a small time horizon with a short step size may produce better accuracy for fine-grained prediction, but performance will degrade as the time horizon or step size increases.

Reviewing the literature reveals that most prior works on traffic prediction aggregate data over a specific time interval, generally no less than 5-minutes. However, this interval may not serve well for informing dynamic operational decisions (e.g., intelligent signal timing systems) that require traffic information in real-time at much shorter intervals. Regarding higher-resolution data, the increasing development of IoT technologies, in conjunction with improved data processing and storage techniques, enables high resolution data to be collected and stored by many traffic systems. Thus, it is possible to leverage the information obtained from high resolution data to predict traffic more effectively. Nevertheless, accurate traffic prediction based on high resolution data is still challenging because this type of data generally exhibits strong fluctuations. This lack of ‘balance’ in high resolution data presents a challenge to forecasting problems (particularly when applying traditional variants of time series techniques): sparsely represented data classes tend to be overlooked by conventional prediction models, especially when the imbalance is pronounced. In essence, the lower frequency labels are treated as an outlier \citep{li2020short} and excluded from the data before model training and evaluation.

\subsubsection{Spatial Resolution}
Within the literature, spatial data resolution generally ranges from a few hundred meters to several kilometers. Different data sources can have varying spatial resolutions. For example, the spatial resolution of FCDs is generally high (data is collected on average every 100 meters) compared to fixed-position detector data with a typical spacing of 1-3 km between detector positions (data collection points). Due to the high resolution, FCD can be plotted directly to illustrate the traffic dynamics. At the same time, the fixed-position detector data must be interpolated to reconstruct the spatio-temporal network dynamics during post-processing. Thus, FCD data is generally more helpful in elucidating spatial correlations within the data because it encompasses information about the entire stretch of a roadway link in contrast to isolated and fixed-position detectors \citep{kessler2018comparing}. Moreover, FCD can provide traffic information in areas of the network where no sensors are deployed.

While the level of spatial resolution is customized in most traffic prediction studies, some datasets introduced specific geographic units for spatial resolution. For example, INRIX architecture utilizes TMC-Segments and XD-Segments as the basis for defining road sections to report speed and incident data. TMC location codes were initially conceived of as points on the road network, typically assigned at significant decision points, interchanges, or intersections to describe locations of traffic incidents (e.g., accidents, construction, traffic slowdowns) in a specific format independent of map vendor. The INRIX XD Segments are similar to TMC Segments in that they delineate a particular section of roadway; however, they address some of the limitations of TMC Segments, such as limited road coverage and segment overlapping and gapping, as well as segment resolution. Besides, XD Segments are defined and maintained solely by INRIX. Therefore, INRIX can create XD Segments for sections of roads or new roads that do not yet have defined TMC location codes or TMC Segments. XD Segments also have a maximum length of 1.5 miles, providing a more consistent, unambiguous, and granular definition of road segments compared to TMC Segments, which vary significantly in length depending on the distance between neighboring TMC location codes \citep{INRIX, FHWA2019, ahsani2019quantitative}.

\subsection{Data Preprocessing}
\label{kms}
To effectively predict traffic and solve problems, collected raw data needs to be preprocessed to remove the irrelevant, redundant, noisy, and unreliable data, thus preventing misleading results while simultaneously producing more accurate insights. In general, data preprocessing is considered a preliminary step for enhancing raw data quality and employs varying techniques such as map-matching, data cleaning, data storage and aggregation, and data compression. In this subsection, we categorize and discuss the different preprocessing methods within the existing literature. 

\subsubsection{Map-matching} One of the most common preprocessing methods is Map-matching. This process assigns the latitude/longitude coordinates within the data to positions on the geographical road networks. Notably, it becomes necessary to apply practical map-matching algorithms to the raw positional data to produce meaningful results, and as such, many studies have been conducted on this problem. While  \citet{quddus2007current,wei2013map,zheng2015trajectory} introduced various map matching categorizations, from a technical perspective, \citet{chao2020survey} proposed four new classifications of the existing methods: similarity models, state transition models, candidate-evolving models, and scoring models. Later, \citet{yuan2021survey} provided a different categorization, arranging the methods into five groups based on the geometric information, topological structure of the road networks, and global optimal matching features of the data: Point-Distance, Path-Distance, Probability-Based, Model-Based, and Learning-Based.

Although Map-matching helps clarify and resolve uncertainties with traffic prediction data, inaccurate measurements and low sampling rates for collected data bring about quality issues, resulting in challenges for map matching approaches \citep{bhowmick2018trajectory,cui2021hidden}. For example, attempting to solve the problem of match uncertainty caused by varying map densities, \citet{quddus2015shortest} proposed a new shortest path and vehicle trajectory aided map-matching (stMM) algorithm to match low-frequency GPS data on a road map. In more recent work, \citet{singh2020genetic} applied a novel genetic algorithm to evaluate both sparse and dense GPS data for map matching. DMM, a fast map matching framework for cellular data that uses a recurrent neural network (RNN), has also been proposed to find the most likely road path given a sequence of cell towers \citep{singh2020genetic}. This method contrasts the commonly used Hidden Markov Models (HMM), which incur heavy computational overhead and scale poorly on large datasets.

\subsubsection{Data Cleaning} Accurate and efficient traffic prediction relies on clean spatio-temporal data. During data cleaning, the corrupted, poorly formatted, duplicate, or incomplete data is fixed or removed from the dataset. Notably, data cleaning techniques vary according to the underlying problem and type of data, but they can be broadly grouped into three categories:

\begin{itemize}
\item \textbf{Missing Data.} Hardware failures, software bugs, and human errors \citep{yuan2021survey} can cause missing data, which reduces the power of a model. Thus, more attention should be paid to precisely estimating and interpolating the missing data. Missing data challenges can be divided into two categories: short-period missing values and long-period missing values. Short-period missing values range include outages ranging from 1 second to about 5 minutes and are chiefly caused by unsteady equipment or a cluttered environment. Temporal smoothness is typically employed to deduce the missing values since the surrounding temporal information is rich in these situations. On the other hand, long-period missing values, which can last hours, or even days, are principally caused by system failures. Due to the lack of surrounding contextual information, it is typical for the mean or mode of the dataset (concerning the time of day) to be used as a substitution for long-term missing values. However, the prediction error is huge in this situation because of the scarcity of surrounding temporal information \citep{tian2018lstm}.

Various imputation methods have been proposed to solve this problem while retaining the total sample size by replacing missing values with substituted values rather than eliminating observations or variables with missing data. \citet{cheng2017two} introduced a two-step spatio-temporal missing data reconstruction (ST-2SMR) method to consider the missing patterns by leveraging a coarse-grained interpolation method. The authors then used a dynamic sliding window selection approach to identify which sample data was most relevant for fine-grained interpolation.

A more recent approach to data imputation was proposed by \citet{chan2021neural}, where Fuzzy Neural Network optimization is leveraged to perform robust missing data imputation in spatio-temporal and multi-dimensional datasets, effectively improving imputation accuracy. Notably, the imputation method was applied to the problem of dynamic vehicle rerouting, and the experimental results demonstrate the effectiveness of the imputed data. Meanwhile, in a different study by \citet{wang2018missing}, missing data is estimated by the Optimum Closed Cut (OCC) method, utilizing both the road topological information and the spatio-temporal correlation among road traffic. 

For the imputation of long-period consecutive missing values, \citet{ma2020transfer} proposed a new methodology, namely transferred long short-term memory-based iterative estimation (TLSTM-IE). In this method, an LSTM model first learns from a complete data sequence of similar distribution to the sequence with missing values. After that, the model employs transfer learning to estimate the missing values for the incomplete dataset iteratively. In this way, the missing values can be intelligently calculated based on the observations of similar time series, in contrast to naive imputation approaches, preserving the underlying long-term data trends. A year later, \citet{kwon2021multilayered} introduced another data imputation algorithm based on the M-LSTM architecture with parameter transfer in V2I networks. This algorithm can adopt an LSTM network to learn the representation of time series traffic data and proposes a multilayered LSTM (M-LSTM) network with parameter transfer for efficient imputation. Specifically, the parameters trained in one layer can be transferred to the following layers, reducing the training and imputing time compared to the prior approaches. Therefore, the proposed algorithm can ensure reliable input for ITS applications leveraging real-time sensor data, even when the sensor is offline.

Another missing data problem results when no historical trajectory data exists: commonly referred to as the cold start problem. Notably, it is challenging to establish predictors without historical data. For example, sensors may be continuously added and removed daily, making the cold start problem a practical consideration in growing transportation networks. In one of the earliest works, \citet{gao2012gscorr} proposed a geo-social correlation (gSCorr) model based on social information to alleviate the cold starts problem in location prediction models. More recently, \citet{tu2019fingerprint} proposed a transfer learning-based generative model for providing personalized location recommendations, which transfers user interest and location features from app usage data to provide location recommendations in the absence of historical data. This method builds a user-location matrix, user-app matrix, and location-app matrix from the data, enabling the model to learn the latent vectors of users, locations, and apps jointly. Meanwhile, \citet{gong2020potential} handled the cold-start challenge for passenger flow prediction by mining supplementary information from urban statistical data to guide the learning process. Lastly, \citet{magalhaes2021speed} proposed the global and cluster-based approaches, where models are trained on data coming from all the sensors in the network (or groups of similar sensors, in the cluster-based case) to build resilient predictive functions. However, among these two approaches, the global-based approach naturally solves the cold start problem because it relies on a single prediction function independent of specific sensors.
 
\item \textbf{Data outliers.} Outliers and erroneous data create severe problems for data analysis and can result from a myriad of problems, including human mistakes, rare exceptions, mislabels, poor calibration, malfunctioning devices, and other data-gathering issues. While each situation is unique, developing methodologies to identify and repair the outliers and erroneous data becomes an essential and challenging task. In the literature, various techniques have been used to detect outliers. Some studies have applied distance-based approaches, which leverage neighborhood computation \citep{wang2018detecting,tran2020real} to classify and exclude outlier points. Density-based approaches have also been explored, where the density of each trajectory is computed, and low-density values are excluded as outliers \citep{tang2017local,duggimpudi2019spatio,riahi2021new,abid2021improved}. Pattern-mining is another methodology where outliers are classified by examining the varied correlations between trajectories \citep{homayoun2017know,cai2020uwfp}. On the other hand, some researchers have also directly applied machine learning approaches to learning the threshold for outlier classification when training on the trajectory data \citep{tian2020identifying,belhadi2021machine}.
    
After an outlier has been identified, it is crucial to understand the cause of the outlier data and why it exists. Specifically, it is desirable to determine if outliers were caused by anomalous network conditions or equipment failure. How to distinguish erroneous data from abnormal traffic conditions is a challenging task \citep{chen2010comparison}. However, some methods and data mining techniques can help cope with this problem. In recent work, \citet{kullberg2021learning} tried to detect outliers resulting from anomalous behavior using statistical hypothesis tests such as T-score and n-step prediction. Meanwhile, \citet{van2021data} leveraged knowledge on the variable of interest and its correlation with other factors to determine whether the detected outliers result from non-recurrent events or equipment errors.
    
While some studies focus on detecting and removing outlier data, others have attempted to repair detected outliers in various ways. For example, \citet{sani2019repairing} first caught outlier behavior based on the occurrence probability of an activity sequence concerning the surrounding contextual behavior. After that, the authors replaced the identified outlier activity sequences with one of higher probability within the same contextual behavior space. Meanwhile, \citet{liu2021experience} introduced TsClean, a uniform framework for identifying and repairing outliers in IoT data detected by various algorithms. Conversely, some research has demonstrated that outlier data can be helpful if analyzed separately from the primary data instead of being repaired or removed \citep{john2021outlier}. 
    
\item \textbf{Data imbalance.} One of the most prevalent problems in datasets is the imbalance of data distribution or labels. While easy to identify in smaller sets, it becomes challenging for machine learning algorithms to identify these rare cases in the large datasets emerging in recent years.Recently, \citet{peng2020examining} attempted to solve the issue of imbalanced data by optimizing the imbalanced classification algorithm concerning the output level, data level, and algorithm level. In another work, \citet{elamrani2020class} employed a Synthetic Minority Oversampling Technique (SMOTE) to rebalance the training sets and cope with a bias towards the majority class. 

\end{itemize}

\subsubsection{Data Storage and Aggregation} The massively increased volume of spatio-temporal data collected by various IoT devices and sensors within the traffic network (e.g., mobile phones, RSUs, traffic cameras) has exposed a need for improved data storage and aggregation systems. While the increased volume of data undoubtedly improves traffic prediction accuracy, it comes with considerable computation, transmission, and storage costs. Furthermore, the complex deep learning models used in the literature for modern traffic prediction generally require the data to be aggregated to a centralized location (e.g., a data center), presenting additional costs. In an effort to address these drawbacks, some recent studies have focused on leveraging various distributed systems, like blockchain networks, to store and shard the data efficiently for use in various ITS applications \citep{li2019blockchain, 9350858, 9107472}. FL is also a new approach to coping with these issues, where devices in the network train prediction models locally and only distribute the model parameters instead of the underlying data \citep{9082655}. That being said, these novel concepts need substantially more research to improve our understanding of their benefits within modern traffic prediction applications. 

\subsubsection{Data Compression} Data compression is another preprocessing method that aims to reduce the size of datasets while preserving the most critical insights. Different online and offline algorithms are proposed in the literature to compress data from large and diverse urban networks, with negligible loss of information and efficient representation of all types of data. The offline methodologies, which can be categorized as either simplification-based \citep{bermingham2017framework,zhang2018trajectory} or road network-based methods \citep{yang2017novel,li2021trace}, are applied to the raw trajectory data in order to eliminate some irrelevant data points and increase overall compression quality. On the other hand, the online methods focus on compressing trajectory data in a timely fashion by using window-based \citep{gao2018online,roy2021compression} or moving attribute-based methodologies \citep{al2018safepath,zhang2021trajectory}.

\section{Traffic Prediction Methods and Models}
\label{Sec4}

The traffic forecasting problem represents a spatio-temporal time series prediction problem, where the input contains a traffic variable (e.g., flow or speed) represented in one or more time series, and the output is a forecast of the future conditions. The prediction equation for a single time series can be represented as \citep{tedjopurnomo2020survey}:

\begin{equation}
 y_{t+T'} =f ([X_{t-T+1}, X_{t-T}, ..., X_t])  
\end{equation}

where $y_t$ is the predicted traffic at time $t$ and $X_t$ is the input denoting the attributes at the $t$th timestamp. $T$ indicates the input sequence length, i.e., how many time steps of historical traffic data are used as the input. $T'$ represents the prediction horizon, while $f$ is an arbitrary function that calculates the traffic prediction based on the input data and depends on the actual model used.

The field of traffic prediction has existed for almost six decades and covers a wide array of methodologies which can be divided into two main categories. Generally, traffic forecasting models are either univariate, where the input data consists of a single time series from one individual sensor, or multivariate, where the time series of multiple sensors throughout the network are considered within a single framework \citep{nagy2018survey}. Multivariate time series forecasting methods inherently assume interdependencies among variables. In other words, each variable depends not only on its historical values but also on the historical values of other variables. Today, most models are multivariate and attempt to capture the complex spatio-temporal relationships between the traffic data collected across the transportation network.

Most notably, the spatio-temporal relationship is fundamental to traffic forecasting. Many approaches for precise traffic prediction consider the topological structure of the urban road network (spatial) and the dynamic network changes over time (temporal). However, only recently has the joint spatio-temporal relationship been considered (e.g., the impact of historical trends from other networked nodes on the future state of a given node). The resulting data-driven traffic forecasting methodologies can be broadly categorized into three groups: (1) statistical methods, (2) traditional machine learning, and (3) deep learning. Statistical methods are especially suitable for smaller datasets, benefiting from a clear and simplified computational structure compared to more advanced machine learning approaches. However, traditional machine learning algorithms are better suited for capturing complex non-linear correlations and processing high-dimensional data inherent in transportation-related data.

Meanwhile, increased computational capabilities and data access have resulted in the popularity of deep learning-based models, which have highly complex structures and have been shown to outperform many of the widely used traditional approaches assuming sufficient data. That being said, given the importance of jointly capturing the localized and network-wide spatio-temporal relationships and the varying trade-offs of the individual models, hybrid approaches combining multiple models have become popular amongst researchers in recent years \citep{gu2019improved, hajirahimi2019hybrid,alsolami2020hybrid}. Generally, the hybrid approaches follow an ensemble-based design \citet{dietterich2000ensemble}, where predictions from multiple models are weighted and fused to produce the final prediction. Recently in traffic forecasting, the new Bayesian combination method (NBCM) proposed by \citet{wang2014new} has been demonstrated to make efficient use of diverse sub-predictors. Alternative hybrid approaches seek to connect different DL architectures in sequence, such that the hidden representation learned and output by one network can be used as input to a subsequent model \citep{li2021hybrid}. In these designs, a convolution-focused model (e.g., CNN, GCN) extracts essential features from the input data with efficiently deep architecture and subsequently inputs the encoded features into an RNN-based gated design to capture the temporal dependencies. We discuss the most interesting hybrid approaches in the existing literature in future subsections.

In the remainder of this subsection, we provide a model-driven review of the existing literature on multivariate traffic forecasting models, highlighting their scientific contributions and unique approaches, which improved our understanding and ability to forecast traffic.

\subsection{Statistical Methods}
Statistical models, which treat forecasting as a regression problem, are widely applied to short-term traffic prediction \citep{yang2020evaluation} due to their clear computational structure and strong theoretical interpretation ability. For the non-learning approaches, such as k-nearest-neighbors (KNN) \citep{zhang2013improved, harrou2020traffic}, historical average (HA), and the conventional vector autoregressive (VAR) models, considering the effects of both the downstream and upstream traffic generally improves prediction performance.

Conversely, non-linear regression \citep{hogberg1976estimation}, average algorithm, smoothing, Bayesian networks (BN), and Kalman filtering (KF) are common parametric techniques for analyzing and forecasting time series using observed ground truth data. In addition, the autoregressive integrated moving average (ARIMA) \citep{williams2003modeling}, and its variants, are among the most consolidated approaches and have been frequently applied to traffic prediction problems \citep{williams2003modeling, shekhar2007adaptive, li2012prediction, moreira2013predicting, lippi2013short, wagner2017functional}. ARIMA, an extended form of the autoregressive-moving average (ARMA) model, is specially used for future points prediction in time series data. Other variants of ARIMA, including Seasonal ARIMA (SARIMA), ARIMA with explanatory variable (ARIMAX), Kohenen ARIMA (KARIMA), and Vector ARIMA also exist in the literature and have been shown to improve traffic flow prediction accuracy. 

Although suitable for smaller datasets with only a few time series having short observation periods, due to a simplified and transparent computational structure, the statistical methods generally lack the complexity to consider the spatio-temporal relationship and instead focus exclusively on temporal time series data. Thus, the prediction capabilities of these statistical models are limited for the traffic prediction application due to their inability to account for the inherent spatial dependencies. To overcome this challenge, some studies developed extended time series-based approaches, where the extension took into account spatial and temporal interactions in a new manner, such as ST-ARIMA \citep{min2009short,min2010urban} and VARMA \citep{min2011real}. Since these approaches use explicit parameterization functions with strong assumptions during the modeling process, they are unsuitable for simulating real traffic application scenarios \citep{cheng2018spatiotemporal}. In contrast, non-parametric spatio-temporal models such as ST-KNNs \citep{wu2014improved}, without any requirement for prior knowledge or explicit mathematical expression, can easily achieve satisfactory portability and comparatively greater prediction accuracy. However, these statistical methods are ineffective at processing complex (high-dimensional) and dynamic time series data. For example, ST-KNNs, which use distance functions and correlation coefficients to identify spatial neighbors and measure the temporal interaction by only considering the temporal closeness of traffic, cannot fully reflect the essential features of road traffic.

Despite the drawbacks of classical statistical approaches when processing the highly-dimensional ITS data, some hybrid methods integrating classical and DL-based methods have emerged to overcome the existing challenges \citep{li2017short}. For example, \citet{li2019learning} recognized the strong ability of state space and hidden Markov models for capturing the importance of exogenous variables (e.g., weather, special events). However, these methods are not suitable for modeling long-term dependencies or non-linear relationships, two key considerations when forecasting traffic. Accordingly, the authors relax the first-order Markov assumption to include a sequence of deterministic states generated by an RNN-based model, effectively capturing the long-term dependencies. A data-driven sparse Bayesian learning approach (SPL) is proposed to determine the relevant importance of features at the current time step, pruning away any low-impact or redundant features while mitigating overfitting to exogenous variables. Consequently, the model can process many exogenous features while selectively considering the most critical subset of external data at the current moment. For MTP, this is a promising approach for training a model to capture the sparsely represented but high-intensity exogenous variables (e.g., a rare accident on a bridge or a monthly concert event).

In summary, exogenous variables are important considerations for traffic prediction to improve our models. However, the case study in this paper is limited and only considers exogenous temporal variables, such as whether it is a workday or peak period. More research is needed to see if this approach to considering exogenous variables can be further improved and integrated into the cutting-edge MTP models.

\subsection{Traditional Machine Learning}

\begin{table}[t]
\centering
\tiny
\begin{threeparttable}
\caption{ Traditional ML traffic prediction methods}
\label{bbbTML}
\begin{tabular}{p{0.001\textwidth}p{0.001\textwidth}clllll}
\hline
\multicolumn{3}{c}{\textbf{Characteristic}}& \textbf{Model}&\textbf{Advantage}&\textbf{Disadvantage}\\\hline

\multirow{14}{*}{\rotatebox{90}{Process high-dimensional }}&\multirow{14}{*}{ \rotatebox{90}{data and capture complex}}&\multirow{14}{*}{\rotatebox{90}{non-linear relationships.}}&\multirow{2}{*}{Feature-based}&\textendash Developing a regression model with &\textendash Limited and Inconsistent performance across 
\\
 &&&\multirow{2}{*}{Models}&human-engineered features &various applications due to reduced  \\
&&&&\textendash Easy-to-implement method&complexity and the human-engineered features\\\cline{4-6}

&&&\multirow{5}{*}{Gaussian process}&\textendash  Use various kernel functions to model the&\multirow{4}{*}{\textendash   Hard to process large amount of data}  \\
&&&\multirow{5}{*}{Models}&inner characteristics of traffic data &\multirow{4}{*}{\textendash  Have higher computational load and storage}\\

&&&&\textendash Effective and feasible in traffic prediction&\multirow{4}{*}{pressure than feature-based models}\\
&&&& based models& 
\\
&&&& \textendash Gets spatial and temporal relationship & \\
&&&& simultaneously& \\\cline{4-6}

&&&\multirow{5}{*}{State-space}&\textendash  Assume that the observations are generated&\multirow{3}{*}{\textendash  The nonlinearity of these models is} \\

&&&\multirow{5}{*}{Models}&by Markovian hidden states.& \multirow{3}{*}{restricted, and they are not always suitable for} \\

&&&&\textendash   Can naturally represent the uncertainty of&\multirow{3}{*}{simulating complex and dynamic traffic data.}\\

&&&&the system and better capture the latent &\\

 &&&&structure of the spatio-temporal data.&\\

\hline

\end{tabular}

  \end{threeparttable}
\end{table}

In addition to the classic statistical methods, ML-based methods are becoming increasingly popular in traffic prediction. Compared to classical methods, ML models can provide more robust generalization capabilities while also being able to learn more complex relationships and better adapt to the changing conditions of the traffic network \citep{yang2020evaluation}. 

As shown in Table \ref{bbbTML}, traditional ML approaches can be segmented into three categories: feature-based methods, Gaussian process models and state-space models \citep{yin2021deep}. Feature-based methods are used to train regression models for some practical traffic prediction problems using human-engineered traffic features \citep{lin2017road,li2018general,guan2018unified,tang2018forecasting, kowalek2019classification}. While the feature-based models are easy to implement, Gaussian process approaches utilize multiple kernel functions to characterize the inner features of traffic data, incorporating both spatial and temporal correlations simultaneously \citep{matthews2018gaussian,salinas2019high,liu2020gaussian}. Although research has demonstrated Gaussian methods to be useful and practical for traffic prediction, they have higher computational complexity ($O(N^3)$) and data storage requirements ($O(N^3)$) than feature-based models \citep{datta2016nearest}. This presents scalability issues when working with large datasets, which are increasingly common in traffic prediction applications.

Meanwhile, state-space models presume that the observations are derived from Markovian hidden states. State-space models excel at capturing latent data structures and can naturally model the uncertainty of a system, which is a desirable property for traffic prediction applications. However, these models have difficulty generalizing non-linear relationships \citep{tan2016short,duan2018unified,shin2018vehicle,ishibashi2018inferring,gong2018network}, and they are not always the best choice for modeling complex and dynamic traffic data, especially for long-term forecasting.

\subsection{Deep Learning for Multivariate Traffic Prediction}
Deep learning (DL) models have become increasingly popular in recent years, and research has demonstrated their strong applicability for MTP and capturing non-linearity \citep{lai2018modeling,shih2019temporal}. Previously, deep learning approaches were widely considered impractical due to their increased data, computation, and storage requirements compared to classical methods. However, stemming from a desire to utilize better the enormous quantities of data generated in the previous two decades, scientists have demonstrated that DL methods have the potential to significantly outperform the widely used approaches if given sufficient resources. This subsection introduces the most popular and widely used DL models within the literature, focusing on the traffic prediction application.

\subsubsection{Basic DNN Models}
\begin{itemize}

\item \textbf{MLP}. The multi-Layer perceptrons (MLP), sometimes called vanilla neural networks, are a feed-forward neural network (FNN) class generally having more than three layers of fully-connected neurons. A defining characteristic of these networks is the absence of loops, meaning that neurons form connections between layers but never connect within a layer. The information flows from the input layer to the output layer, hence the name feed-forward. MLP uses the backpropagation training technique, a supervised learning algorithm that learns the desired output from varied input data. Like RNN, exploding and vanishing gradients present challenges and can produce poor results when training on large MLP networks. To overcome this challenge, a resilient backpropagation algorithm, Rprop, was proposed to update the learning rate for each neuron connection dynamically. Later,  \citet{oliveira2016computer} showcased the benefits of Rprop in network traffic prediction by comparing the model with standard MLP, RNN, and stacked autoencoder (SAE). 

Additionally, \cite{song2017traffic} compared the MLP-based traffic prediction model with a CNN-based approach, and the results showed CNN outperforming MLP models. Notably, MLP has issues reflecting local dependencies in the traffic data and is also more prone to noise than other approaches. While initially popular for traffic prediction due to its ability to generalize the nonlinear relationships in traffic data, MLP is falling out of use in favor of more powerful DL models better suited for traffic prediction (e.g., RNN and CNN).

\item \textbf{AE}. Autoencoder (AE) methods are a variety of artificial neural networks that learn an efficient encoding of unlabeled data in an unsupervised manner. In traffic prediction, AEs are primarily leveraged to impute or effectively compress data \citep{ghosh2017denoising, wang2018network}, reducing its dimensionality while retaining the most vital information. The network design is relatively simple at the highest level, with two main portions: (1) an encoder that compresses the input $(x)$ into fewer bits; and (2) a decoder that takes the sparse representation of the input $(x')$ and outputs the original value.

More specifically, AEs \citep{liou2014autoencoder} are based upon unsupervised learning (unlabeled training example) techniques, having a deliberately placed bottleneck between the encoder and decoder for learning a compressed representation of the input. Moreover, AEs comprise the same number of input and output units along with one or more hidden layers (Fig. \ref{fig:AE}). Eqs. \ref{AE1} and \ref{AE2} represent the structure of AEs with a single hidden layer where $f$ and $g$ are encoding and decoding functions, respectively; $b$ is the bias value; $W$ and $W'$  are encoding and decoding weight matrix. AEs are utilized for tasks such as dimensionality reduction, classification, and anomaly detection because they generate similar input at the output layer. In particular, the bottleneck is an essential feature of AE that restricts the amount of data allowed to traverse the network. By learning to recreate the input from a compressed representation, AEs learn the most fundamental aspects of sparse input material. Besides, AEs use backpropagation (BP) techniques to train an approximation identity function that generates a target value comparable to the input. However, they demand a pretraining stage that may lead to a vanishing gradient problem \citep{mahmud2018applications}.

\begin{figure}[t]
    \centering
    \includegraphics[width=.5\textwidth]{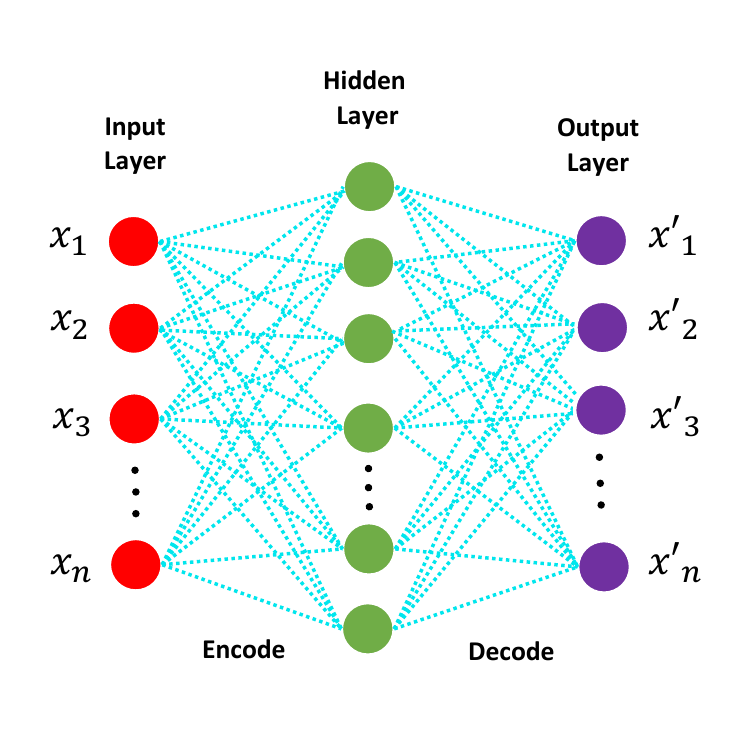}
    \caption{Structure of AE.}
      \label{fig:AE}
      \end{figure}

The following example presents a simple introduction to the internal workflow of an AE. First, let us assume a set of training samples $x_1, x_2, x_3, ..., x_n$. After passing through the AE, the first input $x_i$ is encoded to hidden form $y(x)$ using Eq. \ref{AE1} and then decoded back to $x_i^{’}$ using Eq. \ref{AE2}. The training procedure aims to minimize the reconstruction error $\delta(x, g(f (x)))=\delta(x, x')$, i.e., difference between input and re-constructed output.
That being said, if the count of neurons in hidden layers is the same or greater than the input and output layers, then the model behaves the same as an identity function (e.g., it just copies the inputs). Therefore, determining the ideal number of hidden units is a challenging and application-specific task requiring careful fine-tuning through trial and error \citep{palm2012prediction}.

\begin{equation}
\label{AE1}
y = f (W x + b) 
\end{equation}
\begin{equation}
\label{AE2}
x' = g(W' y + b')
\end{equation}

For MTP applications, AEs can be applied in multiple ways, including for data imputation \citep{ghosh2017denoising}, data compression \citep{wang2018network}, and feature extraction \citep{liu2019traffic}. Specifically for traffic prediction, AEs are typically stacked in sequence to hierarchically extract the most important features from the traffic data. Referred to as Stacked AEs (SAEs), the hidden representation from the first encoder is passed to the subsequent AE layer, and this process can continue for a variable number of layers. The dimensionality is reduced as the input data passes through each layer, and only the most important spatio-temporal features are passed to the next layer. In this way, SAEs present another method for performing hierarchical feature extraction, in contrast to convolutional or gated recurrent network architectures \citep{wang2020scalability}. However, deep SAEs require a large volume of highly-dimensional input data for training, which must be preprocessed, and suffer from long training times before using the output features to train the final prediction model or layer. Furthermore, due to the black-box nature of NNs and the hidden representation output by the encoder, the features learned by the SAE are not well understood.
\end{itemize}

\begin{itemize}

\item \textbf{RBM $\&$ DBN}. Restricted Boltzmann Machines (RBMs) are stochastic neural networks that learn in an unsupervised fashion and are widely used to model binary variables in the literature (Fig. \ref{fig:RBM}). Different from the traditional neural network models, RBM joins the feature learning part based on the original multilayer neural network. The feature learning portion imitates the human brain to handle data signal classification. The specific operation is to increase the partial connection of the convolution layer and dimension layer in front of the original fully connected network layer. In simple terms, the traditional shallow neural network projections step is from the feature mapping to the value, and the characters are artificially selected. RBM projections step is from the signal to the feature and then to the value. The network freely chooses the data characteristics.
For example, \citet{ma2015large} applied a stacked RBM to extract temporal and spatial correlations from traffic flow data for traffic congestion forecasting. Notably, RBMs effectively extract representative characteristics from traffic flow data and capture the interaction between variables. Additionally, multiple variations of the RBM have been utilized in the literature, including Deep Belief Network (DBN) and Deep Boltzmann Machine (DBM). DBN combines several RBMs to capture low-pass network traffic components and elucidate the underlying data's long-range dependencies. More specifically, DBN can learn internal representations from the input data (higher layers capture complex statistical structures) and make inferences quickly (hidden layers state computation). 

\begin{figure}[t]
    \vspace*{-3cm}
    \centering
    \includegraphics[width=.95\textwidth]{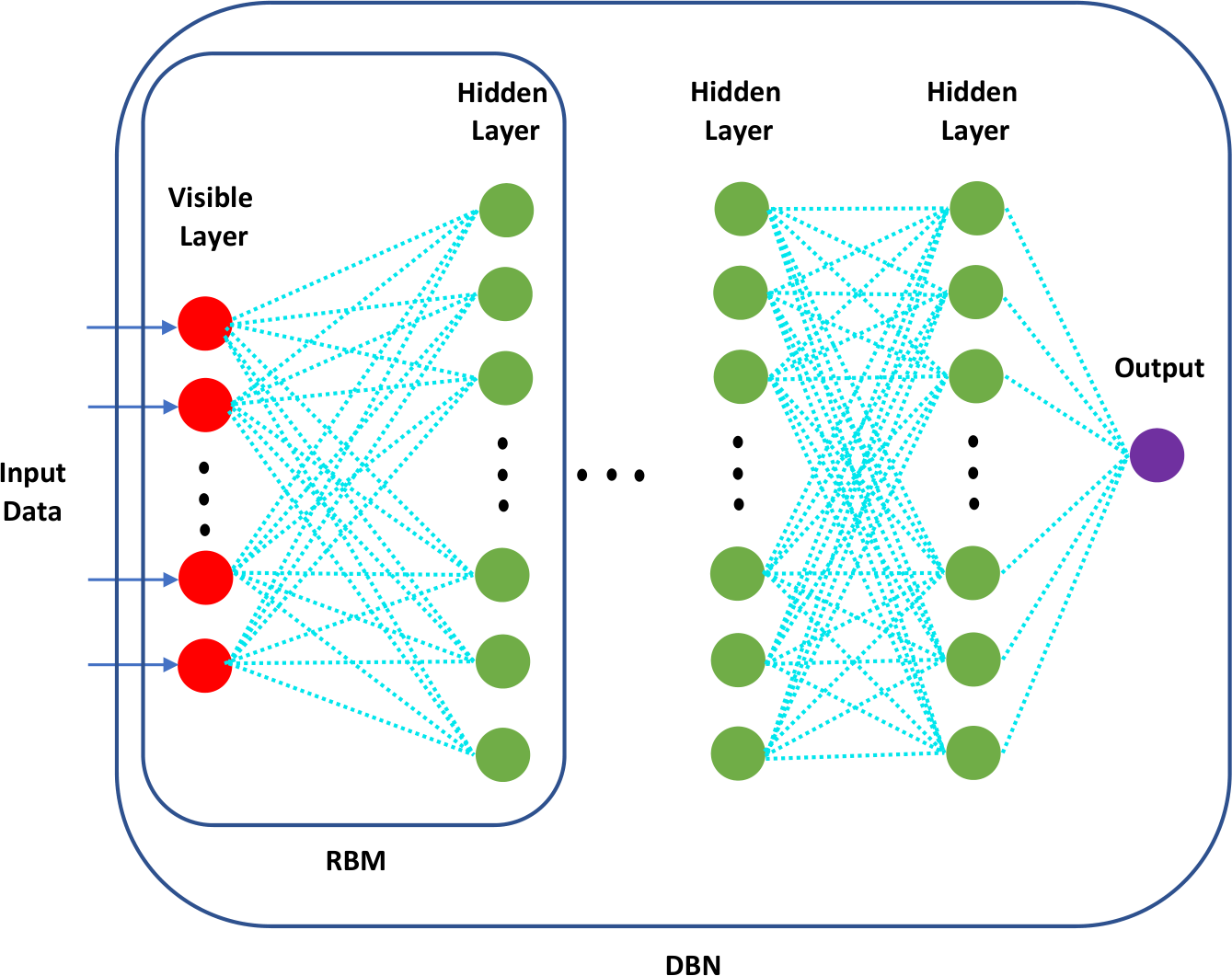}
    \caption{Structure RBM $\&$ DBN.}
      \label{fig:RBM}
      \end{figure}

Although DBN can effectively learn the object features in an unsupervised manner, this model has difficulty mapping the complex relationships between the elements in the traffic system. To overcome this challenge, \citet{bao2021improved} introduced an improved DBN by integrating SVR with DBN to map the complex relationship between the elements in the traffic system. On the other hand, DBM is a modified Boltzmann Machine that is unrestricted, containing more hidden layers and directionless connections between nodes. DBM provides additional advantages resulting from its top-down feedback loop, allowing higher-level information to be leveraged for resolving uncertainty regarding basic or intermediate-level features.

\end{itemize}

\subsubsection{CNN}
\begin{figure}[t]
    \centering
    \includegraphics[width=.99\textwidth]{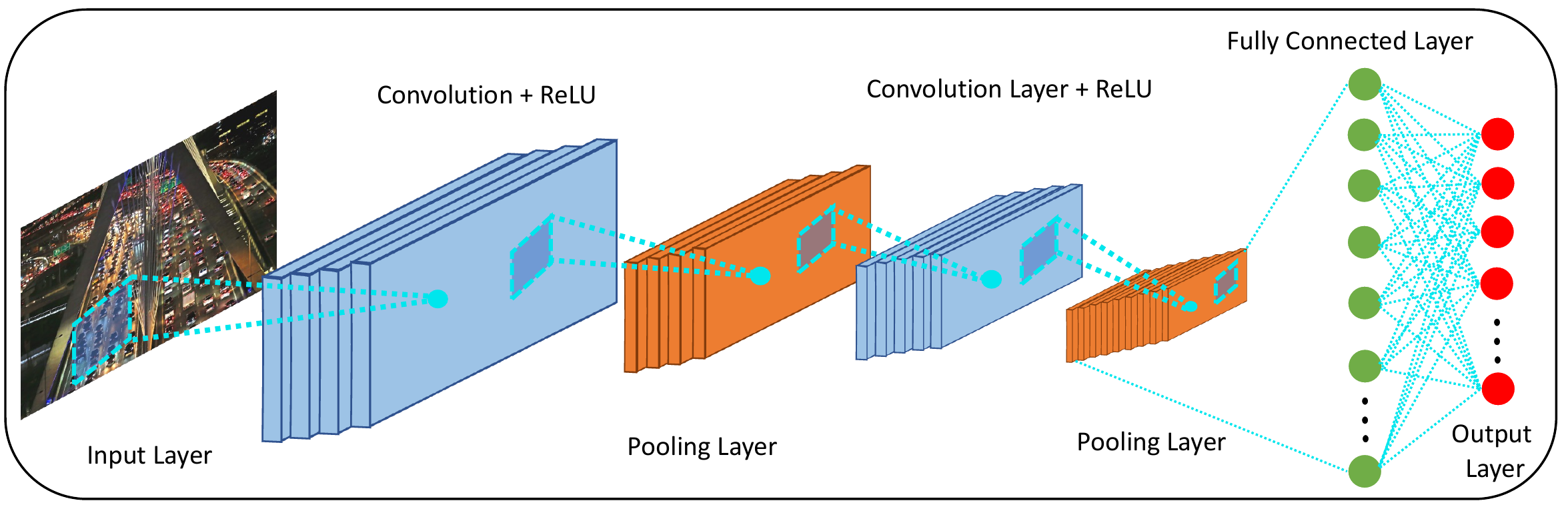}
    \caption{Structure of CNN.}
      \label{fig:CNN}
\end{figure}

Convolutional neural networks (CNNs) are one of the most representative supervised deep learning approaches for modeling spatial or temporal correlations. CNN architecture contains four main layers: convolutional, ReLU (activation), pooling, and fully connected. Moreover, some studies denote the input and output as two additional layers positioned at the beginning and end of the network, respectively. During the training process, the model first extracts features from image data into convolution layers, which are then convolved using a filter with learnable weights. Notably, different models and applications require varying numbers of convolutional layers. Following the convolution layer, the ReLU layer uses an activation function to set the negative values in the incoming data to 0. Next, in the pooling layer, the dimensionality of the input is reduced by subsampling (pooling) layers through a feature map size reduction method. This layer also helps to reduce the number of parameters by discarding unimportant information while keeping the most marginal and critical parameters. Lastly, the input is flattened and fed into fully connected layers to map the input to the required output dimension. The bundle of convolution layer and pooling layer can repeat multiple times to generate a deeper neural network. For example, Fig. \ref{fig:CNN}  shows a CNN architecture with two convolution layers and two pooling layers. With more layers in the CNN network, higher-level information is hierarchically captured. The drawback is that more layers add additional parameters, which may lead to over-fitting and increased training times.
 
For MTP applications (e.g., flow or speed), CNNs provide one method for capturing the localized spatio-temporal dependencies between links or nodes within the roadway network. The intuition behind applying CNN to MTP problems revolves around the convolution and pooling operations, which collectively learn localized features from the input data. Notably, these features are discovered in an unsupervised manner spontaneously from the raw input, enabling the model to understand previously unknown localized dependencies. Since traffic time series data from nodes in close proximity along the same link are likely to be related, multiple works have demonstrated the effectiveness of leveraging convolution-based DL models for extracting localized spatial dependencies from the multivariate time series data \citep{wu2018hybrid}.

In one of the first works to explore using CNN for MTP, \citet{song2017traffic} compared two versions of the well-researched MLP architecture with a CNN-based model for predicting traffic speed across four directly connected roadway links within Seoul. In this design, a CNN model is developed for each link to capture the intra-link node dependencies, while a fifth CNN models the temporal features. Lastly, each model's output is aggregated to produce the final prediction. The case study demonstrated that the CNN approach could better represent the localized spatial dependencies between nodes within the same link, improving prediction accuracy by upwards of 6\% compared to the state-of-the-art MLP approaches. However, the CNN model used in this paper only performs convolution on the subset of nodes within a single link, ignoring the inter-link dependencies. Expanding the CNN model's spatial dimension presents a trade-off between performance and computational demand. Capturing the long-range interlink dependencies between distant infrastructure (e.g., the relationship of flows between a suburban freeway and the CBD/downtown area) requires many convolutional layers and increases the network depth.

Later, in an alternative approach by \citet{wu2018hybrid}, a hybrid model is proposed by combining CNN with LSTM to capture the spatial and temporal traffic trends separately. Specifically, the CNN model extracts the spatial features of the traffic time series, while the LSTM model extracts the short- and long-term temporal patterns. Motivated by related work in computer vision, \citet{springenberg2014striving}, the authors decide to perform full convolution within their CNN model by removing the pooling layers. Notably, this contrasts with previous work where the conventional CNN design is applied. Since the primary purpose of pooling is to reduce the size of feature maps, removing the pooling layers can help preserve the underlying structure of the roadway network, which would otherwise be distorted. Interestingly, related work in crowd flow prediction also reported improved prediction accuracy when using a full convolutional network \citep{zhang2017deep}. 

Building upon the earlier approaches, \citet{liu2019deeprtp} proposed DeepTRP: a modified convolutional model for MTP forecasting, which also removed the pooling layer. In the case study, the authors take a big data approach to regional traffic forecasting, where an area consisting of 17,017 road segments is divided into a \emph{$4\times8$} grid of 32 regions. In contrast to \citet{wu2018hybrid}, DeepTRP performs 2D convolution on a 2D matrix representation of the traffic data while pre-processing the data into hourly, daily, and weekly groups. By using 2D convolution, the inter-regional relationships can be better learned during model training, whereas \citet{wu2018hybrid} outlined that 1D convolution is best suited for small study areas consisting of only a few road segments. However, 2D convolution and deeper convolutional networks are more challenging to train, can be prone to exploding gradients, and run the risk of overfitting. To mitigate against this, \citet{liu2019deeprtp} proposed stacking a variable number of residual units between the first and last convolutional layers, a method previously successful in related areas such as image recognition \citep{he2016deep}. Notably, their sensitivity analysis revealed that increasing the number of residual units beyond two harmed performance and that more fine-grained partitioning of regions also increased prediction error. For future work, experimentation with a significantly deeper convolution-based model is necessary to assess whether additional convolutions can better capture the complex and fine-grained regional dependencies and allow for a finer-grained regional representation.

Additionally, other works have identified the importance of considering regional dependencies within the existing models. Notably, the standard convolutional methods proposed in many works operate on a fine-grained representation of the transportation network, focusing only on the spatio-temporal relationships between directly adjacent nodes (e.g., neighborhoods). Recognizing this deficiency, \citet{pan2019matrix} proposed a matrix factorization method for extracting the high-level, regional relationships from multivariate traffic flow data. The most notable contribution of this work is the design approach taken by the authors, who ensure their proposed regional feature extraction method is compatible with the existing state-of-the-art finer-grained methods (e.g., \citet{zhang2017deep}). In future research, it would be interesting to experiment with a combined weighted prediction approach, where the regional predictions can be weighted and combined with the outputs of existing cutting-edge but finer-grained methods.

Another modification to the traditional CNN structure proposed the convolutional LSTM (Conv-LSTM) to jointly consider the spatio-temporal features within the multivariate traffic time series data \citep{zheng2020hybrid}. The primary contribution of this work is to combine the CNN with the LSTM network sequentially, enabling the model to jointly consider the short- and long-term spatio-temporal trends from a single set of input data. Prior hybrid approaches, including \citet{wu2018hybrid} treated the spatial and temporal features separately, which may overlook some of the intricate relationships between the spatial location of the infrastructure and the period of interest (e.g., the impacts of rush hour CBD congestion on the traffic in residential areas). Using the convolutional layers' output as input to the LSTM network, the spatial relationships between nearby flow sequences are captured concerning changing temporal network dynamics, such as differences in weekday and weekend traffic patterns. Case study results demonstrate that the Conv-LSTM model can outperform the standalone LSTM and SVM models, while \citet{dai2020spatio} also utilized a Conv-LSTM for traffic forecasting with great success. In \citet{dai2020spatio}, the performance of a hybrid model combining the Conv-LSTM and GCN architectures is explored, incorporating cutting-edge convolutional methods in both the temporal and spatial feature extraction modules. The comparison highlighted that combining both models yielded about a 4\% increase in RMSE accuracy compared to using a Conv-LSTM or GCN alone. Consequently, it is vital to consider specialized approaches for extracting the study area's spatial and temporal features.

While the works mentioned above apply more traditional CNN model structures to MTP problems, related research proposed modified convolution-based networks optimized for time series forecasting. Notably, the seminal work of \citet{lea2016temporal}, who first introduced the temporal convolution network (TCN) for capturing global patterns across high-dimensional time series with diverse scales in the action segmentation field, has inspired innovation. Before this work, the primary method involved using two high-level DL models, usually a CNN and an RNN, to capture local and global patterns. In contrast, the proposed TCN method provides a unified approach for hierarchically extracting temporal features at the local, intermediate, and global levels using a combination of 1D convolutions, pooling, and channel-wise normalization. 

Notably, the 1D Convolution layer in TCN has two differences compared to standard CNNs: (1) Causality, which implies that each layer output is produced looking only at the most recent
historical samples, and (2) Dilation, which ensures that each filter only examines a subset of historical samples. More simply, rather than performing convolution on a contiguous time window of input samples, dilation inserts a fixed step between convolution inputs, thus increasing the receptive field while keeping the number of parameters low \citep{burrello2020predicting}. Besides, the 1D convolution-based design improves training time significantly for time series modeling compared to RNN-based models due to the activation functions being computed hierarchically instead of sequentially, as in RNN. While the TCN was designed for action segmentation, it solved a shared problem with MTP models: how to jointly capture the low-level and localized spatio-temporal features with the high-level temporal trends (e.g., seasonality).

More recent work has attempted to improve the original TCN design and apply it to MTP forecasting. \citet{sen2019think} proposed DeepGLO: a TCN-based hybrid forecasting model designed for a large-scale MTP problem, including time series from 228 traffic sensors. An essential contribution of this work is the ability of DeepGLO to consider the global network patterns during prediction. In contrast, alternative approaches (e.g., CNN, RNN) generally focus on only the local past data (e.g., the past data from a specific detector only), despite being trained on the entire set of time series. DeepGLO thinks globally but acts locally by using a temporally regularized matrix factorization (TRMF) model, normalized by a TCN, to output features representing the global trends. Notably, the regularization process by the TCN is required to capture non-linear relationships, as the standard TRMF approach can only describe linear trends. The output global-level features are then used as covariates for another TCN model, producing a final model that jointly considers the historical local and global trends during prediction. Case study results demonstrate the effectiveness of DeepGLO for MTP compared to the existing cutting-edge methods, including the spatio-temporal graph convolutional network (STGCN), a competing approach discussed later. However, DeepGLO is not directly compared to other competing hybrid models, making cross-comparison difficult.

In summary, research has demonstrated the effectiveness of CNN-based models for capturing complex spatio-temporal dynamics in multivariate traffic time series data. Both 1D and 2D convolution show promise for identifying relationships between adjacent time series at different spatial granularity; however, they require different input representations and varying sequentially stacked layers and filters. Most recently, with great success, cutting-edge approaches have integrated the convolutional operation into new hybrid models, such as Conv-LSTM, TCN, and DeepGLO. However, it is crucial to consider that the traditional convolution operation is designed for data in Euclidean space, which can distort the underlying roadway network structure and destroy critical spatial relationships. Furthermore, the pooling operation traditionally included in the CNN model further distorts the road network through subsampling, and multiple research works found success by using fully convolutional approaches instead. That being said, cross-comparison between the various methods within the literature is challenging because the studies use varying datasets and compare their methods to different baselines. Standardized baseline datasets for other MTP applications and a standardized set of baseline models for comparison with new approaches would greatly enrich the existing literature.

\subsubsection{RNN}

\begin{figure}[t]
    \centering
    \includegraphics[width=.99\textwidth]{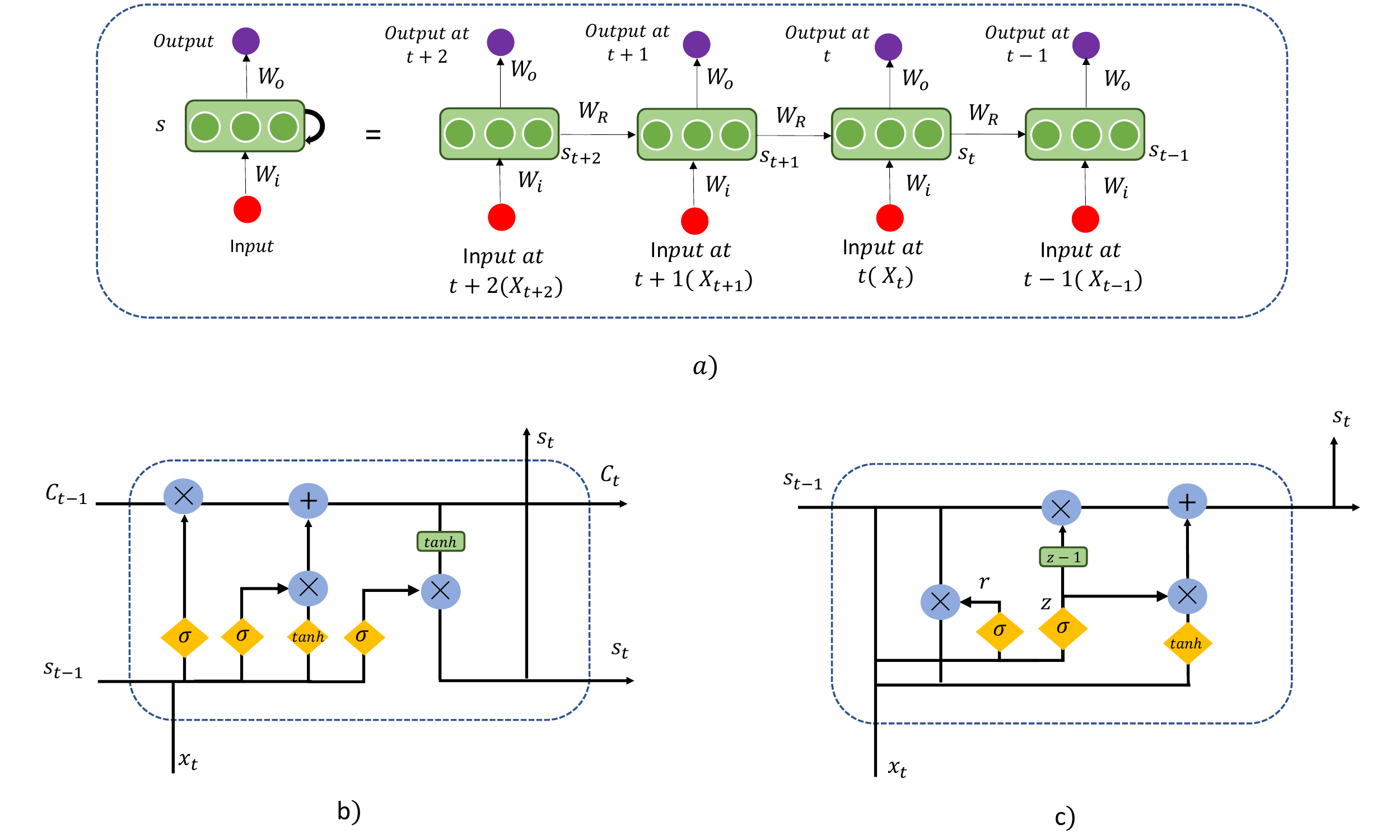}
    \caption{Structure of a) RNN, b) LSTM and c) GRU.}
      \label{fig:RNN}
\end{figure}

Recurrent neural networks (RNN) represent an evolution of the FNN model, designed to improve performance for time series data. The conventional FNN ANN designs do not perform well when modeling sequences and time series as they lack a memory component. In contrast, RNNs work well with sequential and time series data due to adding a memory cell. More specifically, RNN models process input sequentially and maintain an internal memory by constantly updating a set of hidden states. In this way, RNN-based models consider the dependency between adjacent time steps, whereas traditional FNNs treat time steps independently. During prediction, the input at the current timestep is taken along with the output from the previous timestep as depicted in Fig. \ref{fig:RNN}(a). In other words, the RNN learns by updating a state $s_t$ depending on the prior output at state ${t-1}$ and the current input $x_t$ at time t. The RNN network is represented by the Eq. (\ref{RNN1}) and (\ref{RNN2}), where $w_i$ is the weight matrix from the input to the hidden layer, $w_R$ is the recurrent weight matrix, and $w_o$ is the weights associated with the hidden to the output layer. Memory is realized by configuring each hidden neuron with a feedback loop that passes the current output as the input during the next step. Due to the RNN network's use of loops, the original Backpropagation (BP) approach is ineffective. Therefore, the network is trained using Backpropagation Through Time (BPTT) \citep{werbos1990backpropagation}. BPTT is a version of normal BP in which gradient descent is applied to an unrolled network. While unfolding all timesteps, the error of the current time step is backpropagated to the first timestep in BPTT. Besides, BPTT can be expensive for high-dimensional time series analyses with several timesteps.

\begin{equation}
\label{RNN1}
s_t = f_s(w_i x_t + w_R s_{t−1} + b_s) 
\end{equation}
\begin{equation}
\label{RNN2}
y_t = f_t(w_o s_t + b_y)
\end{equation}

The first proposed RNN was a two-layer, fully-connected neural network with a feedback loop in the hidden layer, with the addition of the feedback loop being the primary contribution. However, this simple design becomes prone to gradient vanishing or exploding when training large networks due to the backpropagation involved, presenting problems when working with extended multivariate traffic time series. More specifically, adding more hidden layers in the network results in more derivatives that must be multiplied during each network pass. As a result, if the products are large, the gradient can explode in large networks and result in arithmetic overload. On the other hand, if the derivatives are very small, the continued multiplication will result in a vanishing gradient and potential arithmetic underflow. Summarily, for conventional RNNs, the effect of past input on output decays exponentially as it cycles around the recurrent connection. Hence, RNNs can only deal with the problem of sequence handling for short-term dependencies and fail when modeling long-term dependencies due to the vanishing gradient problem. The literature has proposed a new class of Gated RNN with multiple design approaches for various applications to address this problem. The most common variations of Gated RNN are the Long short-term memory (LSTM) \citep{hochreiter1997long} and Gated Recurrent Unit (GRU) \citep{cho2014learning} models. 

For MTP applications, LSTM is favored over conventional RNN due to its ability to remember long-range dependencies selectively. Consequently, the LSTM network can improve modeling performance for time series containing both long- and short-term dependencies. Similar to the conventional RNN, the LSTM NN includes an input layer, a recurrent hidden layer, and an output layer. However, it treats the hidden layer as a memory unit that helps remember the longer-term dependencies. The hidden layer comprises a series of memory blocks called cells, each with a self-connection that enables retaining a dynamic temporal state. In addition, each block consists of three multiplicative gates, i.e., one input gate, one output gate, and one forget gate, as shown in Fig. \ref{fig:RNN}(b). The multiplicative gate enables cells to keep and retrieve information over a long period, which helps mitigate the vanishing gradient problem. Each cell transfers cell state and hidden state to the immediate next cell. Here, the input gate controls whether the memory cell is updated or not. In contrast, the forget gate indicates whether the memory cell should be set to 0, i.e., the LSTM network no longer requires the information. This way, the LSTM model can selectively discriminate between useful information as conditions change over time. 

Moreover, a primary difference between LSTM and RNN is that the former uses a forget gate to control the state of a cell and ensure that it does not degrade, further improving the long-range dependency modeling performance. The Output gate controls if the current information of the memory cell is to be made visible or not, i.e., it generates output for the LSTM network, which can be an interpretable prediction or a hidden state. In cases where the output is a hidden state (e.g., latent representation), the hidden state is generally used as the input features for another layer or model, such as an MLP or fully connected layer, to produce the final prediction output \citep{ziat2017spatio}. In other words, the hidden representation can be interpreted as a compressed representation of the input data containing the most important features. Consequently, if the hidden representation is learned, such as in the RNN-based STNN model proposed in  \citet{ziat2017spatio}, it can provide one method for extracting the most important features from the stochastic and noisy traffic data. Eqs. (\ref{DI})-(\ref{DE})  depict the working of the LSTM network:

\begin{equation}
\label{DI}
i = \sigma (W_i x_t + U_i s_{t−1} + V_i C_{t−1} + b_i)
\end{equation}
\begin{equation}
f = \sigma (W_f x_t + U_f s_{t−1} + V_f C_{t−1} + b_f )
\end{equation}
\begin{equation}
o = \sigma (W_o x_t + U_o s_{t−1} + V_o C_{t−1} + b_o)
\end{equation}
\begin{equation}
C_t = f * C_{t−1} + i ∗ g(W_c x_t + W_c s)_{t−1} + b_c) 
\end{equation}
\begin{equation}
s_t = o ∗ h(C_t)
\label{DE}
\end{equation}

where $i$, $f$, $o$ represent input, forget, and output gates activation vectors, respectively; and $C_t$ is the cell state vector, while $h_t$ is the output vector. $W$, $U$, and $V$ are weight matrices.

To address the issue of vanishing gradient that the LSTM network experienced, \cite{chung2014empirical} introduced the GRU model. Compared to LSTM, GRU training is more efficient due to a smaller number of learnable parameters, as there is no separate memory block. Despite having less complexity, GRU has been demonstrated to perform similarly to LSTM for most MTP applications. The reset gate and the update gate of the memory cell of a GRU unit determine what information should be sent to create the output. As shown in Fig. \ref{fig:RNN}(c), unlike LSTM, GRU merges the forget gate and input gate into a reset gate. The reset gate functions similarly to the forget gate of LSTM in that it determines whether or not to integrate present and past information. In contrast, the update gate defines how much prior knowledge should be retained. The following equations define the GRU NN's inner workings \citep{miglani2019deep}:

\begin{equation}
\label{DV}
z = \sigma(x_t U_z + s_{t−1} W_z) 
\end{equation}
\begin{equation}
r = \sigma(x_t U_r + s_{t−1} W_r)
\end{equation}
\begin{equation}
h = tanh(x_t U_h + (s_{t−1} ∗ r)W_h)
\end{equation}
\begin{equation}
s_t = (1 − z) * s_{t−1} + z ∗ h
\end{equation}

where, $x_t$ indicates input vector, $z$ shows update gate vector, $r$ is the reset gate vector, $s_t$ indicates the output vector and $W$, $U$ are weight parameter matrices. $\sigma$ is the sigmoid function that outputs a number between $0$ and $1$ where $1$ signifies that all information gets to pass through the cell state, whereas $0$ signifies no information is passed through. However, a $tanh$ function outputs value a between $-1$ and $1$.

In the recent literature, LSTM and GRU have played essential roles in improving our MTP capabilities and understanding. With the emergence of ITS, data generation has increased dramatically, and many traffic time series datasets are highly dimensional (e.g., one dimension for each detector in the network). Adding a memory cell within the LSTM and GRU enables the models to learn the long-term temporal dependencies in the data, even if they are sparsely represented, making them highly applicable to MTP applications. Notably, in recent approaches, LSTM and GRU are seldom used alone; instead, researchers propose unique ways to incorporate them into hybrid designs \citep{dai2020spatio}. 

In \citet{luo2019spatiotemporal} study, a modified KNN-LSTM architecture was designed to optimize the LSTM-based model for traffic flow prediction. Notably, the KNN algorithm extracts the spatio-temporal correlations between time series (sensing stations) within the network by computing the similarity between the measurements at each station. The most novel contribution of the paper is the use of a rank-exponent method for weighting the predictions from the K-nearest stations to produce the final prediction at any test station. Case study results reveal that integrating KNN and weighted rank-exponent methods with the 2-layer LSTM improved prediction performance compared to existing ML and DL approaches, including standalone LSTM. However, the success of the proposed method is heavily dependent on similarity in the magnitude and distribution of time series between different stations. If the traffic stream between stations exhibits high variance, the weighted rank-exponent process could produce inconsistent results when predicting flows for some stations.

During the same year, \citet{zhao2019t} proposed an alternative RNN-based hybrid MTP model by combining GRU with GCN to create the T-GCN architecture. In the novel approach, a T-GCN cell is designed, which performs the graph convolution on input data before feeding it through a GRU-inspired series of gates (e.g., reset and update gates) to produce the final prediction. The GRU was chosen over LSTM due to its similar performance but reduced complexity \citep{chung2014empirical}, having fewer parameters to train. Graph convolution was used to extract the spatial features in favor of euclidean convolution, as graphs have been demonstrated better to represent the transportation network's link and node structure. Notable experimental results showcase that the T-GCN design can outperform the predictions of other popular models, including the standalone GCN and GRU, highlighting the importance of jointly considering the spatio-temporal trends in traffic data. Interestingly, the model predicts poorly at the local minima/maxima, which could be a byproduct of the smooth filters applied during graph convolution. Moreover, the small changes resulting from smoothing are likely incapable of mimicking the harsh and stochastic changes in inter-link dependencies within the transportation network, especially in peak periods.

Another work by \citet{dai2019short} introduced an alternative approach for mining the spatial traffic features for RNN-based models, in contrast to KNN or GCN. The proposed Spatio-temporal feature selection algorithm (STFSA) uses the Pearson correlation coefficient to extract the most relevant traffic features for the current prediction task. The approach is similar to the KNN feature mining in the study of \citet{luo2019spatiotemporal}, except the method for computing similarity between time series at each station differs. STSFA is an exciting contribution to MTP models because it offers a simple and easy-to-interpret method for spatio-temporal feature generation. Notably, the STFSA approach could be used with various RNN-based models to improve feature extraction and capture significant traffic flow trends. The temporal component of STSFA determines the ideal input length by selecting a subset of historical input with a strong correlation to the current period. At the same time, the spatial component selects the six most highly correlated detectors from the set of time series. After running STSFA, the input data matrix is constructed using the optimal subset of input data from each selected time series. Next, the GRU model is used to predict future traffic flow. Compared to standalone GRU and CNN models, the STSFA combined with GRU produced the best predictions regardless of the time horizon (e.g., 5 minutes, 15 minutes). However, the study area for this paper was only a single road segment, and the effectiveness of STSFA on a larger region of the roadway network has not been explored. At the same time, cross-comparison between this approach and the aforementioned competing designs is not meaningful because the studies use different datasets in their experiments.

In summary, RNN models have demonstrated promise for modeling the temporal characteristics of the multivariate traffic time series data. While earlier approaches primarily used LSTM-based RNN architectures in their designs, multiple works have highlighted the competitive performance of GRU-based models and prefer them for their reduced complexity and faster training time. Moreover, the most recent cutting-edge MTP approaches integrate RNN-based designs within more sophisticated, hybrid model architectures designed to consider spatio-temporal traffic trends jointly. The motivation behind applying RNN-based architectures to MTP stems from their strong ability to capture short- (e.g., daily periodicity) and long-term (e.g., seasonality) trends within the time series data. However, recent research has demonstrated that hybrid approaches are becoming necessary to jointly consider the spatial relations between links and nodes within the road network, such as the impacts of heavy congestion in the CBD on the downstream residential district. The spatial relationships also have an inherent temporal component, as the effects of an event at one location of the network need time to propagate, further highlighting the importance of jointly considering the spatio-temporal features. Thankfully, scholars have proposed multiple promising hybrid approaches for enhancing the RNN-based models with spatial capabilities, including nonparametric methods (e.g., KNN), graph-based convolution (e.g., GCN), convolutional RNNs (e.g., Conv-LSTM), and traditional statistics (e.g., Pearson correlation coefficient). That being said, despite each study demonstrating the superiority of their proposed method compared to some existing popularized methods, the datasets used in experiments differ, and the models are not directly cross-compared. Consequently, future research is needed to elucidate if an ideal model architecture exists for MTP applications.

\subsubsection{GCN}

\begin{figure}[t]
    \centering
    \includegraphics[width=.99\textwidth]{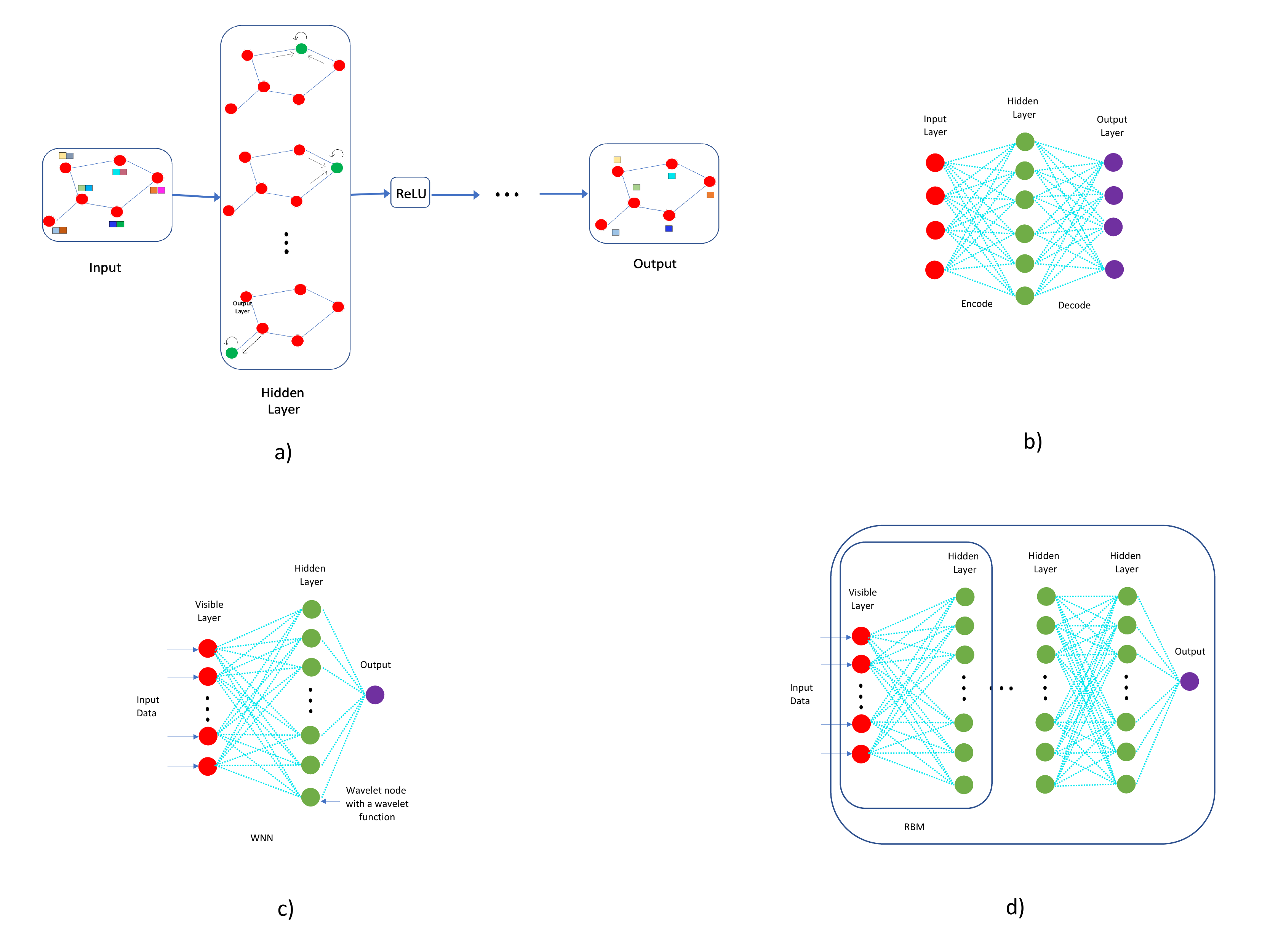}
    \caption{Structure of GCN.}
      \label{fig:Model3}
\end{figure}

The Graph convolution network (GCN) \citep{kipf2016semi} is an evolution of the CNN model designed to perform convolution on structured graphical input, in contrast to the euclidean-structured input data of traditional convolution. As outlined in the CNN section, the convolution operation is one well-researched method for extracting the spatial features of a roadway network in MTP models. As shown in Fig. \ref{fig:Model3}, during convolution, the model learns many filters for detecting interlink relationships (features) concerning the traffic flow or speed, improving prediction performance by accounting for the inherent time-dependencies between related links. However, modeling the roadway network in euclidean space necessitates a grid-based representation, distorting meaningful connections between nodes and limiting the effectiveness of the models. For example, two adjacent roadways may lie very close in euclidean space but be unconnected and exhibit vastly different characteristics, which limits the effectiveness of traditional image-based convolutional approaches for MTP models. Consequently, scientists have begun experimenting with graphical convolution in their MTP models as an alternative to the widely used euclidean convolution.

Prior to the introduction of GCN, \citet{gori2005new} and \citet{scarselli2008graph} identified the lack of a NN architecture capable of processing graphical input, proposing the graph neural network (GNN) as a solution. Succinctly, GNN is a form of RNN that learns a matrix-based representation of the input graph while encoding each node with a set of learned weights relating to the dependencies between directly connected nodes. The intuition behind the approach is that the nodes within the graph represent objects, while the links express the relationship between objects. Interestingly, this design directly maps to the practical interpretation of the node and link structure in roadway networks; however, this method still relies on a euclidean representation of the graph. More recently, \citet{bruna2014spectral} generalized the popular CNN model to operate directly on graph-based analogs as input, combining spectral graph theory with DNNs. The results show that their approach can significantly reduce the required parameters while providing faster forward propagation for input data with location-based features. This approach was later improved by proposing a method for computing fast localized convolutional filters \citep{defferrard2016convolutional} and was further simplified to make GCN practical for modeling high-dimensional time series \citep{kipf2016semi}.

Despite the noticeable applicability of the aforementioned graph-based convolutional models to MTP applications, they were sparsely considered by researchers until the emergence of GCN \citep{kipf2016semi}. The primary contribution of this work is to simplify graph convolutions with a localized first order approximation of the spectral graph convolution approach in \citet{bruna2014spectral}, further reducing the number of parameters and improving learning speed. Specifically, convolution is localized to consider only the \emph{K}-nearest nodes resulting in truncated filters represented as \emph{K}-order polynomials. By truncating the representation, many filters can be stacked sequentially, enabling deeper networks with better approximation ability which can still be trained efficiently. While this seminal work laid the foundation for many future MTP works, the original GCN operates on undirected graphs which cannot represent the complex traffic diffusion in transportation networks (e.g., upstream and downstream traffic dependencies, directed roadways). Moreover, the approximations leveraged in GCN to reduce computational complexity implicitly assumes equal locality which attributes an equal relationship between all edges connected to a node. In practice, this assumption does not hold because roadways are designed with different classifications, capacities, and goals despite being interconnected, limiting the performance of the original GCN for MTP applications.

While the earlier iterations of GCN excelled at capturing the spatial dependencies between interconnected nodes, they failed to jointly consider the temporal features of structured time series data. Prior MTP approaches proposed to mine the spatio-temporal features through embedding convolutional operations within RNN structures \citep{shi2015convolutional}, but are unable to fully exploit the spatial features of the roadway due to a fixed euclidean input representation. Moreover, they are computationally heavy to train due to the iterative propagation inherent to RNNs. To improve upon these drawbacks, the seminal work of \cite{yu2018spatio} proposed the spatio-temporal graph convolutional network (STGCN) to jointly capture the spatio-temporal features without using an RNN cell. STGCN achieves this by introducing the spatio-temporal convolutional block (ST-Conv block), which sandwiches a GCN layer between two gated and fully convolutional temporal layers \citep{gehring2017convolutional} to capture temporal and spatial features, respectively. In this way, graph convolution is combined with 1D convolution to jointly capture the spatial and temporal trends, respectively. Due to the simpler internal structure and lack of dependency constraints within CNN, STGCN achieves parallelization over the input and is faster to train compared to convolutional RNNs. Moreover, the ST-Conv blocks can be stacked sequentially to build deep models and facilitate MTP in large and complex transportation networks. The authors also conduct a case study on the PeMSD7 dataset, demonstrating that STGCN can be trained about 14 times faster while having nearly $95\%$ less parameters compared to the RNN-based alternatives. These unique properties of STGCN laid the foundation for many of the recent advancements in GCN-based MTP models.

In an alternative approach, the diffusion convolutional recurrent network (DCRNN) is proposed to improve the spatio-temporal modeling capabilities of spectral graph convolution. This model generalized the graph convolution approach to operate on both directed and undirected graphs \citep{li2018diffusion}. The key contribution of this work is the proposed diffusion convolution operation, which relates the traffic flow stream to a diffusion process and can effectively capture stochastic and noisy data trends. More specifically, the diffusion process is represented by a bidirectional random walk of the graph, where the stationary distribution of the Markov process is estimated using a weighted combination of a finite number of random graph walks. In contrast to earlier works, the authors consider the directionality of the network graph and propose both a forward and reverse direction diffusion process to capture the influence of upstream and downstream flows on the sensor of interest.

Moreover, to account for the temporal relationships, the authors integrate diffusion convolution into the GRU cell by replacing the traditional matrix multiplication operation with diffusion convolution. The modified GRU layers are then stacked in an encoder-decoder architecture to create DCRNN. Notably, the authors also propose to use a sequence learning approach \citep{sutskever2014sequence} with scheduled sampling \citep{bengio2015scheduled} to improve the model performance for long-term forecasting (e.g., 1 hour in the future) by minimizing error propagation when expanding the prediction horizon. Experimental results measure the benefits of DCRNN by comparing diffusion convolution to the prior spectral graph convolution approach, highlighting that considering the directionality of traffic through diffusion convolution along a directed graph considerably improves the model prediction accuracy.

Subsequent research attempted to address the shortcomings of GCN for MTP by proposing modifications to the original GCN structure. \cite{wu2019graph} extended the capabilities of GCN by proposing Graph WaveNet: a spatial-based graph convolution method combining graph convolution with dilated casual (1D) convolution. Graph Wavenet dynamically learns the adjacency matrix without requiring an explicit and fixed graphical structure, resulting in two primary benefits: (1) nodes interdependence can be updated dynamically as network conditions change; and (2) hidden influences between nodes without direct connections can be elucidated by the model. Moreover, this approach can dynamically incorporate the addition of new time series as the network topology changes (e.g., addition or removal of sensors and links), presenting the opportunity to learn an online MTP model from evolving live traffic data streams. Notably, Graph WaveNet employs stacked dilated 1D causal convolutions, enabling the creation of deep MTP models capable of processing increasingly long traffic sequences without the vanishing or exploding gradient problem inherent to RNNs. Experiments highlight that Graph WaveNet performs better than STGCN and DCRNN on multivariate datasets with more than 200 sensors, and it can train five times faster than DCRNN but is about twice as slow as STGCN. Graph Wavenet also achieves the fastest inference time of all aforementioned approaches because it outputs twelve predictions each run, while the alternatives depend on historical predictions for inference at larger time horizons.

Similarly, \citet{yu2020forecasting} also proposed incorporating a dynamic adjacency matrix into the GCN architecture; however, this approach learns multiple adjacency matrices from the input data corresponding to specific time intervals in the past. Likewise to diffusion convolution, the intuition behind this approach is to allow the model to imitate the natural phenomenon of traffic propagation, where the traffic correlations between any two nodes are time-dependent and propagate based on direct upstream and downstream connections. Most notably, the authors encode additional features into each node, including the capacity and length of each road segment, demonstrating that incorporating prior domain knowledge into the deep learning models can improve performance. 

Since then, \citet{song2020spatial} further improved graphical approaches for MTP by directly considering an often overlooked spatio-temporal relationship between nodes: the impact of neighboring nodes on the future state of their neighbors. The authors define three high-level relationships the models should consider: (1) the influence of a node on its neighbors at the current time-step (spatial dependency); (2) the impact of a node on itself in the future time-step (temporal correlation); and (3) the impact of neighboring nodes on the future state of a given node (spatio-temporal correlation). Previous works considered spatial dependencies and temporal correlations separately. Specifically, the output features from the spatial modeling component (graph convolution) are used as input to a gated temporal unit (e.g., GRU or LSTM design), thereby indirectly accounting for the spatio-temporal dependencies. In the approach by \citet{song2020spatial}, a new model denoted Spatial-Temporal Synchronous Graph Convolutional Networks (STSGCN) is proposed to consider all three relationships for each node directly. One of the primary contributions of this approach is a unique localized spatial-temporal graph construction, where the adjacency matrix is modified to include a self-connection between all nodes in the current period with its future state in the subsequent time-step. In this way, the impact of neighbors on a given node in the future time step can be represented and learned by the model. Compared to \citet{wu2019graph} and other cutting-edge designs, which consider only the first two relationships, STSGCN demonstrated higher prediction performance across four subsets of the PEMS data. Interestingly, this work also demonstrated the effectiveness of using the gated linear unit (GLU) activation function over the conventional ReLU due to its increased number of parameters and ability to better model the resulting complex dependencies.

Most recently, \citet{guo2021hierarchical} proposed a modified GCN-based model, the hierarchical graph convolution network (HGCN), to investigate the benefits of considering higher-level, regional relationships during high-dimensional MTP. Intuitively, the transportation network contains macro and micro layers, where the microlayers represent the low-level node and link connections. The macro layer represents hot traffic regions (e.g., CBDs, churches, schools, and residential areas). Previous DL-based studies ignore the macro layer, representing the transportation network with only the micro-layer and focusing on the localized relationships between neighboring nodes. Notably, this fine-grained network representation and focus on localized convolution to extract spatial features could obfuscate more significant trends, especially when the distance between interdependent regions in the macro layer is considerable. HGCN can simultaneously model trends at the micro and macro layers by representing the region and road networks separately. GCN-based convolution is applied to both graphs, and the resulting regional features are merged with the localized road segment features to generate the final prediction. While an extensive case study verifies that considering both the micro and macro traffic trends can improve prediction performance compared to existing baselines, more research is needed to determine the best macro-level representation for MTP applications. In this initial approach, a simple spectral clustering method is proposed; however, employing more complex and optimized processes for detecting the hot regions in a network could improve performance.

In summation, GCN models have been demonstrated as a practical component in MTP modeling, specifically for extracting the spatial relationships within multivariate time series. The seminal work by \citet{kipf2016semi} proposed the first GCN to generalize the CNN model to operate on arbitrarily structured graphs, in contrast to Euclidean space. For MTP applications, graphs present an intuitive representation of the spatial relations within the transportation network, whereas a grid-based euclidean representation can destroy the complex underlying structure. Early applications of GCN combine its robust spatial feature extraction with fully convolutional or gated RNN-based layers in sequence to jointly capture the spatio-temporal trends. However, some significant trends can be lost due to the sequential design. Furthermore, early attempts leverage a static adjacency matrix that does not map well to the real-time phenomena of traffic propagation, where the relationship between any two road segments is dynamic. Recent works have proposed multiple methods for learning dynamic adjacency matrices, including \citet{song2020spatial} and \citet{wu2019graph}, which improves the prediction performance of GCN-based models. Additionally, it has been demonstrated that considering both the macro and micro network layer relationships during modeling can improve the prediction results, but determining the best macro-level representation of the roadway network is still an open research problem.

\begin{figure}[t]
    \centering
    \includegraphics[width=.5\textwidth]{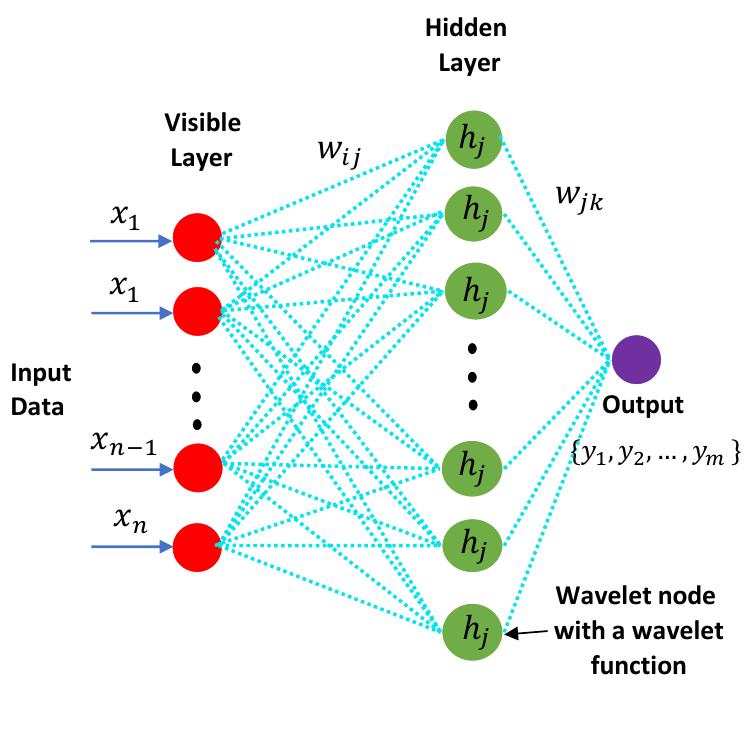}
    \caption{Structure of WNN.}
      \label{fig:WNN}
      \end{figure}

\subsubsection{WNN}
A wavelet neural network (WNN) is an integrated network combining wavelet transformations and ANNs, which was introduced by \citet{zhang1992wavelet} to solve the convergence performance problem. In summary, while the authors could not theoretically guarantee that their approach would converge successfully, they demonstrated significant experimental improvement compared to the old approach (which typically diverged while their approach consistently converged). In general, WNN aims to find a group of wavelets in feature space that can reflect complex relationships within the original signal. WNNs were created as a tool for function approximation, allowing them to solve concerns with traditional NNs' weak convergence. 

As shown in Fig. \ref{fig:WNN}, along with an input and output layer, WNN consists of one hidden layer that utilizes a non-linear wavelet basis function as an activation function. Notably, adding scale and shift factors improves WNN's approximation results and provides better accuracy. The equations defining WNN are expressed below: assume input and output layer has $n$ and $m$ neurons respectively, i.e., $x_1, x_2, ..., x_n$ are inputs and $y_1, y_2, y_3, ..., y_m$ are outputs of NN and $w_{ij}$, $w_{jk}$ are connection weights. With input $x_i (i = 1, 2,....,m)$, the output of hidden layer is given as follows:

\begin{equation}
h(j)= z_j [\frac{\sum_{i=1}^{n} W_{ij} x_i- b_j}{a_j}] \, \, \, \, \, \, \, \, = 1, 2, ..., p
\end{equation}
where $h(j)$ is the output of the $j$th node in the hidden layer, $w_{ij}$ is the weight between the input and hidden layer of the network, $a_j$ is the scaling factor of the wavelet basis function, $b_j$ is the translation factor of the wavelet basis function $z_j$, $z_j$ is the wavelet activation function, and $p$ represents the number of hidden layer nodes. WNN uses a gradient descent algorithm to adjust connection weights, shift and scale factor. Generally, the Morlet Wavelet basis function is selected as the activation function for the hidden layer, which extracts the magnitude and phase information of the analyzed signal as shown in Eq. \ref{WNN2}.

\begin{equation}
\label{WNN2}
y = cos(1.75x)e^{−x^2/2} 
\end{equation}

Finally, the output of WNN can be represented as:

\begin{equation}
y(k) = \sum_{j=i}^{p} W_{jk} h(j) \, \, \, \, \, \, \, \, \,k = 1, 2, ...,m 
\end{equation}

where $w_{jk}$ represents the connection weight of the hidden layer to the output layer, $h(j)$ is the output of the $j$th hidden, and $m$ represents the number of output layer nodes.

Given the inherent similarities between time series and signals, recent research has experimented with the WNN for predicting short-term traffic flow. WNNs can be a desirable architecture for MTP applications because of their strong nonlinear processing power, self-organization, self-adaptation, and learning ability. However, the choice of parameter optimization algorithm during training is a crucial consideration. Like many DL models, the traditional WNN structure uses a stochastic gradient descent algorithm for learning the weights. This method is easily trapped in the local extremum, resulting in slow convergence and low prediction accuracy. To address these challenges for flow prediction in ITS, \citet{chen2020short} proposed an improved particle swarm optimization (I-PSO) algorithm to address the slow convergence rate and the local optimality problems of stochastic gradient descent. Compared to the standard WNN, or WNN with traditional PSO, WNN with I-PSO can achieve the highest accuracy while converging the fastest. However, the authors note that the proposed approach is ineffective during non-recurrent network events and does not consider external factors such as weather, presenting much room for future research.

Parallel research by \citet{shen2020short} applied the harmony search algorithm to WNN, denoted HS-WNN, as an alternative optimization method for learning the WNN weights. While particle swarm optimization imitates the social behavior of flocking or herding biological species (e.g., fish, birds), the harmony search optimization heuristic is inspired by the improvisation process of jazz musicians. Notably, the case studies in both papers demonstrate the improvements of alternative optimization approaches on the convergence speed and optimal solution reliability, compared to stochastic gradient descent, for WNN-based MTP models. However, the datasets used in both papers are poorly documented, making cross-comparison difficult and leaving the readers wondering which optimization approach is the best for traffic prediction.

\subsubsection{Attention-based}
Attention-based methods for DL first appeared in machine translation, where they were demonstrated as an effective measure for overcoming degraded performance in AEs when working with long input sequences \citep{bahdanau2015neural}. With the emergence of ITS and big data processing, transportation networks are generating a vast amount of time series data across a diverse set of geospatial sensing locations within the network. However, many of the cutting-edge short-term MTP approaches focus only on the near-term (e.g., hourly, daily, or weekly) dependencies due to limitations with the existing architectures. For example, GCN only considers relationships between the \emph{K}-nearest nodes, while LSTM and GRU suffer from vanishing gradient problems when dealing with longer sequences. Attention-based mechanisms are an exciting method for overcoming these challenges by enabling the model to learn a set of high-impact features indicating the relative importance of each subset of input for the current prediction interval.

Recently, attention-based methods have been applied to various MTP models to improve their spatio-temporal feature extraction capabilities and learning efficiency. \citet{wu2020combined} proposed integrating an attention mechanism within a bi-directional LSTM architecture for improving MTP for high-dimensional time series. In this approach, the attention mechanism can dynamically learn the relative importance of each adjacent time series on the series to be predicted, capturing the most impactful spatio-temporal dependencies. Wavelet decomposition is also used to decompose each time series into a low-frequency and noise series, where the low-frequency series represents the important high-level trends with the stochastic noise removed. Focusing on the input's low-frequency representation makes learning the high-dimensional time series traffic data more efficient. At the same time, the prediction accuracy is preserved because the attention mechanism dynamically weights the time series in order of importance.

Complementary work around the same time was performed by \citet{liu2019traffic}, where attention was integrated into two bi-directional LSTM models to capture the spatial and temporal traffic features in high-dimensional time series data. A unique contribution of this work is the proposed preprocessing module, which can cluster the time series data into groups with similar distributions before model training, improving the convergence time and prediction accuracy. Compared with \citet{wu2020combined}, the case studies in both papers demonstrate the benefits of incorporating an attention mechanism into LSTM-based MTP models, notably for high-dimensional time series modeling. In this way, the model can learn new hyperparameters relating the importance of input segments to the current prediction task, in contrast to condensing input into a fixed-length representation regardless of size. That being said, the papers conduct experiments using different datasets and study areas, so cross-comparison becomes challenging.

While the previous works focus on integrating attention mechanisms into RNN-based designs, \citet{guo2019attention} proposed an attention-based convolutional model (ASTGCN), which employs both graph-based and standard convolution with attention mechanisms for capturing the spatio-temporal traffic trends. The authors improve upon prior approaches to dynamically learning the GCN adjacency matrix, combining an attention mechanism with the GCN. More specifically, the attention method learns a separate hyperparameter matrix indicating the relative importance of each node to the current prediction and time, which is used to weight the adjacency matrix dynamically. Consequently, the latent dynamic and time-dependent relationships between time series are effectively considered. An extensive case study demonstrates the effectiveness of this approach for dynamically assessing the temporal traffic dependencies within the GCN framework.

More recently, \citet{shi2020spatial} leveraged attention mechanisms to further improve our understanding of high-dimensional MTP modeling by integrating attention and a modified LSTM design to overcome the vanishing gradient problem for long-range time series modeling. Notably, a recurrent structure with skip connections is proposed to alter the traditional LSTM design and capture pair-wise long-range dependencies between distant time series without a vanishing gradient. A spatial attention mechanism in the encoder learns the impact of each node on the entire graph, giving the most consideration to the small subset of nodes with the most significant effect on the current time-step. Moreover, the proposed approach is directly compared to \citet{guo2019attention}, demonstrating that the skip-connections with attention can effectively consider the long-range and pair-wise dependencies better than a GCN-based architecture. However, the proposed model has almost twice as many parameters as ASTGCN, requiring more data and processing power to achieve better results.

In an alternative approach to improving the reliability of long-term MTP, \cite{zheng2020gman} proposed the graph multi-attention network (GMAN) by integrating attention mechanisms into an encoder-decoder architecture. One of the most notable contributions of GMAN is the transform attention mechanism, which is an attention layer between the encoder and decoder. It is responsible for modeling the direct relationship between the future and previous time steps, enabling the model to consider only the most important historical features while limiting error propagation. In other words, it transforms the encoder output using an attention mechanism prior to feeding the output into the decoder. Moreover, GMAN stacks multiple spatio-temporal attention blocks within the encoder and decoder, forming a deep network capable of extracting the traffic trends in large sensor networks. Within each spatio-temporal block, the spatial attention mechanism dynamically updates the pair-wise correlation between vertices with respect to the current time step, while the temporal attention mechanism dynamically weights the impact of previous and future time steps on the current step. Notably, the input data must be first transformed via a spatio-temporal embedding process to incorporate the time-dependent graph data into the attention mechanisms. Experimental results highlight the benefits of the transform attention mechanism, with GMAN outperforming both DCRNN and Graph Wavenet for prediction horizons of 30 and 60 minutes. However, GMAN performed worse than Graph Wavenet for shorter (15 minutes) predictions while requiring more time for training and inference.

In summary, DL models are widely applicable to MTP applications due to their superior ability to capture stochastic and non-linear relationships in traffic data. Recent research has proposed many efficient hierarchical feature extraction approaches, including spectral and spatial graph-based convolution methods, attention mechanisms, diffusion convolution, dilated convolution, and RNN-based methods. The performance of each approach is comparable in terms of prediction accuracy at different time horizons and computational complexity, and both must be considered when selecting the ideal modeling approach. In Table \ref{modelDL}, we compare the above-discussed models for reference. In the following subsection, we review the limitations of Deep Learning-based methods.

\begin{landscape}

\begin{table}[h]
\centering
\begin{adjustbox}{width=1.4\textwidth}
\begin{threeparttable}
\caption{Comparison of DL architectures.}
\label{modelDL}
  \begin{tabular}{cllllll}
\hline
\textbf{Model}&\textbf{Description}&\textbf{Advantage}&\textbf{Disadvantage}&\textbf{Example}\\\hline

\multirow{4}{*}{\textbf{MLP}}&
\textendash a simple ANN& \textendash Simpler and more interpretable than  &\textendash time series data requires much  &\\
&\textendash Consists of three layers:input layer, layer to&mother methods&preprocessing&\multirow{2}{*}{\cite{song2017traffic}}\\
&features from the input, output
layer&\textendash Slow
convergence&\textendash Outperformed by cutting-edge DNN 
\\
&\textendash Layers are composed of several neurons&\textendash Occupies a
large amount of memory&designs\\\hline

\multirow{6}{*}{\textbf{AE}}&\textendash Consists of three parts:encoder (converts the&\multirow{4}{*}{\textendash Extracts the most useful and high-level}  &\multirow{4}{*}{\textendash Multiple forward passes create a}  &\multirow{3}{*}{\citet{ghosh2017denoising}}\\
&input features into an abstraction (code))&\multirow{4}{*}{features}&\multirow{4}{*}{long training and preprocessing time}&\multirow{3}{*}{\citet{wang2018network}}\\
&, code $\&$ decoder blocks ( Try
to reconstruct 
&\multirow{4}{*}{\textendash Works with big and
unlabeled data}&\multirow{4}{*}{\textendash Hard to find the code
layer size}&\multirow{3}{*}{\citet{liu2019traffic}}\\
&the input features)&&&\multirow{3}{*}{\citet{wang2020scalability}}\\
&\textendash Use some non-linear hidden layers to reduce\\
&the
input features\\\hline

\multirow{4}{*}{\textbf{RBM$\&$DBN}}& \textendash DBN is stacked by several RBN
&\multirow{2}{*}{\textendash Effectively extracts representative} &\textendash Difficulty mapping complex  &\multirow{3}{*}{\citet{ma2015large}} \\
&\textendash Take advantage of the greedy
learning process to&\multirow{2}{*}{ data characteristics}&relationships  between network& \multirow{3}{*}{\citet{bao2021improved}}\\
&initialize the model parameters
and then &\multirow{2}{*}{\textendash DBN:  fast inference time}& elements&\\
&fine-tunes the whole model using
the label&&\textendash High computational cost&\\\hline

\multirow{5}{*}{\textbf{CNN}}& \textendash  Consists of a number
of convolution 
&\multirow{2}{*}{\textendash    Adept at capturing spatial depen-}&\multirow{3}{*}{\textendash  Long training time is needed}&\citet{lea2016temporal,zhang2017deep}; \\

 &and pooling layers (to reduce the
&\multirow{2}{*}{dencies}&\multirow{3}{*}{\textendash Sequence and time series data} &\citet{song2017traffic,wu2018hybrid};\\
&dimensions of features and find useful &\multirow{2}{*}{\textendash Extracts
relevant features}&\multirow{2}{*}{ are not supported }&\citet{wu2018hybrid,liu2019deeprtp};\\
&patterns)
followed by a fully connected&\multirow{2}{*}{\textendash Have high
competitive performance}&&\citet{sen2019think,liu2019deeprtp}; \\
&layers (used for
classification.)&&&\citet{zheng2020hybrid,dai2020spatio}\\\hline

\multirow{4}{*}{\textbf{RNN}}&
 \textendash Consists of numerous successive recurrent&\multirow{3}{*}{\textendash   Strong performance on time series}&\textendash   Inability to consider spatial  & \multirow{3}{*}{\citet{wang2018learning}}\\
&layers, which are sequentially modeled in &\multirow{3}{*}{ data}&dependencies &\multirow{3}{*}{\citet{zhang2017deep}}\\
&order to map the sequence with other &  &\textendash Exploding and vanishing gradient\\
&sequences&&problem\\\hline

\multirow{6}{*}{\textbf{LSTM}}&\textendash An extension of RNN&\multirow{4}{*}{\textendash Memory cell enables better}&\multirow{4}{*}{\textendash Longest training time}&\multirow{2}{*}{\citet{luo2019spatiotemporal}}\\
&\textendash Created as the solution to
short-term memory&\multirow{4}{*}{performance on long sequences}&\multirow{4}{*}{\textendash High model complexity}&\multirow{2}{*}{\citet{fu2016using}}\\
&\textendash Has internal mechanisms
called gates&\multirow{4}{*}{\textendash Good for sequential data}&\multirow{4}{*}{\textendash High computational cost}&\multirow{2}{*}{\citet{wu2020combined}}
 \\
&(forget gate, input gate, and output
gate)&&&\multirow{2}{*}{\citet{liu2019traffic}}\\
&to learn which data in a sequence is
important&&&\multirow{2}{*}{\citet{li2021hybrid}} \\
&to keep or throw away\\\hline

\multirow{3}{*}{\textbf{GRU}}&\multirow{2}{*}{\textendash Similar to LSTM
but has fewer parameters} &\textendash Similar performance but faster&\multirow{2}{*}{\textendash Less efficient in accuracy
than}&\multirow{2}{*}{\citet{zhao2019t}} \\
&\multirow{2}{*}{\textendash Has two gates( update
gate $\&$ reset gate)}&training time  compared to LSTM&\multirow{2}{*}{LSTM}&\multirow{2}{*}{\citet{dai2019short}}  \\
&&\textendash Good for sequential data\\\hline

\multirow{4}{*}{\textbf{GCN}}&
\multirow{2}{*}{\textendash A spectral-based convolutional GNNs} &\textendash   Exploits graph-based structure of &\multirow{3}{*}{\textendash Relies on human-defined topological} &\citet{wu2019graph}\\
&
\multirow{2}{*}{\textendash Defined by introducing filters from graph } &transportation networks &\multirow{3}{*}{structure}&\citet{yu2020forecasting} \\
&\multirow{2}{*}{signal processing}&\textendash Effectively captures complex spatial &&
 \citet{song2020spatial}\\
&&dependencies in large networks&&\citet{guo2021hierarchical}\\\hline

\multirow{7}{*}{\textbf{Attention}}&
\textendash Consists of an
encoder network (to map input &\multirow{2}{*}{\textendash   Provide strong results in presence of} &\multirow{4}{*}{\textendash Increase training time of long input  }&\multirow{3}{*}{\citet{bahdanau2015neural}}\\
&acoustic vectors into a higherlevel representation), &\multirow{2}{*}{ long input sequences}&\multirow{4}{*}{sequences data by adding more weight  } &\multirow{3}{*}{\citet{wu2020combined}}\\
&an attention model( to summarize the output of &\multirow{2}{*}{\textendash More interpretable than the other DL} &\multirow{4}{*}{parameters to the model}&\multirow{3}{*}{\citet{guo2019attention}} \\
&the encoder based on the current state of the &\multirow{2}{*}{ models because of the attention weights}&&\multirow{3}{*}{\citet{liu2019traffic}}\\
&decoder), $\&$ a decoder network( to model an &\multirow{2}{*}{\textendash  Demonstrates outstanding results in} 
&&\multirow{3}{*}{\citet{shi2020spatial}}  \\
&output distribution over
the next target cond-&\multirow{2}{*}{NLP models}&\\
&itioned on the sequence of previous predictions)\\\hline

\multirow{3}{*}{\textbf{WNN}}&\textendash Consist of one hidden layer that utilized a &\textendash Time Frequency localization of wavelet &\multirow{2}{*}{\textendash Can be bias towards local minima} &\multirow{2}{*}{\cite{chen2020short}}\\ 
&non-linear wavelet basis function as an&\textendash Self-learning&\multirow{2}{*}{and maxima}&\multirow{2}{*}{\cite{shen2020short}}\\
&activation function&\textendash Fault-tolerance\\\hline

\hline
\end{tabular}
  \end{threeparttable}
  \end{adjustbox}
\end{table} 
\end{landscape}

\subsubsection{Limitations of the Deep Learning-based Methods}

In comparison with classical methods, the literature highlights many advantages of leveraging deep learning models for MTP applications. Despite this, the emerging deep learning approaches have some notable limitations:

\begin{itemize}

\item \textbf {Computational complexity and efficiency}. Compared to other machine learning methods, deep learning models have increasingly complex data structures and require higher computing power for training. Recently, researchers have expressed concerns over the high resource requirements of cutting-edge deep learning models across various application areas. To better understand and quantify the computational limits of existing deep learning methods, \cite{thompson2020computational} examined 1058 articles on deep learning methods across five application areas (image classification, object detection, questioning, named entity recognition, and machine translation). They identified a clear trend in the literature, highlighting that advances in accuracy heavily rely on simultaneous advances in computing power (p-value $<$ 0.01). Specifically, the authors leverage statistical learning theory to estimate that computational requirements scale as a fourth-order polynomial, at a minimum, concerning performance (but practically, the scaling can be demonstrated to be worse). One explanation for this relationship is that NNs overparameterize the network by design because it has been shown that performance is significantly improved when NNs have more parameters than available training instances \citep{soltanolkotabi2018theoretical}. For example, one of the best-performing image recognition NNs, NoisyStudent, has 480 million parameters trained on only 1.2 million data points. Notably, the authors conclude that the current trajectory of deep learning advances will quickly become economically, technically, and environmentally unsustainable due to the relationship between advancement and computation requirements.

On the bright side, DL for traffic prediction has yet to reach the same computational ceiling as the application areas studied in \cite{thompson2020computational}. There is a need for re-examining the existing DL-based approaches for traffic prediction from a scalability perspective. The statistical learning analysis highlights that there is much room for improving the efficiency of the current methods by reducing computational complexity and developing high-performing but lightweight deep learning architectures. In recent works, some researchers have attempted to improve the efficiency of NNs by leveraging optimization to search for the hyperparameter set, which maximizes accuracy while minimizing computation \citep{pham2018efficient}. On the other hand, another group of approaches, denoted model pruning strategies, seeks to compress an already trained network and reduce inference computation costs while maintaining good performance \citep{deng2020model}, but this does little to improve efficiency from a model training perspective. Notably, if the pruned model representations can be leveraged to train new models with sparser parameter counts, the training efficiency can also potentially be increased. Lastly, exploring alternative model architectures to DL for solving traffic prediction problems may be necessary if increases in efficiency are not achievable. In summation, computation requirements for improving DL approaches are rapidly outpacing advances in hardware or efficiency, highlighting a need for further research to enhance DL efficiency.

\item \textbf{Optimal model choice}. For MTP applications, the ideal model can capture the three critical trends: (1) the influence of a node on its neighbors at the current time-step (spatial dependency); (2) the impact of a node on itself in the future time-step (temporal correlation); and (3) the impact of neighboring nodes on the future state of a given node (spatio-temporal correlation) \citep{song2020spatial}. While the appropriate model selection always involves application- and data-specific considerations, the literature is mainly undecided on the best approach for multivariate speed or flow forecasting, and new techniques are frequently proposed. Moreover, in addition to the trends above, the ideal model should also capture both the short- and long-term dependencies in the spatial (e.g., dependencies between distant but related nodes such as CBD and residential) and temporal (e.g., recent, daily-periodic, seasonal) areas. Existing works have revealed specific architectures optimized for accomplishing a subset of the modeling goals, but no unified architecture has emerged as the best for all considerations. Furthermore, after deciding on your model architecture, determining the best model parameters (including how many hidden layers, the number of hidden nodes, desired learning rate, hyperparameters, choice of activation function, evaluation method, etc.) is challenging and generally requires a brute-force approach. Encouragingly, some recent research has proposed and implemented optimization methods that search for the ideal hyperparameter set given dynamic application requirements \citep{balaprakash2018deephyper, andonie2019hyperparameter}.

\item \textbf{Lack of interpretability}. DL algorithms are generally considered ''black box" models with limited practical interpretability compared to traditional approaches. However, traffic forecasting models are leveraged in many high-stakes decision-making applications (e.g., signal timing and planning, route guidance), where interpretability of the resulting predictions is key to understanding why specific approaches perform better or worse than others. At the same time, many DL algorithms lack the ability to quantify the uncertainty of their prediction output due to the dynamic nature of the prediction errors and resulting uncertainty when modeling stochastic and noisy traffic data. Under these dynamic conditions, the conventional approaches for measuring uncertainty (e.g., Kalman or Bayes filters) may fail due to the assumption of a fixed covariance matrix. Moreover, models in high-stakes applications are generally regarded as useless unless their decision-making process can be verified by a domain expert \citep{yu2020forecasting}. Consequently, the full potential of DL approaches is not realized in critical decision-making systems due to the potential for catastrophic failures. Recent research has proposed new methods for adapting DL models to learn a deep uncertainty covariance matrix dynamically while also considering both the epistemic (uncertainty of model parameters) and aleatoric (noise within the data) uncertainties \citep{russell2021multivariate}. However, the work makes many assumptions, including assuming the aleatoric and epistemic uncertainties follow a multivariate Gaussian distribution, which may not hold for highly nonlinear systems where error distributions are multimodal. Moreover, the measurement uncertainty is assumed to be uncorrelated in time between measurements. Both assumptions limit the applicability of the proposed approach for MTP models, where the inherently stochastic data may impair the performance, necessitating further research.

In the traffic prediction domain, some studies have also proposed variations of DL architectures that focus on improving the interpretability of the prediction results. For example, \citet{wang2019traffic} enhanced the interpretability of traffic speed forecasting by proposing a bi-directional and path-based LSTM model for urban network speed prediction. Precisely, the deep learning network is mapped to the physical roadway network resulting in interpretable output from the hidden layer using visualization and qualitative analysis. However, more research is needed to explain better how the newly interpretable information can elucidate insights that drive future improvements. In another related work, \citet{wang2020interpretable} introduced an integrated approach for short-term traffic flow prediction combining SARIMA and Group Method of Data Handling (GMDH) to improve our understanding of the causality between the historical spatio-temporal traffic information and future conditions. Their interpretability analysis concluded that the causality could be interpreted as a polynomial function of historical temporal data and spatial traffic information. The approach demonstrates high prediction accuracy compared to existing and less interpretable methods. That being said, the generalizability of this approach has yet to be explored in other traffic prediction applications. Further research is needed to improve the interpretability of DL-based systems and elucidate methods for leveraging the new insights to enhance the models' predictive capabilities further.

\item \textbf{Limited transferability.} The cutting-edge traffic forecasting approaches are generally network-specific, rendering them useless for predicting traffic outside the specific context in which they were trained. More specifically, the spatial relations learned by the model depend on the topological structure of the sensor network. In other words, two urban road networks are unlikely to be highly similar in structure. It is not efficient to train a network-specific prediction model for every transportation network. Thus, methods for improving model generalizability (transferability) interest the research community greatly. Recently, transfer learning is an emerging ML-based research area to apply knowledge gained while training one model to speed up the training time of a new model. For example, traffic prediction transfer learning can leverage a Seattle-based traffic flow model to help train a flow prediction model for the Boston transportation network. However, transfer learning is a new research area, and little literature exists on transfer learning approaches for traffic prediction problems.

In one very recent work, \citet{mallick2021transfer} proposed a transfer learning approach for diffusion convolution recurrent neural networks (DCRNN), one of the state-of-the-art model architectures for highway traffic forecasting. DCRNN cannot perform transfer learning in its traditional form because it inherently learns location-specific traffic patterns. In the proposed approach, traffic is realized as a function of network connectivity and temporal patterns, enabling generalization to unseen networks with varying spatial representations. The experimental results demonstrate that the approach can successfully perform transfer learning between the Los Angeles and San Francisco transportation networks. However, the study focuses only on highway networks and does not consider external data sources in its analysis. In an alternative approach to cross-city transfer learning, RegionTrans is proposed to effectively transfer knowledge learned from a data-rich city to a data-scarce city \citep{wang2019cross}. Notably, this approach relies on matching a data-scarce city with a data-rich city exhibiting a similar network topology, limiting practical applicability. While these recent works demonstrate progress in applying transfer learning to traffic prediction problems, additional research is needed to refine the approaches and analyze their benefits compared to existing ones.
\end{itemize}

\subsubsection{Deep Learning Evaluation Methods}

\renewcommand{\arraystretch}{1.3}

\begin{table}[!t]
\tiny
\centering
\begin{threeparttable}
\caption{ Deep Learning Evaluation Methods}
\label{Eval}

\begin{tabular}{lll}
\hline
{\bf Evaluation Parameter}&{\bf Formula}&{\bf Characteristics}\\\hline\xrowht[()]{15pt}
Mean Absolute Error (MAE)&$\frac{1}{n}\sum_{i=1}^{n} |\hat{y}_i - y_i| $&\\\cline{1-2}\xrowht[()]{15pt}
Root Mean Square Error (RMSE)&$[\frac{1}{n}\sum_{i=1}^{n} (\hat{y}_i - y_i)^2 ]^{\frac{1}{2}}$&\\\cline{1-2}\xrowht[()]{15pt}
Mean Square Error (MSE)&$\frac{1}{n}\sum_{i=1}^{n} (\hat{y}_i - y_i)^2 $&\\\cline{1-2}\xrowht[()]{15pt}
Mean Relative Error (MRE)&$\frac{1}{n}\sum_{i=1}^{n} \frac{|\hat{y}_i - y_i|}{y_i} $&\\\cline{1-2}

\multirow{2}{*}{Absolute Percentage Error (APE)}&\multirow{2}{*}{$\frac{\hat{y}_i - y}{y_t}$}&\textendash Measure the prediction error\\
&&\textendash The smaller the value is, the\\\cline{1-2}
Variance Absolute Percentage Error (VAPE)&$Var(\frac{\hat{y}_i - y}{y})\times100\%$& better the prediction effect\\\cline{1-2}\xrowht[()]{15pt}
Normalized Root Mean Square Error &\multirow{2}{*}{$\frac{[\frac{1}{n}\sum_{i=1}^{n} (\hat{y}_i - y_i)^2 ]^{\frac{1}{2}}}{max(y_i)-min(y_i)}$}\\
(NRMSE)&&\\\cline{1-2}\xrowht[()]{15pt}
Mean Absolute Percentage Error (MAPE)&$\frac{100\%}{n}\sum_{i=1}^{n} |\frac{\hat{y}_i - y_i}{y_i}| $\\\cline{1-2}
  Symmetric Mean Absolute Percent  Error &\multirow{2}{*}{$\frac{2}{n}\sum_{i=1}^{n} \frac{|\hat{y}_i - y_i|}{\hat{y}_i + y_i} * 100\%$}&\\
(SMAPE) &&\\\hline
  
\multirow{4}{*}{Coefficient of Determination ($R^2$)}&\multirow{4}{*}{$1-\frac{\sum_{i=1}^{n} (\hat{y}_i - y_i)^2}{\sum_{i=1}^{n} (\hat{y}_i - \bar{Y})^2}$}&\textendash Calculate the correlation coefficient, \\
&&which measures the ability of the  \\
&&predicted result to represent the\\
&&actual data\\\cline{1-2}
\multirow{2}{*}{Explained Variance Score ($v$ar)}&\multirow{2}{*}{$1-\frac{Var(Y-\hat{Y})}{Var(Y)}$}&\textendash The larger the value is, the better\\
&&the prediction effect\\\hline

\multirow{3}{*}{Accuracy}&\multirow{3}{*}{$1-\frac{||Y-\hat{Y}||_F}{||Y||_F}$}&\textendash Detect the prediction precision\\
&&\textendash The lager the value is, the better \\
&& the prediction\\\hline

\end{tabular}
\begin{tablenotes}
      \small
      \item $\hat{y}_i$: The $i$th predicted value, $y_i$: The $i$th target output, $\hat{Y}$: Set of $\hat{y}_i$, $Y$: Set of $y_i$, $\bar{Y}$ : Average of $Y$, n: the total number of samples. 

    \end{tablenotes}
  \end{threeparttable}
\end{table} 

Deep learning models can leverage various parameters to evaluate their traffic prediction accuracy, and the most notable are summarized in Table \ref{Eval}. These parameters generally represent different methods for calculating the prediction accuracy or error. Among these methods, MSE, MAPE, and RMSE are the most commonly used in the literature and are negatively correlated with the training data size. Specifically, as more instances are added to the training set, the model prediction error should decrease, resulting in more accurate prediction capabilities. 

That being said, it is also essential to consider the problems of overfitting and underfitting when deciding the training set size and desired number of features. At this moment, more research is needed to elucidate the overfitting problem within the context of deep learning models, which challenge our conventional understanding due to their strong generalization ability even when they appear to be overfitting the training data. Until recently, the community assumed that a model with an increasing number of parameters and training data would lead to poor generalization performance on testing data; however, new research into deep learning highlights that this is not always the case in practice \citep{belkin2019reconciling}.

\section{Common Traffic Prediction States and Associated Applications} 

\label{Sec5}
Traffic prediction approaches focus on forecasting network-based metrics (e.g., flows, speeds, travel time, demand) at varying scopes based on the dynamic spatio-temporal network conditions. In addition to the previous model-focused literature discussion, the existing traffic prediction methods can also be categorized based on the prediction problem, namely, what is being forecasted. In this section, we outline and describe some of the literature's most frequent and relevant classes of prediction problems. We also categorize recent traffic prediction studies according to their respective application task, area, data type, the model used, and other descriptive dimensions in Table \ref{trafficlit}.

\subsection{Multivariate Traffic Flow and Speed Forecasting}
The traffic forecasting problem represents a spatio-temporal time series prediction problem, where the input contains a traffic variable (e.g., flow or speed) represented in one or more time series, and the output is a forecast of the future conditions. For the most part, multivariate and short-term traffic forecasting literature focuses on the microscopic level by analyzing and predicting traffic conditions (e.g., flow or speed) at low resolutions, such as at the detector or link levels. However, a growing body of literature has shifted its focus to predicting traffic flows at the regional level. In regional flow prediction, the traffic network is segmented into a collection of regions, and the higher-level features and relationships between areas (e.g., residential and CBD) are considered. Notably, this can provide benefits of reduced computation overhead compared to predicting regional traffic flow using a combination of fine-grained models.

\subsubsection{Regional Flow}
In practice, regional flow prediction models are paramount to the application of urban planning. Stakeholders rely on high-level regional insights about the transportation network to plan for future improvements. For example, predicting the traffic flow at entrances to a regional expressway system could be helpful for planners seeking to ensure vehicles enter the system safely and effectively \citep{gao2021synchronized}. Two primary studies attempt to identify the impacts of weather on regional traffic flow concerning region-specific factors (e.g., age, income, road density, hotel density, attraction density) \citep{ding2015dissecting, ding2017detecting}. In particular, inclement weather was demonstrated to have a considerable impact on traffic. At the same time, region-specific features, such as age, also are shown to influence the travel behavior within a region during inclement weather. However, this work is still in its infancy, and the authors express the need to consider the impact of additional external factors on the region-specific traffic flow in future work.

Building upon this research, \citet{liu2019deeprtp} presented the first work to predict traffic congestion at the region level. The proposed approach employs a CNN to extract the spatial features from the traffic data. After that, a novel Traffic State Index (TSI) metric is proposed to measure regional traffic conditions and classify traffic data into three categories. These are then used to capture hourly, daily, and weekly traffic patterns in three residual NNs. Lastly, the outputs of the NNs are carefully synthesized to produce the region-level traffic flow prediction. The authors validate their method with a comparative analysis against five alternative methods, highlighting its superiority. However, the prediction accuracy degrades as the number of regions within the partitioned network increases, presenting the need for future research into deeper and more complex NNs for modeling regional traffic flow. For future work, clustering regions with similar properties has the potential to help overcome the aforementioned challenge in more extensive networks. Still, this approach has yet to be explored for regional flow prediction. Interestingly, related work on taxi demand prediction proposes considering the heterogeneity between different regions in their analysis, which could serve as a basis for future work in regional flow prediction \citep{zhang2021mlrnn}.

\subsubsection{Network Flow}
In contrast to analyzing traffic at the region level, estimating and attributing traffic flow through each roadway and intersection within a transportation network is the focus of the network flow forecasting state. With the emergence of ITS, network flow models can enable more proactive and intelligent traffic management while providing input for real-time route guidance and advanced traveler information systems. However, forecasting network flow is an exceedingly challenging problem due to the intricate spatial relationships between links in the roadway network, the consideration of intersections and traffic control devices, and the volatility and uncertainty of vehicle flow over time. Two primary concerns for network flow forecasting that impact modeling challenges are the scope of the study area and the prediction horizon. Increasing the prediction horizon or expanding the study area to forecast traffic flow on an increased number of links requires more historical data, elevates computational requirements, and demands more complex modeling approaches.

Accordingly, the earlier approaches leveraged statistical ARIMA-based models for short-term network flow prediction across only a single arterial roadway \citep{kumar2015short, kumar2017traffic} or a small subset of the urban network (e.g., 30 links) \citep{cai2016spatiotemporal}. At the time of publication, these approaches outperformed the existing baselines but also presented some drawbacks to the ARIMA approach. Specifically, ARIMA-based methods struggle to handle complex, non-stationary time series data, instead assuming the data is stationary (ideal stationary assumption). Consequently, the capability of ARIMA models to predict traffic during periods where the underlying relationships are more complex, such as during peak periods, is limited.

In an attempt to improve upon these drawbacks, more recent research has leveraged DL-based modeling approaches for predicting network flow at greater scopes. \cite{wu2018hybrid} proposed a novel hybrid model, combining fully-connected NNs, RNNs, and CNNs, to jointly mine the underlying spatio-temporal features from the traffic data and predict network flow within a 45-minute time horizon. The scope of this study included 33 detectors along a major interstate highway. In an alternative approach, \citet{ma2020hybrid} introduced a combination of statistical (ARIMA) with deep learning (MLP) models to predict traffic flows across a single arterial highway with data from 44 double-loop detectors. Although both studies present improved prediction results compared to the existing baselines, their study area's scope raises questions about generalizability: it is unknown whether these approaches will perform well in large-scale urban transportation networks.

To improve large-scale network traffic flow prediction, \citet{zhang2019trafficgan} proposed integrating an attention-based mechanism in conjunction with the CNN and RNN networks to predict traffic flow under an adversarial learning framework with a study area consisting of 1,250 road segments. Despite the improved performance of this approach, leveraging a CNN to extract spatial features from the roadway requires sampling traffic data in regular grids for input, destroying the underlying spatial structure of the network. Most recently, researchers have experimented with alternative methods for spatial feature extraction in traffic flow prediction approaches, including graph-based models. \citet{zhao2019t} integrated GCN and GRU for network flow prediction and conducted a case study using data from 156 major roads and 207 sensors across two urban transportation networks. The approach outperformed five existing baselines (HA, ARIMA, SVR, GCN, GRU). It demonstrated its superior ability to capture the relationship between network topology characteristics concerning the dynamic temporal traffic state. That being said, the performance analysis highlights that this approach performs poorly near local minima/maxima, presenting room for future research.

Later, \citet{guo2020optimized} identified that existing graph-based methods learn the spatial representation using simple, empirical spatial graphs in contrast to a data driven approach, which may hinder performance. To explore the benefits of building the graph in a data-driven way, the authors propose an optimized graph convolution recurrent neural network, which develops an optimized graph that identifies the latent association between road segments directly arising from traffic data. In their performance analysis consisting of data from 370 roadway segments, the authors prove their method outperforms existing baselines for prediction horizons of 15, 30, and 120 minutes ahead; however, it underperforms when predicting 60 minutes ahead. Due to the black-box nature and limited interpretability of DL models, elucidating the model's shortcomings in this scenario is challenging, requiring further research.

\subsubsection{Speed}
While some studies focus on predicting traffic flow, related approaches focus on speed instead, and the literature contains a diverse set of methods for predicting traffic speed. Practically, traffic speed prediction and analysis can aid in dynamic congestion detection and mitigation applications. More specifically, speed information at the segment level, when compared to the predicted (expected) speed, can potentially reveal problematic congestion in real-time and inform the mitigation actions of TMCs. While most of the early studies only considered speed with respect to time \citep{vanajakshi2004comparison,guo2010real}, there is a recent ongoing effort to incorporate the underlying spatial dependencies into the models \citep{dai2020spatio, zhou2022comprehensive}. 

However, despite the rich body of literature on traffic speed prediction, most existing studies have limited scopes and focus on forecasting speed for a particular roadway topology by learning from historical data collected from only a few specific roads (e.g., PEMS data). Furthermore, few studies focus on speed prediction in this network-wide context due to the increased complexity of modeling urban road networks. To fill this research gap, \citet{liu2019traffic} introduced a novel deep learning architecture, leveraging a temporal clustering and hierarchical attention (TCHA) mechanism to capture the latent spatio-temporal correlations in traffic data dynamically. Notably, this method elucidated numerous correlations between the target road and the nearby roads, which can be leveraged to improve prediction results.

In another recent related work, a novel vertical stack of two LSTM models was applied by \citet{wang2019traffic}, in conjunction with a network path segmentation algorithm, to effectively capture and elucidate both spatial and temporal correlations from network-wide high dimensional traffic data. Compared to the previous studies, this work naturally captures domain-related knowledge of transportation systems and presents realistic physical meaning; mainly, the output of the hidden layer is interpretable through visualization and qualitative analysis. Later, to leverage graph-based DNNs to improve network-wide spatial feature extraction, \citet{cui2020learning} introduced a graph wavelet gated recurrent (GWGR) neural network for capturing localized characteristics/features from the complicated geometric or topological structure of transportation networks. Specifically, the new feature extraction process enables increased flexibility by removing the need for an explicitly given human-defined topological network structure, such as the case in GCN.

Most recently, \citet{james2021citywide} proposed a real-time citywide traffic speed prediction model leveraging a new geometric deep learning approach. This design accounts for the complex structure of the urban road network directly in the learning process. Additionally, to further increase the model's ability to process massive transportation networks, this study suggested a tailor-made feature design and training technique combined with an attention mechanism. This design performs multi-step-ahead short-term traffic speed prediction while considering the correlations between prediction variables through improving basic regression trees. The result provides more information and better predicts the longer-term trend of traffic speed in large urban networks compared to a single-step approach.

While both aforementioned approaches improve our understanding of feature extraction and improve prediction accuracy compared to the existing baselines, there is still much room for improving the reliability and adaptability of the associated approaches. In particular, the methods extract network-specific spatial features from the underlying transportation data, which results in poor performance when attempting to generalize the models to other networks. Moreover, these models generally are trained on heavily preprocessed data where anomalies have been regularized or removed, limiting their prediction accuracy during non-recurrent network congestion or anomaly events. Further research is needed to improve upon these drawbacks, and methods such as transfer learning present promising avenues for future research.

\subsection{Travel Time Forecasting}
In addition to flow and speed forecasting, travel time estimation represents another related transportation application that can benefit from DL-based MTP approaches. The existing literature can be segmented into two broad categories: (1) path travel time approaches, which focus on a predicting the travel time for a predefined path or road segment; and (2) OD travel time estimation, where the model infers the driven path and outputs a prediction solely using the OD pair and timestamp as inference input. Below we describe both approaches in more detail.

\subsubsection{Path Travel Time}
In the path travel time prediction state, the prediction models take specific route(s) as input (derived from GPS trajectory data) and output the estimated travel time. Accurately estimating travel time for a given route is an important consideration for many transportation applications, including route planning (e.g., Google Maps), traffic monitoring, and taxi/delivery dispatching (e.g., Uber, Grubhub, Lyft). However, path travel time estimation is challenging due to the complex spatial correlations, external factors, and temporal dependencies in transportation networks.

In general, there are two primary methods for estimating path travel time: (1) individual, where the entire path is segmented into smaller chunks, and the travel time for each portion is estimated and summed together \citep{pan2012utilizing, yang2013travel, wang2014travel, wang2016traffic}; or (2) collective, where the entire path is considered and a single overall travel time estimate is generated \citep{jenelius2013travel}. The problem becomes more straightforward in the individual method by decomposing the path into smaller sub-problems. Still, these methods fail to capture the complex dependencies between the sub-paths resulting from intersections, traffic lights, and other network infrastructure considerations. On the other hand, the collective method can successfully capture and consider the intricate dependencies along the entire path but produce unreliable estimates for longer paths due to probe data sparsity (e.g., some paths are traveled considerably less than others). Consequently, the most recent research attempts to rectify these issues by proposing collective methods using DL models, which can better learn the underlying road network structure and improve the travel time prediction accuracy and reliability for longer trips \citep{wang2018will, li2020travel}. These works demonstrate that considering the network topology within the travel time prediction models can significantly improve accuracy and reliability, especially for longer trips. That being said, these studies fail to consider external factors (e.g., weather, special events, accidents, construction) in their analysis, which can further improve the prediction accuracy.

Most recently, two research works have identified this literature gap and integrated external factors into their modeling approach with great success \citep{chen2020long,li2021new}. Despite this, the cited works generally only consider a subset of external effects, such as just weather and planned special events, instead of considering all contributing factors. Moreover, unpredictable external events (e.g., accidents) and their impacts on prediction performance are not analyzed, presenting a future research direction. At the same time, the existing works primarily leverage trajectory data collected from taxis in their training and analysis of the models. However, taxi drivers are more experienced than traditional roadway users. They may not make mobility and routing decisions the same way, potentially hindering the generalization capability of the models for other drivers. Lastly, path travel time approaches can also be impeded by probe data density, as the methods need to infer the path taken by drivers between consecutive trajectory data points. When the time between data collection increases, the vehicle may have traversed multiple links and intersections between successive data points, making it challenging to infer the chosen path accurately. To resolve this challenge, some researchers have attempted to propose origin-destination (OD) travel time models, which can predict the travel time given only an origin and destination pair without inferring the path. We discuss the OD travel time state in more detail below.

\subsubsection{OD Travel Time}
In contrast to the path travel time prediction state, OD travel time seeks to leverage vehicle trajectory data to predict total trip travel time without requiring the exact path traversed by a given vehicle. The OD travel time prediction approach provides the benefits of reducing computation time while eliminating inference errors when using sparse trajectory data. Practically, the OD travel time state is of importance to the same applications as path travel time (e.g., route planning and traffic monitoring). Similar to path travel time, existing research has demonstrated that considering the spatial dependencies between links within the transportation network can improve prediction accuracy \citep{jenelius2013travel}. Other work has also improved prediction accuracy by leveraging the travel time of previous trips in their models to learn an encoding of OD and trajectory pairs for different departure times \citep{yuan2020effective}.

Moreover, considering external factors, including meteorological data, time-of-day, and historical travel time within the models also improves the predictive performance \citep{ li2018multi, xu2018trip, wang2018learning}. Notably, these studies have some drawbacks requiring future research. First, the generalizability of the models for predicting travel time for general traffic scenarios, in contrast to opportunistic probe vehicles (e.g., taxis), is not considered. Second, existing approaches assume the driver will always take the shortest distance path between origin and destination. This assumption is not always practical, as network conditions impact the travel time and chosen route (sometimes, the shortest distance path is not always the fastest).

Furthermore, these models generally consider limited modality data or leverage a single model without considering the one-sidedness of such a design. Consequently, a recent study proposed a combination of gradient boosting decision trees with a DNN to improve prediction accuracy while incorporating external factors such as meteorological data \citep{zou2020estimation}. While the case study proves this approach outperforms the existing baselines, the authors highlight room for further improvement. Specifically, additional external factors, utilizing more fine-grained weather data, and considering traffic flow data in conjunction with the trajectory data are not explored.

Additionally, following the emerging trend of learning the network topology as a graph, \citet{wang2021graphtte} proposed a combination of GCN and GRU to improve the spatial representation and better capture the dynamic spatio-temporal network conditions for OD travel time prediction. A case study is conducted, and the proposed method can reduce the MAPE error by upwards of 1.78\% compared to the state-of-the-art baselines. However, the model takes considerable time to train and perform inference, even on strong hardware, presenting room for improving the efficiency of this approach. This method also does not consider external data in its model, which has improved prediction accuracy for OD travel time models. It would be interesting to see if considering external factors and their impacts could further enhance the performance of this approach. For interested researchers, we provide the following simple baseline from the recent literature to assist in conducting a comparative analysis of future work \citep{wang2019simple}.

\subsection{Travel Demand Forecasting}
Accurate travel demand forecasting is essential to practical transportation applications such as dispatching and scheduling mobility and delivery services (e.g., Taxi, Uber, GrubHub). However, predicting travel demand is challenging given that the overall roadway demand is decentralized and results from many areas (e.g., demand for conventional taxi service, ride-hailing applications, service delivery, and non-recurrent events). In general, travel demand forecasting can be split into two subcategories: region-level demand prediction and OD demand prediction. In literature, there exists a large historical body of work on region-level demand prediction \citep{davis2016multi,liao2018large,yao2018deep,liu2020predicting}. The progression of the methods within the literature for the region-level travel demand forecasting state is similar to the other states: the earlier works focus on leveraging statistical models \citep{davis2016multi}, while the subsequent studies prove that combinations of DL-based models outperform the earlier statistical approaches \citep{liao2018large,yao2018deep}.

Later, to improve prediction performance in large-scale urban transportation networks, \citet{liu2020predicting} demonstrated the effectiveness of integrating a novel attention-based mechanism with DL-based models, outperforming the existing baselines. In an alternative approach to better capture the exogenous dependencies and underlying correlations of travel demand, \citet{guo2020residual} introduced a novel residual spatio-temporal network (RSTN) model. This approach combines CNN and LSTM while reformulating the prediction problem to learn a residual function based on the travel density in each temporal epoch. The resulting model is easier to train, and the case study demonstrated its improved accuracy compared to other state-of-the-art baseline approaches. Moreover, at the multi-region level, \citet{tang2021multi} recently proposed a multi-community spatio-temporal graph convolutional network (MC STGCN) to overcome the challenge of capturing inter-region demand correlations of complex urban road network structures with irregular shapes and arrangements.

Despite much research attention to region-level demand prediction, few attempts have been made to explore the origin-destination (OD) demand prediction problem \citep{wang2017deepsd,wang2019origin,liu2019contextualized}. One problem with the existing studies is that they either do not suitably account for the non-Euclidian demand structure in the data or only capture the pair-wise relationships using undirected and static graphs. Additionally, it is significantly more difficult to perform OD-based demand forecasting compared to the region-level due to the challenge of capturing the spatio-temporal correlation between two different OD pairs. Attempting to overcome this challenge, \citet{ke2021predicting} proposed OD graphs for characterizing non-Euclidian pair-wise geographical and semantic connections between various OD pairs and demonstrated that their design outperforms the baseline algorithms by a significant margin. In a different approach, \citet{zhang2021dneat} captured the Euclidean and non-Euclidean spatial-temporal relationships of OD graphs using a novel Dynamic Node-Edge Attention Network (DNEAT) model, which incorporates a temporal node-edge attention mechanism. Experimental results demonstrated that the model outperformed six baselines across two different datasets while showing increased resilience to sparse training data.

\newgeometry{left=1.2cm,right=1.2cm,top=0cm,bottom=0.2cm} 

\begin{landscape}
\begin{center}
\tiny
\begin{longtable}[!h]{lclccllllc}

\caption{A summary of most recent traffic prediction studies .} \label{trafficlit} \\

\cline{1-9}

\multirow{3}{*}{\bfseries Author}&\multirow{3}{*}{\bfseries Area}&\multirow{2}{*}{\bfseries Dataset} &\bfseries Data &\multirow{2}{*}{\bfseries External} &\multirow{2}{*}{\ \bfseries Application}&\multirow{3}{*}{\bfseries Model}&\multirow{2}{*}{Evaluated} &\multirow{2}{*}{ \bfseries Comparative}&
\\
&&\multirow{2}{*}{\bfseries(Type)}&\bfseries Interval&\multirow{2}{*}{\bfseries Data}&\multirow{2}{*}{ \bfseries Task}& &\multirow{2}{*}{Algorithm}&\multirow{2}{*}{ \bfseries Model}&
\\
&&&\bfseries (min) &\bfseries& &&& &
\\\cline{1-9}

\endfirsthead

\multicolumn{10}{c}{{\bfseries \tablename\ \thetable{} -- (Continued)}} \\

\cline{1-9}

\multirow{3}{*}{\bfseries Author}&\multirow{3}{*}{\bfseries Area}&\multirow{2}{*}{\bfseries Dataset} &\bfseries Data &\multirow{2}{*}{\bfseries External} &\multirow{2}{*}{\ \bfseries Application}&\multirow{3}{*}{\bfseries Model}&\multirow{2}{*}{Evaluated} &\multirow{2}{*}{ \bfseries Comparative}&
\\
&&\multirow{2}{*}{\bfseries(Type)}&\bfseries Interval&\multirow{2}{*}{\bfseries Data}&\multirow{2}{*}{ \bfseries Task}& &\multirow{2}{*}{Algorithm}&\multirow{2}{*}{ \bfseries Model}&
\\
&&&\bfseries (min) &\bfseries& &&& &
\\\cline{1-9}

\endhead

\cline{1-9}\multicolumn{10}{l}{{S: Spatial Dependency, T: Temporal Dependency, F: Freeway(s), E: Expressway(s), H: Highway(s), U: Urban Arterial/Road Networks}} \\

\endfoot

\cline{1-9}\multicolumn{10}{l}{{ S: Spatial Dependency, T: Temporal Dependency, F: Freeway(s), E: Expressway(s), H: Highway(s), U: Urban Arterial/Road Networks}} \\

\endlastfoot

\multirow{3}{*}{\citet{tran2022short}$^{S,T}$}&\multirow{3}{*}{U}&\multirow{3}{*}{Haiphong
(GPS)}&\multirow{3}{*}{10}&\multirow{3}{*}{$\times$}&\multirow{3}{*}{Speed}&\multirow{3}{*}{LSTM}&\multirow{3}{*}{RMSE, MAE, MDAE}&CNN, AR, ARMA, ARIMA, &\\
&&&&&&&&SARIMAX, PROPHET, MLP,&\\
&&&&&&&&SES,
HWES&\\\cline{1-9}

\multirow{3}{*}{\citet{zafar2022applying}$^{S,T}$}&\multirow{3}{*}{U}&FCD(GPS)&\multirow{3}{*}{15}&\multirow{3}{*}{$\checkmark$}&\multirow{3}{*}{Speed}&\multirow{3}{*}{LSTM-GRU}&\multirow{3}{*}{ RMSE, MAE, MAPE}&\multirow{2}{*}{KNN, XGBoost,ANN,}&\\
&&OSM&&&&&&\multirow{2}{*}{Linear Regressor,MLP} &\\
&&ETA\\\cline{1-9}

\multirow{2}{*}{\citet{jian2022spatiotemporal}$^{S,T}$}&\multirow{3}{*}{F}&
PeMSD4,PeMSD8&\multirow{2}{*}{15,30,60}&\multirow{3}{*}{$\times$}&\multirow{2}{*}{Speed}&\multirow{2}{*}{ST-DWGRU} &\multirow{2}{*}{RMSE, MAE,
MAP}&HA ,ARIMA,STGCN, DCRNN,
&\\
&&
PeMS-BAY(Detector)&&&&&&GWN,ASTGCN,LSGCN\\\cline{1-9}

\multirow{2}{*}{\citet{jin2022gan}$^{S,T}$}&\multirow{2}{*}{U}&\multirow{2}{*}{AutoNavi}&\multirow{2}{*}{15}&\multirow{2}{*}{$\times$}&\multirow{2}{*}{Speed}&\multirow{2}{*}{PL-WGAN}&MAE,
RMSE,ACC,&HA, ARIMA, SVR, GRU, GCN,
&\\
&&&&&&&$R^2$,VAR&
ST-GCN,T-GCN,
A3T-GCN&\\\cline{1-9}

\multirow{2}{*}{\citet{lee2022ddp}$^{S,T}$}&\multirow{2}{*}{F}&\multirow{2}{*}{SeoulTaxi(GPS)}&\multirow{2}{*}{5}&\multirow{2}{*}{$\times$}&\multirow{2}{*}{Speed}&\multirow{2}{*}{DDP-GCN}&\multirow{2}{*}{MAPE,MAE,RMSE}& HA,VAR,LSVR,ARIMA,STGCN&\\
&&&&&&&&FC-LSTM,DCRNN&\\\cline{1-9}

\multirow{2}{*}{\citet{lu2022traffic}$^{S,T}$}&E&Guangzhou(Detectors)&10&\multirow{2}{*}{$\times$}&\multirow{2}{*}{Speed}&\multirow{2}{*}{DENN}&RMSE, MAE,$R^2$,&HA,ARIMA,SVR,LSTM,FDL,&
\\
&F&PEMS-D8(Detectors)&5&&&&ACC&STNN,BTF
,GLA,T-GCN&\\\cline{1-9}

\multirow{3}{*}{\citet{zhao2022attention}$^{S,T}$}&\multirow{2}{*}{F}&\multirow{2}{*}{METR-LA,PEMS-BAY,} &\multirow{3}{*}{5}&\multirow{3}{*}{$\times$}&\multirow{3}{*}{Speed}&\multirow{3}{*}{ADSTGCN}&\multirow{3}{*}{RMSE,MAE,MAPE}&HA,ARIMA,FNN,FC-LSTM,&\\
&\multirow{2}{*}{H}&\multirow{2}{*}{PEMA-S (Detectors)}&&&&&&DCRNN,STGCN,SLCNN,&\\
&&&&&&&&GMAN,STAWnet&\\\cline{1-9}

\multirow{2}{*}{\citet{ma2022novel}$^{S,T}$}&\multirow{2}{*}{U}&\multirow{2}{*}{Hangzhou(Detectors)}&\multirow{2}{*}{5}&\multirow{2}{*}{$\times$}&\multirow{2}{*}{Network Flow}&\multirow{2}{*}{STFSA-CNN-GRU}&\multirow{2}{*}{MAPE,MAE,RMSE }&ARIMA, SVR, CNN, RNN,
&\\
&&&&&&&&LSTM,GRU\\\cline{1-9}

\multirow{2}{*}{\citet{wang2022attention}$^{S,T}$}&\multirow{2}{*}{F}&PeMSD4,PeMSD8&5&\multirow{2}{*}{$\times$}&\multirow{2}{*}{Network Flow}&\multirow{2}{*}{ASTGAT}&\multirow{2}{*}{MAE,RMSE}&HA,ARIMA,VAR,LSTM,GRU,&\\
&&(Detectors)&&&&&&STGCN,GeoMan,ASTGCN(12)&\\\cline{1-9} 

\multirow{2}{*}{\citet{aljuaydi2022deep}$^{S,T}$}&\multirow{2}{*}{F}&MRWA&1&\multirow{2}{*}{\checkmark}&\multirow{2}{*}{Regional Flow}&\multirow{2}{*}{1-D CNN LSTM}&\multirow{2}{*}{MAE, RMSE}&\multirow{2}{*}{MLP}&\\
&&WebEOC&15\\

\cline{1-9}

\multirow{2}{*}{\citet{sun2021modeling}$^{S,T}$}&H&METR-LA(Detectors)&\multirow{2}{*}{NA}
&\multirow{2}{*}{$\times$}&\multirow{2}{*}{Network Flow}&\multirow{2}{*}{GST+GAT}&\multirow{2}{*}{MAE,RMsE,MAPE}&ARIMA,SVR,LSTM,STGCN,&\\
&U&PeMS-BAY(Detectors)&&&&&&ASTGCN,STANN, GMAN\\\cline{1-9}

\multirow{2}{*}{\citet{li2021multistep}$^{S,T}$}&\multirow{2}{*}{F}&NDW-Netherlands&\multirow{2}{*}{1$\sim$5}&\multirow{2}{*}{$\times$}&\multirow{2}{*}{Network Flow}&\multirow{2}{*}{DGCN}&\multirow{2}{*}{MAE,MAPE,RMSE}&HA,KNN,GAT,DCNN,&\\
&&(Detectors)&&&&&&DCRNN,STGCN,Graph Wavenet&\\\cline{1-9}

\multirow{3}{*}{\citet{wang2021graphtte}$^{S,T}$}&\multirow{3}{*}{U}&\multirow{2}{*}{Chengdu Taxi (GPS)}&\multirow{3}{*}{10}&\multirow{3}{*}{\checkmark}&\multirow{3}{*}{OD Travel Time}&\multirow{3}{*}{GraphTTE}&\multirow{3}{*}{MAPE,RMSE,MAE}&Average, GBDT, DeepTTE, &\\
&&\multirow{2}{*}{Xi'an Taxi (GPS)}&&&&&&GTTE-D,GraphTTE,GTTE-S,&\\
&&&&&&&& DeepOD&\\\cline{1-9}

\multirow{2}{*}{\cite{li2021new}$^T$}&\multirow{2}{*}{E}&\multirow{2}{*}{OPENITS(Detectors)}&5 &\multirow{2}{*}{\checkmark}&\multirow{2}{*}{Path Travel Time}&soft set theory&MAE, MAPE,  R$^2$,&\multirow{2}{*}{BPR}&\\
&&&1&& & model&RMSPE&\\\cline{1-9}

\multirow{2}{*}{\citet{james2021citywide}$^{S,T}$}&\multirow{2}{*}{U}
&NavInfo(Trajectory)&\multirow{2}{*}{5}
&\multirow{2}{*}{$\times$}&\multirow{2}{*}{Speed}&\multirow{2}{*}{$GA^2$}
&MAPE&HA, AK, SVR, B-LSTM, 
&\\
&&OSM&&&&&&
DCRNN
, T-GCN
, A3TGCN
\\
\cline{1-9}

\multirow{3}{*}{\citet{guo2021hierarchical}$^{S,T}$}&\multirow{3}{*}{U}&\multirow{2}{*}{Didi Chuxing GAIA} &\multirow{3}{*}{10}&\multirow{3}{*}{$\times$}&\multirow{3}{*}{Speed}&HGCN&\multirow{3}{*}{MAE, MAPE, RMSE}&HA,OTSGGCN,  LSTM,
GRU,  &\\
&&\multirow{2}{*}{Initiative (GPS)}&&&&HGCN-WH&&GatedSTGCN,GCRN,ARIMA,  &\\
&&&&&&HGCN-WDF&&OGCRNN, GWNET&\\\cline{1-9}

 \multirow{2}{*}{\citet{li2021hybrid}$^T$}&\multirow{2}{*}{H} &\multirow{2}{*}{England}&\multirow{2}{*}{15}&\multirow{2}{*}{$\checkmark$}&\multirow{2}{*}{Network Flow}&\multirow{2}{*}{W-CNN-LSTM}&\multirow{2}{*}{RMSE,MAE,R$^2$}&LSTM,MLP,ARIMA,&\\
&&&&&&&& CNN-LSTM\\\cline{1-9}

 \multirow{3}{*}{\citet{ke2021predicting}$^{S,T}$}& \multirow{3}{*}{U}& \multirow{2}{*}{NYC Taxi \& Limousine }& \multirow{3}{*}{30} & \multirow{3}{*}{$\times$}& \multirow{3}{*}{Travel Demand}& \multirow{3}{*}{ST-ED-RMGC}& \multirow{3}{*}{RMSE,MAE,MAPE}&HA,GBDT,RF,XGB,LASSO,&\\
&& \multirow{2}{*}{Commission(GPS)}&&& &&&MLP,LSTM,Spatial LSTM,\\
 &&&& &&&&MGC,ED-MGC,RMGC\\\cline{1-9}
 
 \multirow{4}{*}{\citet{zhang2021dneat}$^{S,T}$}&U&Didi-Chuxing(GPS)&10& \multirow{4}{*}{$\times$}&\multirow{4}{*}{Travel Demand}&\multirow{4}{*}{DNEAT}&\multirow{4}{*}{RMSE,MAPE}&\multirow{3}{*}{HA,LSTM, TCN,DySAT}&\\
& \multirow{2}{*}{U}&NYC Taxi \&Limousine & \multirow{2}{*}{15}&&&&&\multirow{3}{*}{ConvLSTM,ST-GCN}\\
&&Commission (GPS)&&&&\\
&U&OSM&&&&\\\cline{1-9}

\multirow{2}{*}{\citet{ma2020hybrid}$^{S,T}$}&\multirow{2}{*}{H}&\multirow{2}{*}{Tel Aviv ( Detectors)}&\multirow{2}{*}{60}&\multirow{2}{*}{$\times$}&\multirow{2}{*}{Network Flow}&\multirow{2}{*}{NN-ARIMA}
&\multirow{2}{*}{MAPE, MSE}&NN,ARIMA,MSVR,&\\
&&&&&&&&MSVR-ARIMA\\\cline{1-9}

\multirow{2}{*}{\citet{zou2020estimation}$^T$}&U&Shanghai Taxi(GPS)&\multirow{2}{*}{20s$\sim$30s}&\multirow{2}{*}{\checkmark}&\multirow{2}{*}{OD Travel Time}&\multirow{2}{*}{TTE-Ensemble}
&\multirow{2}{*}{MAE,MAPE,MARE}&\multirow{2}{*}{LRM, GBDT, DNN}&\\
&U&Chengdu Taxi (GPS)&&&\\
\cline{1-9}

\multirow{4}{*}{\citet{yuan2020effective}$^{S,T}$}&\multirow{4}{*}{U}&Chengdu Taxi (GPS)&3s&\multirow{4}{*}{\checkmark}&\multirow{4}{*}{OD Travel Time} &\multirow{4}{*}{DeepOD}&\multirow{4}{*}{MAE, MAPE,MARE} &\multirow{3}{*}{TEMP,LR,GMB,}&\\
&&Xi'an Taxi (GPS)&3s&&&&&\multirow{3}{*}{STNN,MURAT}\\
&&Beijing Taxi (GPS)&1&&\\
&&OSM&&&\\\cline{1-9}

\multirow{2}{*}{\citet{xu2020mtlm}$^{S,T}$}&\multirow{2}{*}{U}&\multirow{2}{*}{GeoLife(Trajectory)}&\multirow{2}{*}{1s$\sim$5s}& \multirow{2}{*}{$\times$}&\multirow{2}{*}{OD Travel Time}&\multirow{2}{*}{MTLM}&\multirow{2}{*}{MAE
, MAPE,RMSE}&GBDT,MlpTTE,Deeptravel, &\\
&&&&&&&&DeepTTE\\\cline{1-9}

\cite{li2020travel}$^S$&U&Sydney Taxi (GPS)&NA&$\times$ &Path Travel Time&SFM&&FM, FFM, CFM&\\\cline{1-9}

\citet{chen2020long}$^T$&F&Taiwan(ETC)&5&\checkmark&Path Travel Time&GB&MAPE,MAE,RMSE&Random Forest&\\\cline{1-9}

\multirow{3}{*}{\cite{guo2020residual}$^{S,T}$}&U&DiDi ChuXing (GPS)&30&\multirow{3}{*}{$\times$}&\multirow{3}{*}{Travel Demand} &\multirow{2}{*}{FCNs+CNN}&\multirow{3}{*}{SMAPE,MAE,RMSE}&\multirow{2}{*}{STN
, LSTM-C
, ARIMA,}&
\\
&\multirow{2}{*}{U}&NYS Taxi \& Limousine &\multirow{2}{*}{30}&& &\multirow{2}{*}{+LSTM}&&\multirow{3}{*}{ST-ResNet, FCL-Net
FCL-Net-v}
\\
&&Commission (GPS)&&&\\\cline{1-9}

\multirow{3}{*}{\citet{kuang2020traffic}$^{S,T}$}&\multirow{2}{*}{U}&\multirow{2}{*}{DiDi-Chengdu(GPS)}&\multirow{3}{*}{NA}&\multirow{3}{*}{$\times$}&\multirow{3}{*}{Regional Flow}&\multirow{3}{*}{ResNet+TCN}&\multirow{3}{*}{RMSE}&HA, ARIMA, XGBoost,SAE,&\\
&\multirow{2}{*}{U}&\multirow{2}{*}{Chengdu Bus (GPS)}&&&&&&GRU,CNN-LSTM,ST-ResNet,\\
&&&&&&&&,LSTM\\\cline{1-9}

\multirow{3}{*}{\citet{guo2020optimized}$^{S,T}$}& U&D.C(Detectors)&5&\multirow{3}{*}{$\times$}&\multirow{3}{*}{Network Flow}&\multirow{3}{*}{GCN+GRU}&\multirow{3}{*}{MAE, MAPE, MSE}&\multirow{2}{*}{HA, SVR, Random
Forest,}&\\
&U&Philadelphia(Detectors)&5&&&&& \multirow{2}{*}{ARIMA, FNN, GRU, GCGRU} \\
&U&PeMSD4(Detectors)&5&&&&  \\\cline{1-9}

\multirow{3}{*}{\citet{song2020spatial}$^{S,T}$}&\multirow{3}{*}{U}&\multirow{2}{*}{PEMS3, PEMS4, PEMS7,} &\multirow{3}{*}{5}&\multirow{3}{*}{$\times$}&\multirow{3}{*}{Network Flow}&\multirow{3}{*}{STSGCM}&\multirow{3}{*}{MAE,MAPE,RMSE}&VAR,SVR,LSTM,&\\
&&\multirow{2}{*}{PEMS08(Detectors)}&&&&&&DCRNN,STGCN, ASTGCN(r),\\
&&&&&&&&STG2Seq, Graph WaveNet \\\cline{1-9}

\multirow{2}{*}{\citet{dai2020spatio}$^{S,T}$}&\multirow{2}{*}{U
}& Ninjing (video  &15, 30, 45, 
&\multirow{2}{*}{$\times$}&\multirow{2}{*}{Speed}&ConvLSTM
&\multirow{2}{*}{RMSE}&\multirow{2}{*}{GCN, ConvLSTM, TGCN}
&\\
&&surveillance devices)&60, 75, 90 &&&+GCN&&\\\cline{1-9}

\multirow{3}{*}{\citet{cui2020learning}$^{S,T}$}&\multirow{2}{*}{F}&\multirow{2}{*}{Loop ( Detectors)}&\multirow{2}{*}{5}&\multirow{3}{*}{$\times$}&\multirow{3}{*}{Speed}&\multirow{3}{*}{GWGR}
&\multirow{3}{*}{MAE, MAPE, RMSE}& ARIMA, SVR,FNN,LSTM,
&\\
&\multirow{2}{*}{U}&\multirow{2}{*}{NPMRDS (Probe veh.)}&\multirow{2}{*}{5}&&&&&LSGC + LSTM
,SGC + LSTM

&\\
&&&&&&&&STGCN
, TGCLSTM, 
&\\\cline{1-9}

\multirow{2}{*}{\citet{wang2019traffic}$^{S,T}$}&\multirow{2}{*}{U}&\multirow{2}{*}{Xuancheng(AVI)}  &\multirow{2}{*}{5}&\multirow{2}{*}{$\times$}&\multirow{2}{*}{Speed}&\multirow{2}{*}{Bi-LSTM NN}&\multirow{2}{*}{MAE}&PBDL-CP,PBDL-RP,PBDL-CP&\\
&&&&&&&&CNN,LSTM NN ,ANN,KNN&\\\cline{1-9}

\multirow{2}{*}{\citet{bogaerts2020graph}$^{S,T}$}&\multirow{2}{*}{U}&DiDiXi'an (GPS)&\multirow{2}{*}{5}&\multirow{2}{*}{$\times$}&\multirow{2}{*}{Speed}&GraphCNN+LSTM+&RMSE,MSE,MAE,&\multirow{2}{*}{k-NN,
LSTM, SVM}&\\
&&DiDiChengdu (GPS)&&&&FNNN&MAPE&\\
\cline{1-9}
 
  \multirow{2}{*}{\citet{liu2019contextualized}$^{S,T}$}& \multirow{2}{*}{U}& NYC Taxi \& Limousine & \multirow{2}{*}{30}
 & \multirow{2}{*}{\checkmark}& \multirow{2}{*}{Travel Demand}& \multirow{2}{*}{CSTN}& \multirow{2}{*}{OD-MAPE,O-MAPE}& HA,LR,XGBoost,MLP, &\\
&& Commission (GPS)&& &&&&ConvLSTM,ST-ResNet \\\cline{1-9}

 \multirow{2}{*}{\citet{wang2019origin}$^{S,T}$}&U&UCAR-Beijing(GPS)&\multirow{2}{*}{60} &\multirow{2}{*}{\checkmark}&\multirow{2}{*}{Travel Demand}&\multirow{2}{*}{GEML}& \multirow{2}{*}{RMSE,SMAPE}&\multirow{2}{*}{ HA,LSTM,LSTNet,GCRN}&\\
 &U&DiDi-Chengdu(GPS)&&&&\\\cline{1-9}

\multirow{2}{*}{\citet{liu2019deeprtp}$^{S,T}$}&\multirow{2}{*}{U}&\multirow{2}{*}{OSM}&\multirow{2}{*}{15}&\multirow{2}{*}{$\times$}&\multirow{2}{*}{Regional Flow}&\multirow{2}{*}{DeepRTP}&\multirow{2}{*}{RMSE}&\multirow{1}{*}{HisAve, HisAve-w, ARIMA,}&\\
&&&&&&&&\multirow{1}{*}{SARIMA,VAR}\\
\cline{1-9}

\multirow{2}{*}{\citet{zhao2019t}$^{S,T}$}&U&SZ-taxi(GPS)&15&\multirow{2}{*}{$\times$}&\multirow{2}{*}{Network Flow}&\multirow{2}{*}{GCN+GRU}&MAE,RMSE,$R^2$,& \multirow{2}{*}{HA,ARIMA,SVR,GCN,GRU}&\\
&H&LOS-loop(Detectors)&5&&&&var,Accuracy&\\\cline{1-9}

\multirow{3}{*}{\citet{zhang2019trafficgan}$^{S,T}$}&\multirow{2}{*}{U}
&\multirow{2}{*}{CTA buses (GPS)}&\multirow{2}{*}{10}&\multirow{3}{*}{$\times$}&\multirow{3}{*}{Network Flow}& \multirow{2}{*}{GAN+CNN}&\multirow{3}{*}{MAE, MRE, RMSE}&ARIMA, SVR,
CNN-SVR&
\\
&\multirow{2}{*}{U}&\multirow{2}{*}{PeMS(Detectors)}&\multirow{2}{*}{NA}&&&\multirow{2}{*}{+LSTM }&&, the scalable deep traffic 
\\
&&&&&&&&flow neural networks,DCRNN 
\\\cline{1-9}

\multirow{2}{*}{\citet{qu2019daily}$^T$}&\multirow{2}{*}{F}&\multirow{2}{*}{DRIVENET( Detectors)}&2,10,15,&\multirow{2}{*}{$\checkmark$}&\multirow{2}{*}{Network Flow}&\multirow{2}{*}{DNN}&\multirow{2}{*}{MAPE}&\multirow{2}{*}{CM}&\\
&&&30,60&&&&&\\\cline{1-9}

\multirow{2}{*}{\citet{liu2019traffic}$^{S,T}$} &\multirow{2}{*}{R}&PeMS&\multirow{2}{*}{5}&\multirow{2}{*}{$\times$}&\multirow{2}{*}{Speed}&TCHA&MAE,MAPE,RMSE&SVR, SAE, LSTM,&\\
&&(Detectors)&&&&&& GRU, hierarchical attention\\
\cline{1-9}

\multirow{2}{*}{\citet{wu2019graph}$^{S,T}$}&H&METR-LA(Detectors)&\multirow{2}{*}{NA}&\multirow{2}{*}{$\times$}&\multirow{2}{*}{Speed}&\multirow{2}{*}{Graph WaveNet}&\multirow{2}{*}{MAE, RMSE, MAPE}&ARIMA, FC-LSTM , WaveNet, &\\
&U&PEMS-BAY(Detectors)&&&&&&DCRNN, GGRU, GGRU, STGCN&\\\cline{1-9}

\cline{1-9}

  \end{longtable}
\end{center}

\end{landscape}
\restoregeometry

\section{Open Research Issues and Future Research Directions}
\label{Sec6}

\begin{figure}[t]
    \centering
    \includegraphics[width=.8\textwidth]{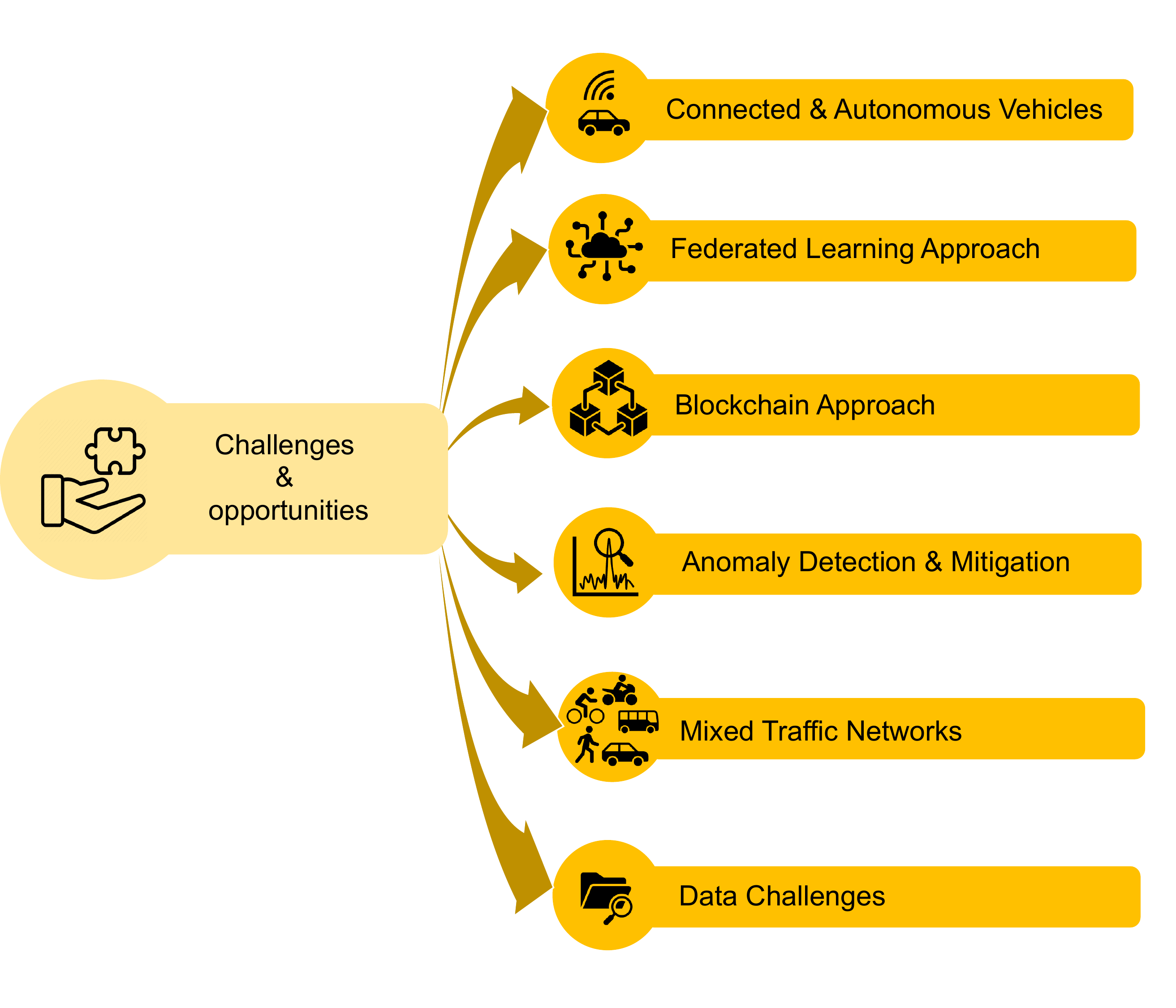}
    \caption{Schematic overview of identified challenges and suggested research
opportunities}
      \label{fig:chal}
\end{figure}

Section 5 outlines the various traffic prediction states in the literature and their associated applications. However, applying the developed models and assessing their real-world benefits presents open research challenges. Consequently, the existing literature generally concludes its performance analysis after quantifying the model's predictive accuracy. In other words, it fails to assess the real-world benefits of their approach for facilitating practical applications (e.g., vehicle routing). As shown in Fig. \ref{fig:chal}, we highlight some of the open research areas and future challenges in DL-based traffic prediction and elucidate how the common traffic prediction states and practical problems influence future research directions in this section.

\subsection{Traffic Prediction For Connected and Autonomous Vehicles}

Reliable traffic prediction methods will be a key enabler of autonomous vehicle applications within the IoV to ensure efficient vehicle routing within the transportation network. AV driving systems can revolutionize societal mobility and provide widespread benefits, including better utilization of existing roadway infrastructure, increased access to various mobility services, reduced negative environmental impact from vehicle usage, and improved safety. AVs are expected to eventually eliminate driver ignorance in favor of dynamic and data-driven intelligent decision systems that adapt their approach to the surrounding environmental conditions. Consequently, new applications with the potential to reduce traffic congestion will become possible, including adaptive cruise control and connected and autonomous traffic control systems \citep{zhong2020influence}. In this work, a simulation analysis elucidates that introducing CAVs into the mixed traffic environment will considerably impact the behavior of human drivers (e.g., increasing lane change frequency). Notably, the performance of these AV-based systems will become inherently dependent on the available traffic prediction capabilities \citep{miglani2019deep, zhong2020autonomous}.

However, the decentralized nature of the IoV environment, where data is collected and stored on a diverse set of heterogeneous devices (e.g., CAV, base station, RSU, sensor, edge device), does not match well with the existing cutting-edge traffic prediction methods due to their centralized nature. For example, network-scale traffic prediction has been studied extensively as a way to predict the traffic for all links within a complex urban transportation network simultaneously to support vehicle routing and traffic management decisions \citep{cui2020learning}. Still, the resulting models require immense amounts of traffic data to be collected throughout the network and aggregated to a central server (e.g., cloud) for processing and model training. This design presents privacy concerns in the decentralized IoV environment, where most trajectory data will be crowd-sourced and collected by Connected Autonomous vehicles (CAVs) owned by multiple stakeholders, in contrast to a single party (e.g., TMC or DoT) collecting processing all of the traffic data. 

Moreover, latency and bandwidth for excessive data transfer present additional challenges, and it is unlikely that it will be efficient to transfer and store all connected vehicle and sensor data in the centralized cloud. Additionally, energy consumption for onboard model training within CAVs presents a critical concern. Energy-efficient computation offloading strategies for ITS networks is an open research area receiving considerable attention \citep{li2020energy}. Notably, the existing research reveals that an optimization strategy considering the computational resources of RSUs, edge devices, and mobile base stations is a promising direction. However, this system will need to interface and account for the capacity of existing mobile edge computing networks and electrical grid infrastructure, presenting a complex challenge for future research.

Consequently, additional research into privacy-preserving and decentralized traffic prediction models will be necessary to support the deployment of CAVs and associated applications (e.g., ridesharing and vehicle loaning) within the emerging IoV-based transportation networks. Recently, FL has been proposed as one method for achieving decentralized and privacy-preserving traffic prediction models \citep{liu2020privacy}; however, the approach is still new and requires significant improvements before the proposed models are practical. Below, we provide more details on FL and its potential to revolutionize traffic prediction within the IoV. 

\subsection{Federated Learning for Traffic Prediction}

FL is an emerging approach to privacy-preserving and distributed machine learning that can potentially revolutionize the future of traffic prediction within the IoV. In FL, participating devices work collaboratively to train a shared prediction model iteratively in a series of rounds. During each round, the traffic collection device will share just the learned parameters (gradients) with the other devices. Specifically, each device uses its own collected data to train a localized model. After training, devices will periodically transmit the local model state to a parameter server, where an algorithm (e.g., Federated Averaging) is used to generate a new global set of parameters from the contributions of each participant. After that, the global parameter set is distributed back to each device, and the process repeats until a convergence condition is achieved. A comprehensive survey on FL is provided in \cite{du2020federated} for interested readers.

Regarding traffic prediction in the IoV, FL can train a more comprehensive traffic prediction model and continuously update it while preserving the privacy of crowd-sourced CAV data and alleviating the need to transmit and store vast quantities of distributed data. Recently, there have been a few attempts to propose FL-based models for traffic prediction; notably, federated versions of the GRU-based traffic prediction model are presented in \cite{liu2020privacy} and \cite{qi2021privacy}. However, the simulations conducted in these studies are not entirely practical as they appear to simulate FL on identical data shards distributed amongst a set of participants. In the practical IoV environment, devices are heterogeneous, and participants are autonomous, resulting in a data skew between participants. Notably, problems of data imbalance, missing classes, missing features, and missing values can limit the practical implementation of FL-based traffic prediction models. When these problems arise, the resulting models perform poorly compared to the existing approaches. As data skew is likely to occur in the distributed IoV environment, further research will be necessary to account for data skew amongst participants and improve the performance of the proposed traffic prediction FL models under this constraint.

In addition to data skew, it has yet to be discovered whether the best-performing traffic prediction model designs can be trained using Federated Learning. As we learned from previous sections, the revolutionary models within the literature are hybrid approaches operating on high-dimensional multivariate time series sensor data. They leverage DL methods (e.g., convolution) to extract the input data's spatial, temporal, and spatio-temporal trends. However, doing so requires an aggregated and pre-processed data structure built upon data sharing. This observation raises the question of whether it is possible to train an online FL-based traffic prediction model that can provide suitable accuracy compared to the centralized approaches. While probe trajectory data collected from mobile devices or CAVs will include some private and identifiable information, the privacy of ground truth traffic sensor data is not a practical consideration. Consequently, one potential option is to relax the data sharing assumption for localized groups or neighborhoods of sensors. In this way, a more complex model benefiting from feature extraction of multivariate sensor data can be realized while simultaneously benefiting from the distributed edge computation and reduced communication overhead provided by FL.

In summary, due to the distributed and heterogeneous nature of data collection in IoV, FL-based models are a promising method for leveraging the benefits of edge computing within the IoV during model training for various MTP applications. Other related disciplines with similar data and network structures, including industrial IoT \citep{lu2019blockchain}, have demonstrated the effectiveness of FL approaches in distributed networks having properties of strong temporal and spatial locality. Since traffic time series data and the IoV closely resemble the industrial IoT environment, we believe FL designs have strong applicability to MTP applications.  

\subsection{Blockchain-enabled Traffic Prediction}
Existing traffic prediction methods based on centralized machine learning need to gather raw data for model training, which involves serious privacy exposure risks when the data is collected by heterogeneous IoV devices owned by multiple stakeholders. Due to the inherent immutability and decentralization qualities of blockchain networks, this technology has enabled considerable advances in traffic prediction capabilities during the last few years. For accurate and efficient traffic jam probability estimation, \citet{9107472} proposed a neural network-based smart contract to be deployed onto the blockchain network. \citet{qi2021privacy} proposed a consortium blockchain-based FL framework is proposed to enable decentralized, reliable, and secure FL without a centralized model coordinator. In addition, \citet{shahbazi2021framework} introduced a blockchain-based framework for traffic demand service prediction. Despite much progress recently, blockchain technology has yet to be fully formed or standardized in the traffic prediction context. For example, existing popular blockchain networks (e.g., Bitcoin and Ethereum) require high energy consumption, provide low transaction throughput, and have storage and bandwidth scalability problems. 

Like FL, blockchain has strong applicability to the IoV and MTP applications because of its inherent decentralized network structure with fault tolerance, strong immutability, and pluggable security protocols \citep{wang2021survey}. Advancements in edge computing have demonstrated the benefits of data locality (e.g., having data and computation exist close to the user or point of access) \citep{carlini2016drivers}. For MTP within the IoV, blockchain can provide an edge solution for distributing computation (e.g., model training and inference), storing, and sharing data. In contrast to distributed databases, blockchain networks support smart contracts, which are chunks of code capable of controlling access to computation and storage resources. This way, a programmable interface enables fine-grained security controls and supports distributed applications. However, the additional functionality comes at the cost of added resource consumption for nodes managing the blockchain network, compared to a distributed database approach. Consequently, if distributed and edge-based data storage and retrieval are the only desired operations, MTP applications may benefit from the more lightweight approach combining local peer-to-peer gossip networks and distributed databases.

That being said, smart contracts present an exciting method for enabling distributing computation within the IoV and have the potential to support much future distributed IoV applications (e.g., ridesharing) \citep{li2020blockchain}. However, concerns of excessive energy consumption, low throughput, and high resource consumption associated with the existing blockchain networks necessitate further research to improve the scalability and efficiency of these distributed networks. Thankfully, much research is being performed to address these challenges using novel approaches such as zero-knowledge proof transaction offloading \citep{ashur2018marvellous, panait2020using, ben2019scalable}, and lighter-weight consensus schemes \citep{guo2021location} designed for location-dependent networks (e.g., IoV).

\subsection{Anomaly Traffic Event Detection and Mitigation}

Non-recurrent anomaly events, such as accidents, work zones, weather, and special events, drastically reduce the capacity of the transportation network, creating traffic congestion. According to the Federal Highway Administration (FHWA), it was estimated that about 50\% of traffic network congestion results from non-recurrent events \citep{fhwaCongestion}. Resolving non-recurrent congestion and returning the network to total capacity requires aggressive action and management by the TMC and associated stakeholders (e.g., police and emergency services). Example response efforts include timely communication to travelers, dynamic rerouting of existing traffic, and removing any obstructions from the traveled way. However, coordinating a response effort and communicating updates to travelers takes time. Consequently, developing methods for rapid anomaly detection is an exciting and essential research area for the transportation community.

While anomaly detection has been studied more extensively in wireless networks, the literature for anomaly detection in transportation networks using traffic data is small. In one related work, \citet{lu2009path}, novel path anomaly detection algorithms leveraging KNNs are proposed for urban-scale transportation networks. The authors conducted a case study and concluded the classification accuracy to be around 90\%. However, the case study was conducted on synthetic path data and not ground truth sensor data and may not be capable of detecting abnormal non-recurrent congestion in real-time. More recently, \citet{hassan2019spatio} proposed a method for detecting spatio-temporal anomalies in ITS leveraging traffic sensor data with a 5-minute resolution. Experiments are conducted to determine the classification precision and optimal parameters for historical data length and training window, with the optimal parameter set having around 70\% classification precision and a running time of less than one second. F1 score is used to account for precision and recall in the performance analysis, demonstrating a trade-off between classification precision and recall as the historical data length and sliding training window sizes increase. However, the authors' dataset consists exclusively of highway ramp detectors, and it is uncertain if this method will operate satisfactorily on links of different functional classes. Moreover, the performance of various network sizes is not analyzed. Further research will be necessary to elucidate the effects of varying network topology and spatial data granularity for anomaly detection in ITS.

Few existing works notably leverage DL models to classify anomaly traffic congestion, providing a potential future research direction. In more recent work, \citet{davis2020framework} proposed an end-to-end deep learning-based anomaly detection method for transportation networks. The authors leverage an LSTM model with a customized extreme value theory-based objective function to classify anomaly events across multiple datasets, including speed, travel time, and taxi demand data. Compared to the existing statistical, ML, and hybrid DL approaches, the proposed approach has the best F1 score for classification across most datasets. The authors also outline additional avenues for future work, such as exploring different objective functions, testing the method on other datasets, and proposing strategies for identifying and quantifying the factors that cause the anomaly.

Moreover, it would be interesting to test different combinations of DL models for classifying anomaly events. LSTM and GCN model combinations have demonstrated considerable success in predicting traffic flow and speed in small-scale and network-wide contexts using low-resolution temporal data sequences (e.g., 5 minutes). A GCN model could extract important spatial features from the network topology for classifying anomaly events in real-time, while the online LSTM model processes recent temporal data sequences to classify the current state as recurrent or non-recurrent. However, the training time can be long in large-scale transportation networks with fine-grained data resolution. Furthermore, these models have degraded performance when predicting under perturbation (e.g., non-recurrent network events). To address these shortcomings, research into a hybrid and real-time system integrating models optimized for non-recurrent event identification and prediction under perturbation, to be used dynamically with the recurrent prediction models, presents an interesting future direction.

Prediction under perturbation presents another problem relating to the response effort for resolving non-recurrent traffic congestion: efficiently rerouting the existing traffic. Practically, rerouting travelers requires selecting less optimal routes, and the response effort may temporarily change the network topology, hindering the performance of route guidance applications leveraging existing DL models. The challenge of predicting under perturbation can be partially attributed to the fact that most literature focuses on the recurrent traffic prediction problem and trains their models with preprocessed data where anomaly events have been normalized or removed. Some researchers have recently challenged the community to rethink how we treat anomaly data, illustrating its potential to elucidate new insights when analyzed separately from the recurrent data \citep{john2021outlier}. Currently, benchmark datasets containing a sufficient density of anomaly events are challenging to find. A standardized benchmark dataset for this application is of sufficient need to the research community. In the interim, some potential datasets for future experimentation found in the existing literature include travel time, vehicle occupancy, and traffic speed data from the Twin Cities Metro Area\footnote{\url{https://github.com/numenta/NAB/tree/master/data}}.

\subsection{Traffic Prediction in Mixed Traffic Transportation Networks}

Automobile-related traffic congestion is a significant problem in transportation networks that adversely impacts travel time reliability. The consensus in the transportation community over the past two decades is that we can not build ourselves out of congestion \citep{downs2004traffic}, emphasizing the need for improving access to multi-modal travel. Consequently, increased emphasis on building multi-modal mobility systems, in contrast to personal vehicle infrastructure, has resulted in a mixed traffic environment, especially in urban areas. One of the most promising applications of traffic prediction for relieving congestion in mixed traffic transportation networks relates to OD travel time prediction problems: scheduling complete trips. A complete trip can be defined as an origin to destination trip encompassing multiple modes (e.g., walking, bicycle, metro, automobile).

Efficiently scheduling complete trips given an origin and destination pair encompasses two essential functions: multi-modal and time-dependent traffic prediction and shortest path algorithms. Ideally, reliable traffic prediction models trained on historical data should inform the shortest path algorithms. The body of literature on multi-modal shortest path algorithms is extensive and includes \citet{li2010activity}, \citet{ zhou2008dynamic} and \citet{ bielli2006object}. However, these works make many assumptions regarding individual decision-making, transit schedules, and travel time reliability. Moreover, they generally only consider a subset of modes and exclude non-automated modes such as walking and bicycling. They also do not employ DL models for predicting traffic for consideration when scheduling paths. It would be beneficial to experiment with integrating reliable multi-modal OD and path travel time DL models to inform the time-dependent shortest path algorithms and expand their multi-modal decision frame. A system for scheduling complete trips while accounting for individual considerations can alleviate automobile congestion and provide more reliable and cost-effective travel opportunities while expanding societal access to mobility.

That being said, mixed traffic environments are the least represented in the literature due to a limited understanding of complex and unknown interactions and relationships between human vehicles, autonomous vehicles, bicycles, pedestrians, public transportation, etc. More research into traffic prediction within the mixed traffic environment will be required to improve our understanding of the diverse interactions and spatio-temporal correlations. Data availability presents one barrier to future research in this area, as some transportation networks have limited spatial data coverage of pedestrian facilities. At the same time, datasets for some modes, such as pedestrian volumes, are also limited in availability. Existing datasets also generally contain data from a single mode, and constructing a multi-modal dataset for benchmarking various approaches would benefit future research. Moreover, an interdisciplinary effort between traffic engineers, behavioral scientists, and computer scientists will likely be necessary to account for the human decision-making process during modeling. In summary, it is currently unknown if DL models can support a multi-modal traffic prediction application, as the existing models primarily focus on traffic prediction of a single mode.

\subsection{Data Challenges in Traffic Prediction}
In this subsection, we outline some existing challenges in traffic prediction research related to input data for training DL models. 

\begin{itemize}

\item \textbf{Limited Data Accessibility and Synthetic data}. The increasing number of sensors deployed within the traffic network in recent years has led to high volumes of collected data in some geographic areas; however, accessing this data for modeling purposes can be challenging in practice. Much of this data is collected by private companies (e.g., Uber, DiDi), as well as government organizations (e.g., TMC and Department of Transportation (DoT)), and is not always made publicly available. Moreover, if the desired data exists but is collected by a private data broker (e.g., INRIX), the acquisition cost can be quite high, restricting access for some researchers. In other cases, the data used in an experiment is provided by a partnership or grant with a local or federal organization and is not readily available to the public for results comparison and validation against other approaches. As a potential solution to data accessibility problems, some researchers have experimented with methods for generating artificial traffic data to augment their existing datasets or to create entirely new datasets that mimic the distribution of the ground truth data. Recently, the GAN attention-based neural network model has shown promise in this area, and many attempts have been made to generate actionable spatio-temporal data for future modeling efforts \citep{gao2022generative}. GAN-based techniques are widely applicable to improving traffic prediction methods where data sparsity is a concern. However, only a few recent research works focus on the traffic data imputation problem in the literature \citep{huang2021deep, kazemi2021igani}.

\item \textbf{Benchmarking Traffic Prediction}. A primary need in future traffic prediction research is a comprehensive database of benchmark datasets for comparative analysis of the proposed methods. As more traffic data become available each year, the diversity of datasets leveraged within the literature for analyzing the performance of proposed approaches has become quite diverse. For example, Table \ref{openacc} provides a list of 40 of the most commonly used large-scale ground truth datasets leveraged for traffic prediction DL models within the literature. The high variability between datasets used in existing case studies is notable because the performance of any prediction model is dependent not only on the model architecture but also on the underlying data characteristics (e.g., distribution, density, quality, features). As a result, cross-comparison of results between approaches trained and evaluated on different datasets does not provide much actionable insight for real-world stakeholders seeking to implement DL models for practical applications. We believe further standardization of traffic datasets within the literature will elucidate new insights into the practical performance trade-offs of the vast array of traffic prediction approaches within the existing literature. Some researchers have recently attempted to pave the way towards standardization by providing open-sourced platforms for accessing diverse sets of unified traffic data, such as LibCity \citep{wang2021libcity}, laying the foundation for future work in the area of standardization.

\item \textbf{External Data Restrictions}. Many works have demonstrated the benefits of considering external data features in their prediction models (e.g., meteorological data, pandemic data, event data, social media data, and work zone data), as illustrated in Table \ref{table:exter}. However, the literature on external data-based traffic prediction is somewhat limited due to the challenges of synthesizing the external data with the ground truth traffic data. This issue is also discussed in another survey \citep{tedjopurnomo2020survey}, where the authors suggest first establishing a benchmark dataset of ground truth traffic data with sufficient spatio-temporal coverage and data density. After that, the traffic data can be concatenated with supplementary data if needed. Lastly, each temporal data instance can be augmented to include the necessary geographical external data, such as weather and accident information. This way, a single comprehensive benchmark dataset is built from possibly disjoint spatio-temporal and external datasets. We believe this is a sound and practical approach to building external datasets for benchmarking. A benchmark dataset in this format would greatly aid future researchers in comparative analysis, enabling the elucidation of underlying spatio-temporal relationships between the traffic and external data.

\item \textbf{Multi-source Data}. We define multi-source data as a comprehensive dataset combining multiple types of ground truth traffic data, such as flow and speed data, with external data sources (e.g., weather, social media, and event data). Most existing traffic prediction research generally focuses on experimenting with a single dataset for their proposed models (e.g., speed data, flow data, trip data). Consequently, the influence of other related factors on the overall prediction outcome is not always thoroughly investigated. Furthermore, non-linear interactions among different data series are generally not considered and are treated as separate features. In the future, when attempting to forecast traffic conditions, we should consider adapting traditional approaches to fuse information from multiple sources and varying datasets to improve our understanding of the underlying correlations between the various transportation data types.

However, finding effective methods for fusing the multitudes of raw data from varying sources into one comprehensive and overarching model becomes a challenging and open research problem. Besides, high-dimensional features create increasingly high computational costs for training, so it also becomes necessary to determine the feature importance and only extract those that will be most beneficial to the application goals. Attention-based DL approaches have recently found success in the literature in quantifying the relevance of various features. SAE DL models can also help to reduce feature counts in multiple applications where the feature set is large (e.g., network-wide traffic prediction) by hierarchically extracting the most important trends and discarding the other information. GCN networks have also demonstrated some success translating traffic data into a graph-based representation capable of capturing and encoding the spatial relationships within the traffic network at the node and link level. We believe further research into these approaches will be fundamental to improving the predictive performance of the resulting traffic prediction models.

\item \textbf{Data Collection Area}. The distinct topological structures of various transportation networks are a primary consideration when training data-driven traffic prediction models. In the cutting-edge approaches for applications such as network-wide traffic prediction, GCN and other graph-based methods are leveraged to capture the spatial properties of the transportation network and translate them to actionable features for temporal prediction models such as LSTM. Consequently, the cutting-edge approaches perform well when predicting traffic within the network they were trained in but tend to fail when attempting to predict outside of this context. Theoretically, we could train specialized prediction models for each subset of the transportation network, but this is not the most efficient approach. Practically, we seek to answer whether these existing models can be adapted to work in other transportation networks or if we can leverage the current models to speed up the training of newer models.

Recently, transfer learning has emerged as a deep learning technique for rapidly building new models by leveraging the wisdom of existing models. However, transfer learning literature for the traffic prediction application is minimal. We believe transfer learning methods will become increasingly important in the future of traffic prediction as a means of solving the adaptability problem of large-scale traffic prediction models. In addition to adaptability, the data collection area also influences the focus of recent studies. Generally, the widely used traffic datasets come from four main collection areas: regional urban transportation networks, expressways, freeways, and highways. Currently, most studies focus on utilizing available freeway, highway, or expressway data to achieve high accuracy. At the same time, traffic prediction in urban arterial networks is more challenging in practice due to the increasingly complicated network structure and the existence of signalized intersections \citep{stathopoulos2003multivariate,advani2020performance}. Notably, less emphasis has been applied to the urban arterial prediction problem. That being said, some current works focus on broadened decision frames that consider not only a single corridor or expressway but also the impacts of neighboring links and intersections within the urban network \citep{yu2017spatiotemporal}, while \citet{advani2020performance} improved the representation of the various vehicle categories found in urban systems, such as cars, bicycles, pedestrians, and public transit vehicles within the models. However, these advances lead to more complex prediction models (e.g., complex hybrid DNNs), which require more data and significantly increased time to train effectively.

\end{itemize}
\section{Conclusion}
\label{Sec7}

Although a variety of traffic forecast methods have been proposed in the literature, ranging from statistical modeling to machine learning, traffic forecasting remains a challenging task due to 
the complex relationships between the data, external factors, non-linearity, and the inherent spatial and temporal dependencies. While deep learning models have made great strides in improving the accuracy and scale of traffic prediction methods, there is still much work to be done. In this paper, we conduct a broad survey on the current state of traffic prediction research. More specifically, we first comprehensively outline the various data types used in traffic prediction and their limitations, and then we survey the existing work on preprocessing this traffic data. Next, we summarize the classical and current traffic prediction methods and provide a taxonomy. Additionally, we survey different traffic prediction tasks and analyze the existing approaches within the literature, focusing on multivariate traffic time series modeling applications. Lastly, some significant challenges, as well as future research directions, are discussed. This paper aims to provide a suitable reference for readers to quickly get up-to-speed regarding the current state of traffic prediction and provide a strong reference for guiding future research work in this field.

\section*{Author Statement}
The contribution of the authors are listed as follows:
\begin{itemize}

\item Maryam Shaygan$^*$\let\thefootnote\relax\footnote{$^*$ These authors contributed equally to this work}: Investigation, Writing - Original Draft.

\item Collin Meese$^*$: Investigation, Writing - Original Draft.

\item Wanxin Li: Review $\&$ editing.

\item Xiaoliang Zhao: Review $\&$ editing.

\item Mark Nejad: Supervision - Review $\&$ editing.
\end{itemize}

\section*{Acknowledgement}
This research is supported in part by a Federal Highway Administration grant: Artificial Intelligence Enhanced Integrated Transportation Management System, 2020-2023. The authors thank Gene Donaldson (DelDOT TMC) for helpful discussions and insights.

\bibliography{cas-refs}

\end{document}